%% file: 00-main.tex
\documentclass{jdmdh}
\pdfoutput=1
\usepackage{array}
\usepackage{pgfplots}
\usepackage[mode=buildnew]{standalone} 
\usepackage{tabularx}
\usepackage{booktabs} 
\usepackage{amsmath,bm} 
\usepackage{subcaption}
\usepackage[scaled=0.86]{helvet} 
\pgfplotsset{compat=1.14}
\usepackage{booktabs}
\newcommand{\ra}[1]{\renewcommand{\arraystretch}{#1}}
\usepackage{amssymb}

\title{Combining Visual and Textual Features for \\Semantic Segmentation of Historical Newspapers}

\author[1]{Rapha\"el Barman}
\author[1]{Maud Ehrmann}
\author[2]{Simon Clematide}
\author[1]{Sofia Ares Oliveira}
\author[1]{Fr\'ed\'eric Kaplan}
\affil[1]{\'Ecole polytechnique f\'ed\'erale de Lausanne, Switzerland} 
\affil[2]{Universit\"at Z\"urich, Switzerland} 

\corrauthor{Rapha\"el Barman}{raphael.barman@epfl.ch}

\begin{document}

\maketitle

\abstract{
The massive amounts of digitized historical documents acquired over the last decades naturally lend themselves to automatic processing and exploration. Research work seeking to automatically process facsimiles and extract information thereby are multiplying with, as a first essential step, document layout analysis. Although the identification and categorization of segments of interest in document images have seen significant progress over the last years thanks to deep learning techniques, many challenges remain with, among others, the use of more fine-grained segmentation typologies and the consideration of complex, heterogeneous documents such as historical newspapers. Besides, most approaches consider visual features only, ignoring textual signal. We introduce a multimodal neural model for the semantic segmentation of historical newspapers that directly combines visual features at pixel level with text embedding maps derived from, potentially noisy, OCR output. Based on a series of experiments on diachronic Swiss and Luxembourgish newspapers, we investigate the predictive power of visual and textual features and their capacity to generalize across time and sources. Results show consistent improvement of multimodal models in comparison to a strong visual baseline, as well as better robustness to the wide variety of our material.
}


\input{01-introduction.tex}
\input{02-background.tex}
\input{03-method.tex}
\input{04-experimental_setup.tex}
\input{05-experiments.tex}

\input{06-discussion.tex}

\section*{Author contributions}
\vspace{1ex}
RB designed and carried out the experiments, wrote the paper; ME designed and supervised the project and experiments, wrote the paper; SC designed and supervised the project and experiments, wrote the paper; SO supervised the project and experiments, helped with the paper writing; FK supervised main directions.

\section*{Acknowledgments}
\vspace{1ex}
We warmly thank the journal \textit{Le Temps} and the Swiss and Luxembourgish National Libraries for giving us access to their newspaper archive collections in the context of the \textit{impresso} project. We also thank Julien Nguyen Dang for his contribution to the annotation of part of the data. Finally, the second and third authors also gratefully acknowledge the financial support of the Swiss National Science Foundation (SNSF) for the project `\textit{impresso} – Media Monitoring of the Past' under grant number CR-SII5\_173719.

\bibliographystyle{plainnat}
\bibliography{references}

\appendix\footnotesize

\end{document}

%% file: 01-introduction.tex
\section*{Introduction}
\vspace{1ex}

For several decades now, digitization efforts are slowly but steadily contributing to an increasing amount of facsimiles of cultural heritage documents. As a result, it is nowadays commonplace for many memory institutions to create and maintain digital repositories that offer rapid, time- and location-independent access to documents, allow to virtually bring together disperse collections, and ensure the preservation of fragile documents thanks to on-line consultation \citep{terras_rise_2011}. Beyond this great achievement in terms of preservation and accessibility, the next fundamental challenge --and real promise of digitization-- is to exploit the \textit{contents} of these digital assets, and therefore to adapt and develop appropriate document and language processing technologies to search and retrieve information from this `Big Data of the Past' \citep{kaplan_big_2017}.

\paragraph{Context} Efforts are, in this regard,  well under way and the libraries, digital humanities (DH), natural language processing (NLP), and computer vision (CV) communities are pooling forces and expertise to push forward the processing of facsimiles, as well as the extraction and linking of the information contained therein.\footnote{These interdisciplinary efforts were recently streamlined within the far-reaching project `Europe Time Machine': \href{https://www.timemachine.eu/}{https://www.timemachine.eu}}
This momentum is particularly vivid in the domain of digitized newspaper archives for which there has been a notable increase of research initiatives over the last years. Those range from individual works dedicated to the development of tools \citep{yang2011topic,dinarelli2012tree,moreux_innovative_2016,wevers-2019-using} or the usage of those tools \citep{kestemont2014mining,Lansdall-WelfareE457}, to evaluation campaigns \citep{rigaud:hal-02304334,clausner_prima_2019}, including the emergence of large consortia projects seeking to apply computational methods to historical newspapers at scale, such as \textit{ViralTexts}\footnote{A project aiming at mapping networks of reprinting in 19th-century newspapers and magazines (US, 2012-2016): {\href{https://viraltexts.org/}{https://viraltexts.org}}}, \textit{Oceanic Exchanges}\footnote{A project tracing global information networks in historical newspaper repositories from 1840 to 1914 (US/EU, 2017-2019): {\href{https://oceanicexchanges.org/}{https://oceanicexchanges.org}}}, \textit{impresso}\footnote{ {\href{https://impresso-project.ch/}{https://impresso-project.ch}}},  \textit{NewsEye}\footnote{A digital investigator for historical newspapers  (EU,  2018-2021): {\href{ https://www.newseye.eu/}{https://www.newseye.eu}}}, and \textit{Living with Machines}\footnote{A project which aims at harnessing digitised newspaper archives (UK, 2018-.):  {\scriptsize\href{https://www.turing.ac.uk/research/research-projects/living-machines}{https://www.turing.ac.uk/research/research-projects/living-machines}} } \citep{Ridge:271329}. 

Overall, this research contributes a pioneering set of text and image analysis tools, system architectures, and interfaces covering several aspects of historical newspaper processing. They usually focus on all or part of the typical digitized newspaper pipeline which consists, essentially, of three main steps: facsimile processing, in order to derive the structure and the text from the document image (via, respectively, optical layout recognition and optical character recognition processes); content enrichment, in order to extract and link relevant information from both textual and visual part of the contents; and, finally, exploration support, in order to search and visualize the enriched resources via e.g. application programming or graphical user interfaces.

\paragraph{Motivation} While encouraging, these efforts are still at an early stage and many challenges have yet to be addressed, especially with respect to document layout analysis (first processing phase). Document layout analysis aims at segmenting a document image into meaningful segments and at classifying those segments according to their contents \citep{eskenazi17_survey_text_doc}. Two types of classification are traditionally distinguished: physical layout analysis, with a focus on the nature of the content (is this segment a textual block, a diagram, a picture, a decoration, a graphic, etc.), and logical layout analysis, with a focus on the function of the content (is this textual block a title, a footer, an article, etc.). Those segments are then fed into optical character recognition (OCR) programs that recognize their textual content.  

With newspapers, these image segmentation and classification processes are particularly difficult because of the complexity and diversity of the object. A newspaper page consists of multiple, heterogeneous elements which feature different layout characteristics (text, map, table, illustration), different contents (regular articles, serial, advertisements) and which, additionally, evolve through time, differ according to newspapers, and are in different languages. Besides, facsimiles can be of variable quality due to the conservation state of the originals and this can also affect layout analysis performances.

Although difficult, layout analysis is however essential for historical newspaper understanding and exploitation, and their quality has a direct impact on downstream processes \citep{10.1145/3355610}. From an information retrieval and user viewpoint, being able to query at the level of meaningful segments such as articles --instead of whole pages--, and to facet over different types of segments are undeniable advantages. From an NLP viewpoint, most analysis of semantic nature such as entity linking, topic modelling or text classification requires and/or performs far better on semantically self-sufficient, autonomous content items. For some processes, it can also be useful to filter out unwanted elements, either because of too noisy in terms of OCR or less relevant in terms of contents (e.g. transport schedule, cross-words, weather reports, TV programs, etc.). Finally, from a media history viewpoint, the automatic classification of content items can enable a better understanding of the evolution of newspaper sections through time and across collections.

Finally, the task of newspaper segmentation is also to be seen within the current context of large-scale newspaper projects. Facing both the digitized newspaper material reality and  user needs, these initiatives help, on the one hand, reveal the defects of legacy layout and text acquisition outputs from libraries and, on the other, emphasize the needs of finer-grained qualification of newspaper sections for scholarship purposes, as well as of efficient large-scale, trans-collection and diachronic processing of newspaper facsimiles. In this regard, the `\textit{impresso} - Media Monitoring of Past'\footnote{\href{https://impresso-project.ch}{https://impresso-project.ch}} project --in the context of which the present work was carried out-- is a case in point. Led by an interdisciplinary team, \textit{impresso} aims at semantically indexing a multilingual corpus of digitized newspapers and integrating the resulting data into historical research workflows by means of a newly developed user interface\footnote{\href{https://impresso-project.ch/app}{https://impresso-project.ch/app}}. By doing so, it appeared desirable to compensate for the deficiencies of old layout analysis.

\paragraph{Proposition} In this context, this paper presents an innovative approach for the semantic segmentation of historical newspapers. `Semantic' in that the targeted image segment typology goes beyond physical and/or logical characteristics and considers fine-grained semantic content item types (e.g. a segment is not only an article, but also e.g. a serial or death notice, or not only a table, but also e.g. election results or stock exchange information). `Innovative' in that the approach makes joint use of visual and textual features, in an attempt to replicate human comprehension which uses both modalities simultaneously when confronted with document images. Already tested in very few recent studies \citep{yang17_learn_extrac_seman_struc_docum, katti18_charg, dang_endtoend_information_extraction, denk19_bertgrid_contextualized_embedding}, we believe it is the first time a multi-modal document image segmentation approach is applied on newspapers, what is more of historical nature. 

\paragraph{Objective} Our objective is twofold. First, we wish to assess whether the combination of visual with textual features can efficiently segment newspapers images. In this regard, the recent advances of deep learning approaches for semantic image segmentation and text processing suggest that positive results can be achieved: visual-based neural architectures trained for natural images have shown good adaptation to document images, and single architectures have demonstrated their capacities to adapt to different tasks \citep{oliveira18_dhseg}. As for text, language models based on embeddings have shown their capacity to support a variety of tasks, from named entity recognition to question answering \citep{collobert2011natural}. Second, we wish to investigate whether this multi-modal representation can better support generalization across time and newspapers. The same newspaper section can indeed change drastically in terms of layout through time and across titles while enjoying a certain stability in terms of textual contents. 

\paragraph{Contributions} We present a series of experiments for the segmentation of several newspapers covering different time periods according to four semantic classes. These experiments are based on a modified version of \textit{dhSegment}, a generic deep-learning approach that operates pixel-wise document segmentation \citep{oliveira18_dhseg}.  Architecture's code, ground-truth data sets as well as models are publicly released.

Section \ref{sec:background} presents prior works and specifies where the present approach sits with respect to them. Section \ref{sec:method} introduces the approach, and Section \ref{sec:setup} details the experimental setup. Section \ref{sec:experiments} reports and discusses three series of experiments and Section \ref{sec:discussion} considers the limits, but also future application scenarios of the approach and concludes.

%% file: 02-background.tex
\section{Related work}
\label{sec:background}
\vspace{1ex}

The survey of \cite{eskenazi17_survey_text_doc} gives an overview of the approaches for the segmentation of textual document images. Approaches are usually divided into three categories: top-down, when starting from the whole page in order to partition it, bottom up, when starting from small components in order to aggregate them, and hybrid. Classical algorithms heavily rely on specific document priors, e.g. having a ``Manhattan'' layout, and/or require large amounts of hand-crafted features. More recent approaches make use of deep neural networks, trading prior, hand-crafted features for the learning capacities of machine learning, especially deep neural networks. Those include the usage of convolutional neural networks 
\citep{chen17_convolutional_neural_networks}, as well as several variants of the fully convolutional network (FCN) introduced by \cite{long15_fully} \citep{he17_multi_scale_multi_task_fcn, xu17_page_segmentation_historical, wick18_fully_convol_neural_networ_page, oliveira18_dhseg}. 

Considering newspapers images, several works have been proposed for their segmentation. \cite{hebert2014automatic} proposed an approach that performs physical and logical segmentation, and detects reading order on historical French newspapers. It is based on conditional random fields and a set of heuristics, targets high-level types such as titles, line and articles and achieves state of the art results with ca 85\% of accuracy. A similar coarse-grained classification (line, image, illustation, text blocks) is done by \cite{gatos1999integrated} on Greek newspapers using an hybrid approach, and by \cite{hadjar2003arabic} and \cite{bouressace2018recognition} on contemporary Arabic newspapers using Run Length Smoothing Algorithm (RLSA). \cite{lorang2015developing} focuses on a more specific type,  that is poetic content items, and make use of manually crafted features to classify crops of newspaper images.

On the other side of the spectrum, another line of research performs newspaper content segmentation using text only (usually when images are not available) via the detection of homogeneous passages based on sentence or paragraph textual similarity \citep{riedl2019clustering}. Those approaches can detect and classify segments of textual nature exclusively, but cannot identify their image boundaries, nor take into account more visual items.

Only a few recent work attempt to make use of image and/or localized, two-dimension text information. \cite{meier17_fully_convol_neural_networ_newsp_artic_segmen} use a FCN based on image and OCR output information in order to detect articles in newspaper images (no further segment types). In this case text is reduced to a binary feature information (a pixel has text or not) and the lexical and semantic dimensions are not taken into account. \cite{katti18_charg} introduced the concept of \textit{chargrid}, a two-dimension representation of text where characters are localized on the image (thanks to the box coordinates) and encoded as a one-hot vector. This information is passed through an architecture that uses two encoders, one for the image information, the other for the character one and two decoders, one that produces semantic segmentation and the other that produces bounding boxes. Different model variants (image only, text only, both) are applied on images of administrative documents (invoices), and experiments show that the models based on both signals achieve better results. This is however opposed to a high-computing cost, as emphasized by the authors.
\cite{dang_endtoend_information_extraction} builds on this work and present an approach based on a multi-stage attentional U-Net using a one-hot encoded character feature. Segmentation of template like administrative documents yield state of the art results in the order of 87\% mIoU (see Section \ref{sec:eval}). \cite{denk19_bertgrid_contextualized_embedding} also extends \cite{katti18_charg}, considering not only characters, but words and their corresponding embeddings, with \textit{BERTgrid} for the automatic extraction of key-value information from invoice images (amount, number, date, etc.). With the same architecture as \cite{katti18_charg}, they obtain best results with document representation based on one-hot character embeddings and word-level BERT embeddings \citep{devlin-etal-2019-bert}, with no image information. Performances differ quite a lot between classes (of key-value types). Finally, \cite{yang17_learn_extrac_seman_struc_docum} jointly uses visual and textual features in a network, via \textit{text embedding maps} where the two-dimension text representation is mapped to the pixel information (cf. Section \ref{sec:method}). Textual features correspond here to sentence embeddings (average of words vectors obtained with word2vec \citep{mikolov2013efficient}), and models are trained on several variants of an end-to-end, multi-modal fully convolutional network for the segmentation and coarse classification of image regions (figure, table, section heading, caption, list, paragraph). Models are tested on various datasets and results show significant, although variable across classes, performance improvements with the model using both visual and textual features.

The method we present builds on the work of \cite{yang17_learn_extrac_seman_struc_docum} in the sense that it also makes use of text embedding maps. It however differs in that we work with historical newspapers --therefore integrating the diachronic dimension--, target a more fine-grained segment typology and experiment with different embeddings.

%% file: 03-method.tex
\section{Method}
\label{sec:method}
\vspace{1ex}

Our objective is to segment newspaper images and to classify detected segments according to a fine-grained newspaper section typology. To this end, we introduce a method which performs supervised, pixel-wise multiclass classification using both visual and textual features. The method builds on \textit{dhSegment}'s architecture.

\subsection{Primary Architecture: \textit{dhSegment}}
\vspace{1ex}

\textit{dhSegment} is an open-source, generic image document segmentation framework\footnote{\href{https://github.com/dhlab-epfl/dhSegment}{https://github.com/dhlab-epfl/dhSegment}} \citep{oliveira18_dhseg}. It consists of a CNN-based pixel-wise predictor coupled with task dependent post-processing blocks. Its network is based on a U-Net architecture \citep{ronneberger15_u_net}, where the encoder follows a deep residual network ResNet-50 \citep{he2016deep} pre-trained on ImageNet \citep{deng2009imagenet}. \textit{dhSegment}  has demonstrated competitive results on multiple tasks, e.g. page extraction, baseline extraction, and layout analysis, thereby paving the way for efficient and generic document image segmentation. Its architecture is here modified in order to incorporate textual features.

\subsection{Text embedding map}
\vspace{1ex}

Considering textual and visual information at the same time supposes to jointly encode their signals. To this end, and as briefly introduced in Section \ref{sec:background}, it is possible to map the one-dimensional representation of textual information (e.g. a word vector) into a three-dimensional one by `positioning' the embedding representation into a two-dimensional space (e.g. a word has a certain width and height when written or printed on a page). This new textual embedding (corresponding to a word or a character) is therefore equivalent to the original vector, augmented with the positioning information (width and height). We refer to this three-dimensional representation of textual information, as introduced in \cite{yang17_learn_extrac_seman_struc_docum}, as a `text embedding map'. 

\begin{figure}[htb]
    \centering
    \hspace*{\fill}%
    \includegraphics[width=0.45\linewidth]{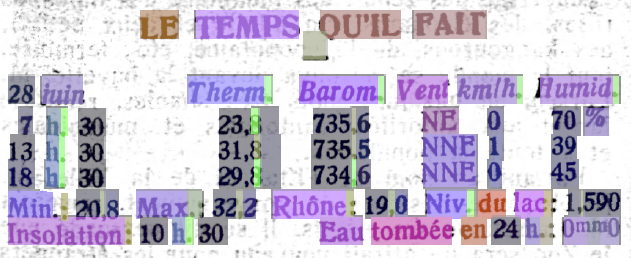}\hfill%
    \includegraphics[width=0.45\linewidth]{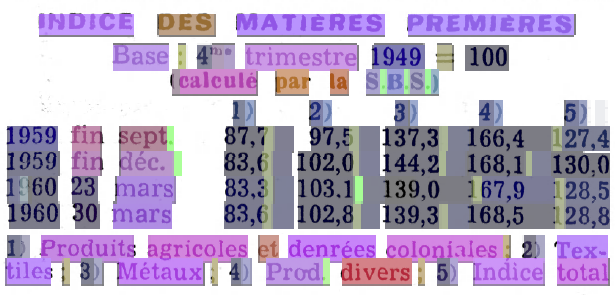}%
    \hspace*{\fill}%
    \captionsetup{width=0.9\linewidth}
    \caption{Visualization of a Flair-based, PCA-reduced text embedding map projected on three dimensions (red, green, blue).
    }
    \label{fig:pca_embeddings}
\end{figure}

The three-dimensional encoding of textual information is generated by using the results of an OCR process which outputs text tokens along with their coordinates on the image. Considering for example the left image of Figure \ref{fig:pca_embeddings}, an OCR engine produces the token ``\textsc{temps}'' located in the bounding box $[(10, 195), (10, 300), (40, 300), (40, 195)]$. Looking the token up in an embedding space returns its (textual) vector, which can then be associated with the bounding box information, thereby creating a three-dimension map.

This process can be formally defined as follows. Given an image of size $H \times W$ and a list of tokens $T$ where each token $t$ is associated with a bounding box $\mathbf{b}_t$ on the image, a text embedding map $G$ of size $H\times W \times N$ is produced, where $N$ is the dimension of the embeddings. Specifically, all pixels contained in the bounding box of a token $t$ are defined as the set $\mathbf{b}_t \in \mathbb{R}^2$ and each pixel $g_{i,j} \in G$ of the text embedding map is computed with

$$
g_{ij} =
\begin{cases}
  E(t) & \text{if } (i,j) \in \mathbf{b}_t \\
  0^{N} & \text{otherwise}
\end{cases}
$$

where $E(t)$ is a mapping of $t \to \mathbb{R}^N$ corresponding for example
to a word embedding, and $0^{N}$ is a null vector in case there is no text in the corresponding pixel.
Each pixel overlapping with a bounding box of a token is therefore mapped to its corresponding embedding. A pixel spanning two bounding boxes is attached to the one that has the closest center.

The final result is a text embedding map, \textit{i.e.} a three-dimension matrix where the first two dimensions correspond to the image-localized representation of the text, and the third to the embedding. Given that it has the same shape as the results of 2D convolutional layers, this construction offers the advantage that it can  be processed directly by classical image processing neural network.

One way of `visualizing' this embedding map is to project each word vector, using principal component analysis (PCA), to a new one of dimension three where each dimension corresponds to a color (red, green, blue). This produces a colored text embedding map where the third dimension (the textual one) is transformed into a color value. The idea here is to `see' the textual information, based on the fact that if two words have the same color, they also share the projection of their embeddings, and therefore textual features. Figure \ref{fig:pca_embeddings} shows such a colouring of textual information with segments of a weather forecast item (left) and a stock exchange table (right). Notwithstanding their similar layouts (\textit{i.e.} a table with same number of columns, with a title on top and some text below) and the drastic dimension reduction (2048 from the original vector to 3), it is possible to observe information about the text, with differences that could not be easily caught by visual features only. For example, numbers are grey, punctuation is green, stop-words have a yellowish tint, and the weather forecast segment contains a column with letters only.

\subsection{Model}
\label{sec:model}
\vspace{1ex}

\begin{figure}[h]
  \makebox[\textwidth][c]{%
  \includegraphics[width=\textwidth]{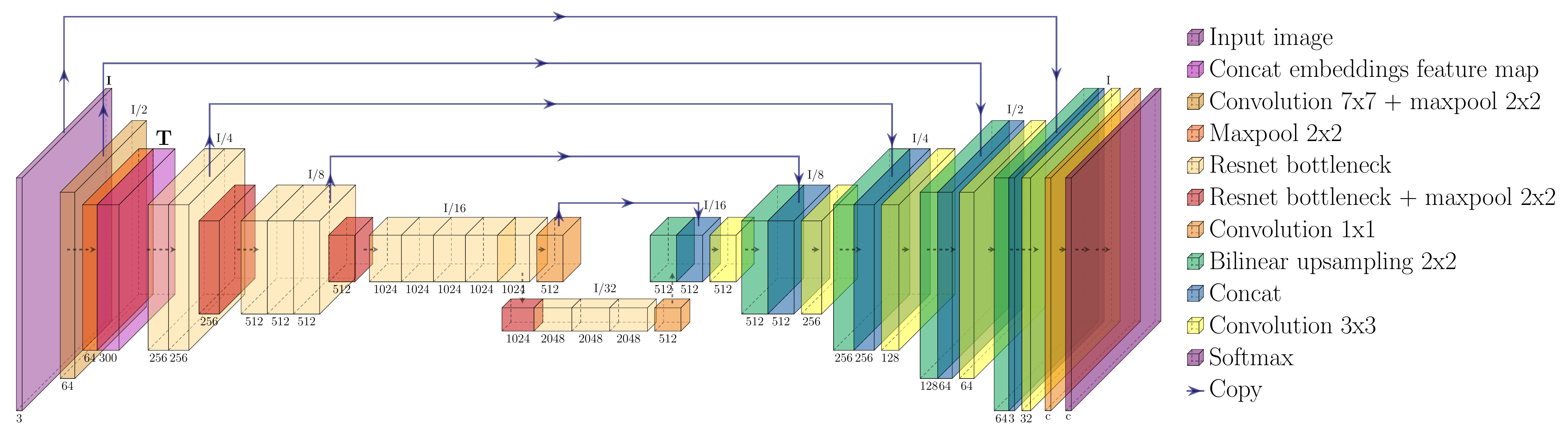}
  }
    \captionsetup{width=0.9\linewidth}
    \caption[The model architecture]{The model architecture used.
    }
  \label{fig:model_architecture}
\end{figure}

Our model architecture is a modified version of \textit{dhSegment},\footnote{More details on \textit{dhSegment} architecture can be found in the original paper.} where the only modification is the addition of the text embedding map. It takes as input an image of a newspaper and its corresponding text embedding map, and outputs a pixel probability map. Figure \ref{fig:model_architecture} displays the architecture, with the \textbf{T} marker indicating where the text embedding maps are concatenated (on the channel axis) to the visual feature maps. The size relative to the original image size $I$ is indicated at each step of the network, and the depth of the feature maps is indicated below the blocks for each step, considering that an embedding feature map of size 300 is input at \textbf{T}.

Variants of this model were tested during a pilot phase, in particular different input levels of the text embedding map. Two options were experimented in this regard: at the beginning of the network, in which case the text embedding map passes through most of the network and the textual signal is treated like the visual one, and at the end, in which case it adds further contextual information to the image feature map and support the final decision of the pixel class. Our preliminary experiments showed that inputting the textual features early in the network is the best option, so do all architectures used in Section\ref{sec:experiments}.

%% file: 04-experimental_setup.tex
\section{Experimental setup}
\label{sec:setup}
\vspace{1ex}

We apply this semantic segmentation method on historical newspapers of the \textit{impresso} project collection, considering four semantic classes. This section introduces the corpora and the typology used for classifying image segments (Section \ref{sec:dataset}), presents the embeddings used for the experiments (Section \ref{sec:embeddings}), specify the training setup (Section \ref{sec:training}) and details the evaluation framework (Section \ref{sec:eval}).

\subsection{Datasets}
\label{sec:dataset}
\vspace{1ex}

\subsubsection{Corpora}
\vspace{1ex}

Since the only freely available historical newspaper image dataset annotated with content item types considers broad categories alone (e.g. article, caption, header, etc.) \citep{clausner2015enp}, we created two new datasets.

\paragraph{Swiss newspapers} The first one originates from the Swiss National Library and the still-existing journal \textit{Le Temps}.\footnote{Both are partners of the \textit{impresso} project: \href{https://www.nb.admin.ch/snl/en/home.html}{https://www.nb.admin.ch/snl/en/home.html} and \href{https://www.letemps.ch/}{https://www.letemps.ch/}} It is composed of three titles in French language from the Romandy region with long publication history, namely: the \textit{Journal de Gen\`eve} (\textsc{jdg}, 1826-1994), the \textit{Gazette de Lausanne} (\textsc{gdl}, 1804-1991), and the \textit{Impartial} (\textsc{imp}, 1881-2017). \textsc{jdg} and \textsc{gdl}, issued in neighboring cities, can be considered as siblings and were merged in 1991. In order to have a long term, diachronic ground truth, newspaper issues were sampled across the whole publication spans with three issues every three years for \textsc{jdg} (used for training and evaluation) and every five years for \textsc{gdl} and \textsc{imp} (used for evaluation only). Because of misalignment problems between facsimiles and token coordinates of the original OCR, all images of the selected issues were re-OCRed with Abbyy FineReader application\footnote{Version 11, \href{https://www.abbyy.com}{https://www.abbyy.com}}. 

This material was manually annotated according to the four semantic classes of our typology (see Section \ref{sec:typo}), using the VGG Image annotator \citep{dutta2019via}. Annotation was done at the pixel level and not at the content item instance level, meaning that each pixel of the image has a label indicating that it belongs to a specific class (or not), but not that it belongs to a specific instance of a class. Several reasons motivate this annotation at pixel level: in most cases, there is only one instance per page, and in case of multiple instances, it might be non-contiguous regions of the same instance. Besides, instance separation and merging can also be done in a post-processing step.

This annotation process yielded a total of 1,982 annotated pages for \textsc{jdg}, 1,008 for \textsc{gdl} and 1,634 for \textsc{imp}. Table \ref{tbl:classes_swiss} shows the class distribution for the three titles. Pages without annotations do not contain any classified content items.

\begin{table}[t]
  \centering
    \begin{tabular}{r|rrrr}
      Class/Newspaper & \textsc{jdg} & \textsc{gdl} & \textsc{imp} & \textsc{luxwort}\\ 
      \midrule
      Serial & 137 & 108 & 103 & - \\ 
      Weather & 156 &  68 &  41 & - \\ 
      Death notice & 153 &  69 & 102 & 1765\\ 
      Stocks & 275 & 135 &  79 & - \\ 
      Pages w/o annotations & 1393 &  697 & 1326 & 17188\\
      \midrule
      Total & 1982 &  1008 & 1634 & 18953\\
    \end{tabular}
    \medskip
    \captionsetup{width=0.8\linewidth}
    \caption{\strut Dataset statistics. Note that page numbers do not add up to the total number of annotated pages because a single page can contain more than one class.\label{tbl:classes_swiss}}
    \hrule height 0pt
\end{table}

\paragraph{Luxembourgish newspaper}\label{sec:luxwort}

The second dataset consists of a single title, the \textit{Luxemburger Wort} (\textsc{luxwort}), a Luxembourgish newspaper from the Biblioth\`eque Nationale du Luxembourg\footnote{A partner of the \textit{impresso} project: \href{https://bnl.public.lu/fr.html}{https://bnl.public.lu/fr.html}} published since 1848 with contents in German, French and Luxembourgish. The library, who outsourced OCR and layout recognition for its newspaper collection, performed a manual check of the recognized segments. Having this at hand, we chose the work with the death notices which, in the \textsc{luxwort}, amount to ca. 90,000 segments in 17,000 page images. For our corpus, we sampled 34 issues per year between 1848 and 1950. This resulted in 18,953 images with 1,765 death notices. 

\subsubsection{Classes}
\label{sec:typo}
\vspace{1ex}

\begin{figure}[htb]
  \centering
  \subcaptionbox{\textit{Serial}\label{fig:ex_feuilleton}}{\includegraphics[width=0.65\textwidth]{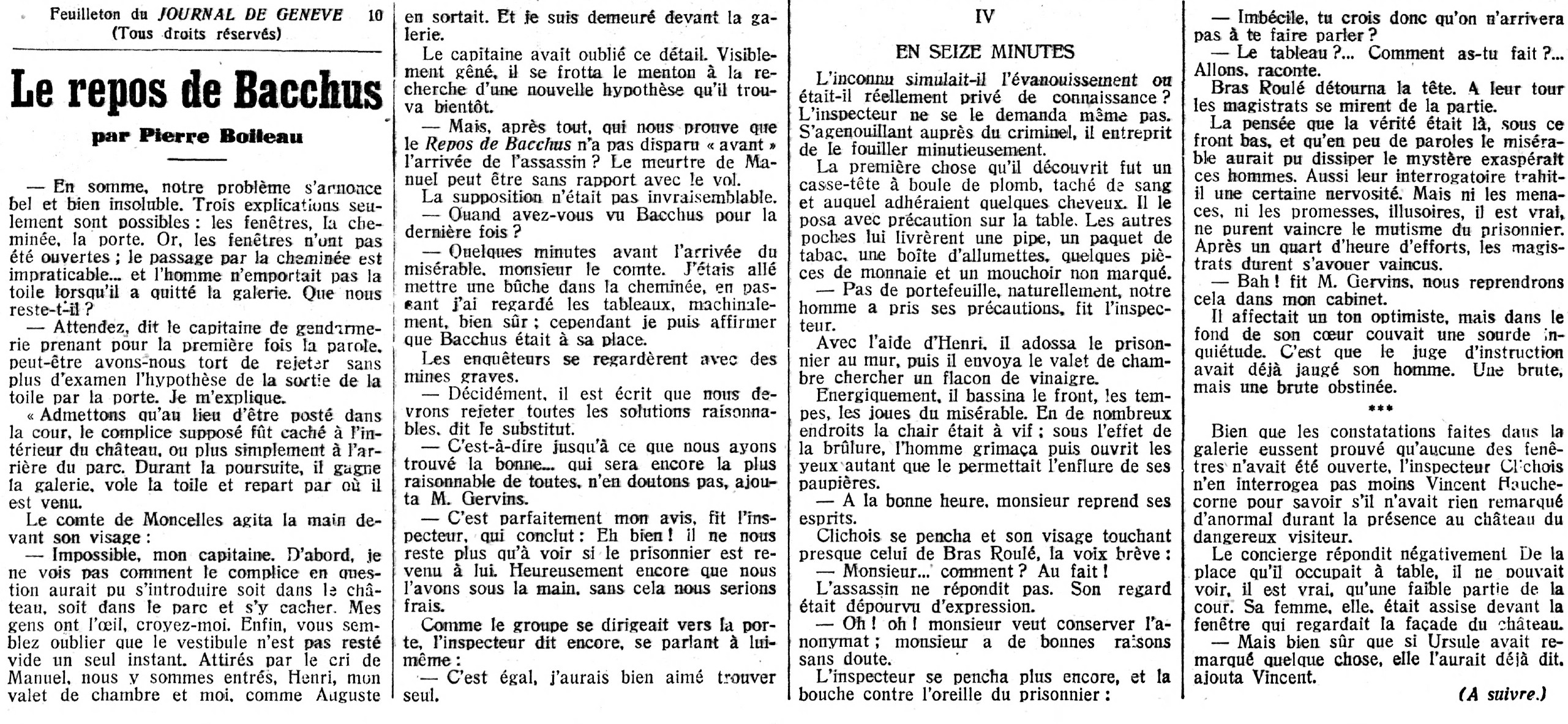}}\hfill
  \subcaptionbox{\textit{Weather forecast}\label{fig:ex_meteo01}}{\includegraphics[width=0.3\textwidth]{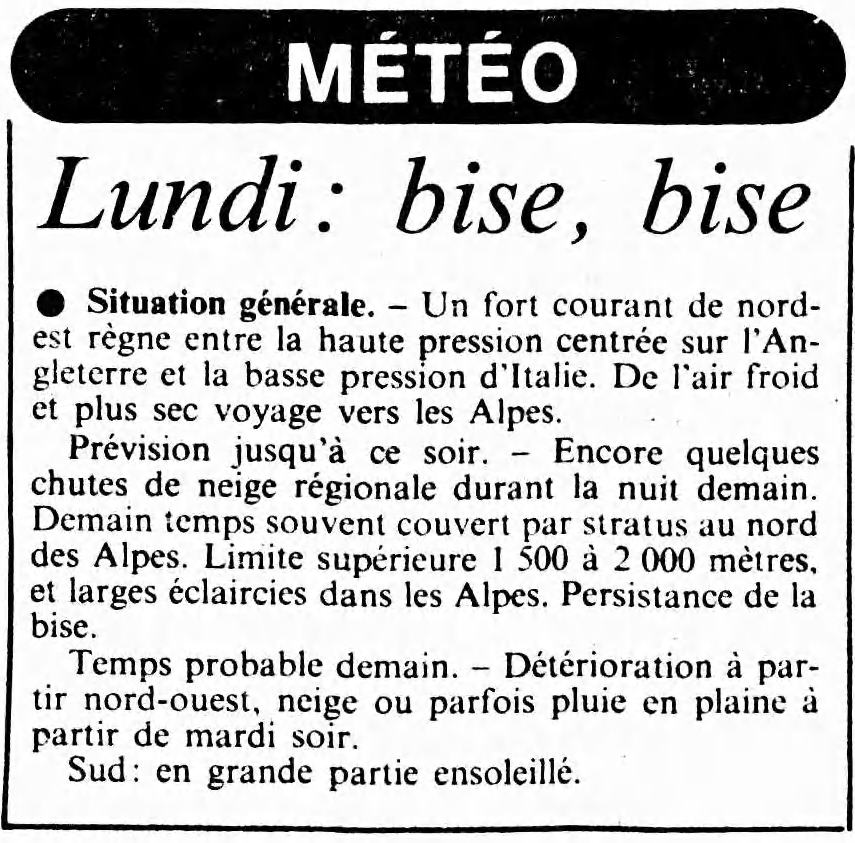}}%
\vspace{0.2cm}
  \hspace*{\fill}%
  \subcaptionbox{\textit{Death notice}\label{fig:ex_obituary}}{\includegraphics[height=2.5cm]{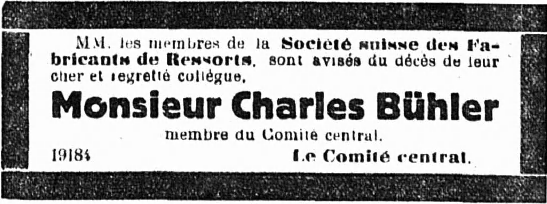}}\hfill
  \subcaptionbox{\textit{Stock exchange table}\label{fig:ex_stocks}}{\includegraphics[height=3.75cm]{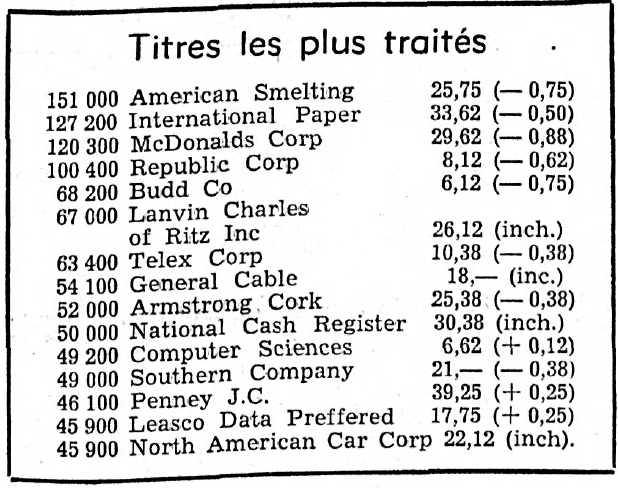}}%
  \hspace*{\fill}%
  \caption{\label{fig:ex_classes} Example images for each of the selected classes. All images are from
    the \emph{Journal de Genève}.}
\end{figure}

As mentioned earlier, newspapers feature a wide variety of contents which change over time and across newspapers. Given our objectives, we selected four classes of content items likely to, on the one hand, be of historical or practical interest (to use them as search facet or to filter them out before processing) and, on the other, present a mix of visual and textual variation:

\begin{itemize}
    \item[-] \textit{Serial}, i.e.~an excerpt of a bigger work published over time in several issues of a newspaper, corresponding to the French \textit{roman-feuilleton}. Serials often span several columns in a horizontal layout and can span several pages.
    \item[-] \textit{Weather Forecast}, i.e.~a text or illustration with the prediction of weather, or even a report of past weather measurements.
    \item[-] \textit{Death Notice}, i.e.~a small notice published by relatives of a deceased person.
    \item[-] \textit{Stock Exchange Table}, i.e.~a table reporting the values of different national stocks.
\end{itemize}

The degree of confusability of document image segments depends on several dimensions. First, the level of refinement of the typology naturally impacts what is confusable with what: distinguishing generic articles from advertisements is less difficult than distinguishing job adverts from purely commercial ones. The typology we consider here is already finer-grained compared to usual newspaper segmentation with e.g. a specific type of table among the tables (\textit{Stock Exchange}) and a specific type of article among the articles (\textit{Weather Forecast}). Next, within a given typology, visual and textual facets are the two main dimensions determining the degree of confusability of segments. Naturally, the more distinct on both dimensions the better. Finally, these dimensions are complemented by the time and source factors, since considering segments from different newspapers, and/or in synchrony or in diachrony also greatly impacts their confusability, not only with other types but also with themselves.

Let's examine our classes, shown in Figure \ref{fig:ex_classes}, in this light. Regarding the visual confusability of \textit{Serials} with respect to other items and themselves, in both synchrony and diachrony, they can be considered as rather distinct and stable: during most of their publication through time and across different newspapers, they are located at the bottom of the front page, topped with a thick black line. This makes them visually distinct compared to other items and should ensure good recognition performances using visual features only. As per their textual contents, these can vary and are, to some extent, confusable with regular journalistic contents. In contrast, \textit{Weather Forecast} segments feature a great visual variability over time (only text, then map and text, then only maps), but a clear textual stability. Here, the consideration of textual features should help improve recall, \textit{i.e.} removing false negatives. \textit{Death notices} are visually very similar to advertisements (small textual segments surrounded by a thick, black frame) but have different textual contents. For this class, one can therefore expect a high confusability with advertisements, and taking into account textual features should help to remove false positives. A similar situation holds for \textit{Stock Exchange Tables}: despite their distinctive layout compared to other content items, they are still visually confusable with other tables (e.g. transport, voting). They, however, enjoy a certain visual and textual stability and their recognition should not drastically suffer across time and sources. 

Overall, these four classes contain various combinations of visual and textual confusability, which makes them suitable for exploring the benefit of adding textual features  for semantic segmentation.

\subsection{Embeddings}
\label{sec:embeddings}
\vspace{1ex}

In order to investigate the effectiveness of different types of embeddings used to build the text embedding maps, we experimented with embeddings having different embedding levels (word or character), contextualized word representations or not (contextual or non-contextual), different languages (mono- or multilingual), and different training data (in- and out-domain). To this end, we use fastText word embeddings, which make use of characters $n$-grams to learn subword embeddings \citep{bojanowski2017enriching}; Byte-Pair encoded subword embeddings (BPEmb), which learn subwords rather than using fixed $n$-grams \citep{sennrich2015neural}; and character-based Flair embeddings \citep{akbik2018coling}, a character-level variant of the contextual string embeddings introduced in \citep{peters2018elmo}. In total, six flavors of these embeddings are considered, with three different stacks. Table \ref{tbl:embedd} summarizes the main characteristics of the used embeddings.

First, four pre-trained embeddings of the Flair library\footnote{\href{https://github.com/flairNLP/flair}{https://github.com/flairNLP/flair}} are used with their default implementation settings, as follows:
\begin{itemize}
    \item[-] \textit{fastText-fr}, \textit{i.e}~the French fastText embeddings of size 300 pre-trained on Common Crawl and Wikipedia; 
    \item[-] \textit{flair-fr}, \textit{i.e}~the French Flair embeddings of size 4096 pre-trained on Wikipedia;
    \item[-] \textit{flair-multi}, \textit{i.e}~the multilingual Flair embeddings of size 4096 pre-trained on the JW300 corpus \citep{agic-vulic-2019-jw300} with more than 300
    languages;
    \item[-] \textit{BPEmb-multi}, \textit{i.e}~the multilingual Byte-pair encoding embeddings of size 300 trained on the 275 most common Wikipedia languages \citep{heinzerling2018bpemb}.
\end{itemize}

\begin{table*}[t]
\centering
\ra{1.2}
    \begin{tabular}{@{}l|rrrrrr@{}}
    \textbf{Name} & \textbf{Dim.} & \textbf{Level} & \textbf{Contextual} & \textbf{Lang} & \textbf{Training data} \\ 
    \midrule
    \textit{fastText-fr} & 300 & word &  & fr & CC \& Wikipedia \\
    \textit{flair-fr} & 4096 & char & \checkmark & fr & Wikipedia \\
    \textit{flair-multi} & 4096 & char & \checkmark & 300 lang & JW300 corpus \\
    \textit{BPEmb-multi} & 300 &  sub-word &  & 275 lang & Wikipedia \\
    \textit{fastText-luxwort} & 300 &  sub-word &  & 3 lang & \textsc{luxwort} \\
    \textit{flair-luxwort} & 4096 & char & \checkmark & 3 lang & \textsc{luxwort} \\
    \textit{fastText-flair-fr} & 4396 & word$+$char & - & fr & CC$+$Wiki (stack)\\ 
    \textit{BPEmb-flair-multi} & 4396 &  sub-word$+$char& - & multi & JW300$+$Wiki (stack)\\
    \textit{fastText-flair-luxwort} & 4396 &  sub-word$+$char & - & 3 lang & \textsc{lux}$+$Wiki (stack)\\
    \end{tabular}
    \caption{Overview of embeddings. The `-' sign means both contextual and non-contextual embeddings (stacks).
    \label{tbl:embedd}}
    \hrule height 0pt
\end{table*}

\bigskip
Next, in order to test the effect of in-domain embeddings, two models were trained on a corpus of 2GB of text of the \textit{Luxemburger Wort} for the period 1848-1950. The first is a FastText model trained on lowercase space-separated input, with at least 3 occurrences per token, a context windows of 8 tokens, a sub-word max character n-gram  length of 6, resulting in embeddings of size 300 (\textit{fastText-luxwort}). The second model is a Flair one trained on the raw OCR output for 96 hours on a NVIDIA Tesla V100, with a batch size of 600, a context length of 250 characters and a hidden size of 2048, resulting in embeddings of size 4096 (\textit{flair-luxwort}). 

Finally, different embedding stacks are considered, combining non-contextual (fastText or Byte-pair) with contextual embeddings (Flair). For experiments related to \textsc{jdg}, \textsc{gdl} and \textsc{imp}, a stack of \textit{fastText-fr} and \textit{flair-fr} is used: \textit{fastText-flair-fr}. For experiments related to \textsc{luxwort}, two different configurations are used: a) a combination of pre-trained embeddings \textit{BPEmb-multi} and \textit{Flair-multi}, referred to as \textit{BPEmb-flair-multi}, and b)  a combination of in-domain embeddings with \textit{fastText-luxwort} and \textit{flair-luxwort}, referred to as \textit{fastText-flair-luxwort}. 

All embeddings and stacks were tested during a pilot phase, and the stacks appeared to be the best in our context. They combine contextual and non-contextual information, as well as word and sub-word information, and seem therefore more suitable to cope with old language and OCR output. Experiments presented in Section\ref{sec:experiments} are based on the stack embeddings exclusively.

\subsection{Training and Post-processing}
\label{sec:training}
\vspace{1ex}

Text embedding maps are pre-computed for all images using the embedding stacks described above and the text associated with the images (original OCR for the Luxembourgish newspaper, \textsc{abbyy} one for the Swiss). Before training, images are resized to fit in $5\cdot10^5$ pixels and are augmented by random scaling ($s \in [0.8, 1.2]$) and rotation ($r \in [-0.01,0.01]rad$). All models are trained for 17,000 steps with a batch size of 4 and  batch renormalization \citep{ioffe2017batch}. We use Adam optimizer \citep{kingma2014adam} with an exponentially decaying learning rate of .95 starting at $10^{-4}$, and a weight regularization of $10^{-6}$. In order to prevent overfitting a development set containing 10\% of the training set is used. The final result for each model is reported on the weights where the loss on the development set was the lowest. Models are trained on a NVIDIA Tesla V100 GPU with 32GB of memory using the Tensorflow library\footnote{\href{https://www.tensorflow.org}{https://www.tensorflow.org}} 1.13.1. 

In terms of post-processing, the class mask is computed from the final output of the network, a pixel probability map. A pixel is considered as belonging to the background if it has probabilities smaller than 50\% for all classes, otherwise to the class with the highest probability. In order to avoid small masks, connected components with an area smaller than 5\% of the size of the image are discarded.

\subsection{Evaluation setup}
\label{sec:eval}
\vspace{1ex}

Given an image and a class, we wish to create a mask that contains pixels of the class. Figure \ref{fig:iou_comp} illustrates this procedure, with Figure \ref{fig:obituary_original} being the input and Figure \ref{fig:obituary_gt} the ground truth, the latter with the mask coloring each pixel according to its class (here yellow for death notices pixels). On this base, several metrics are used to evaluate the models.

\subsubsection{Metrics}
\vspace{1ex}

\begin{figure}[htb!]
\centering
\hspace*{\fill}
\subcaptionbox{Original image\label{fig:obituary_original}}{\includegraphics[width=0.2\textwidth]{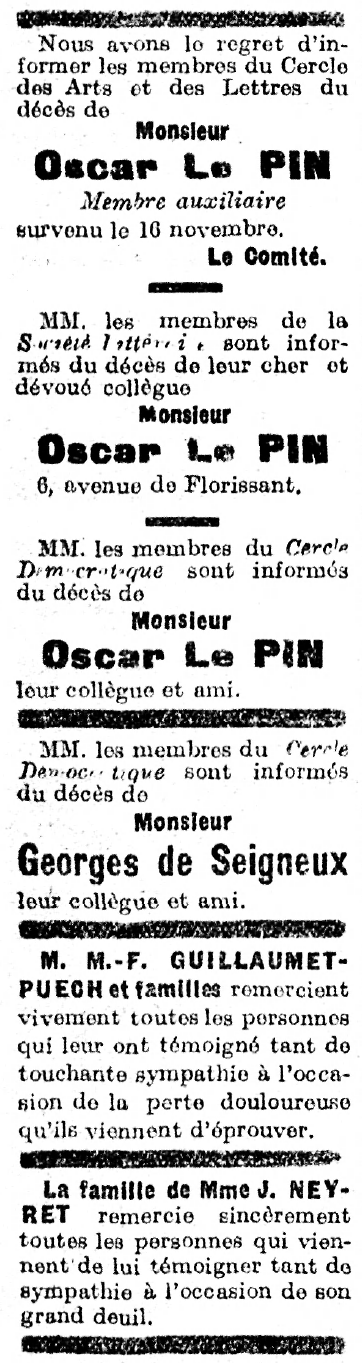}}\hfill%
\subcaptionbox{Ground truth\label{fig:obituary_gt}}{\includegraphics[width=0.2\textwidth]{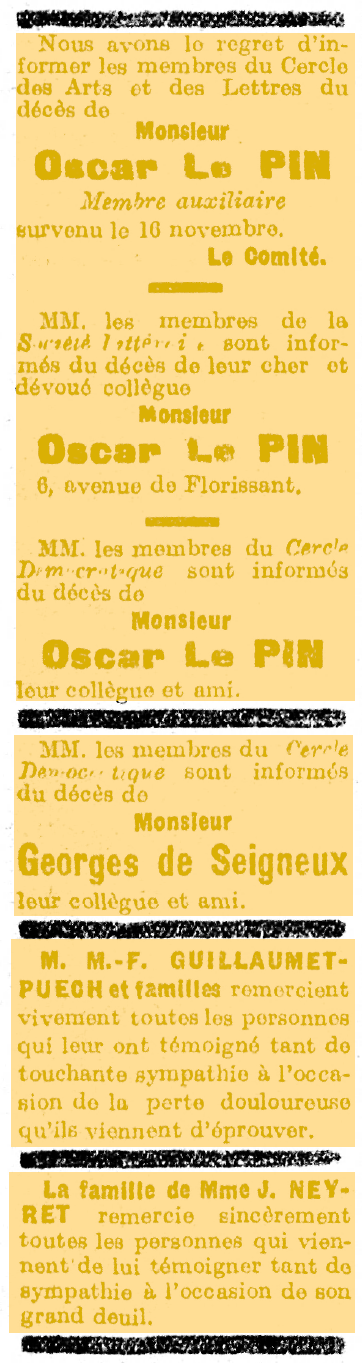}}\hfill%
\subcaptionbox{IoU = 0.23\label{fig:obituary_bad}}{\includegraphics[width=0.2\textwidth]{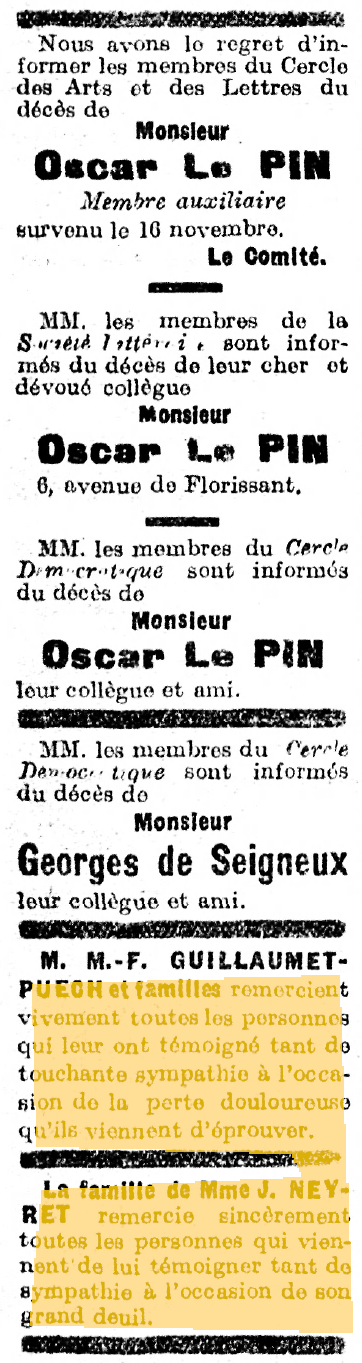}}\hfill%
\subcaptionbox{IoU = 0.86\label{fig:obituary_good}}{\includegraphics[width=0.2\textwidth]{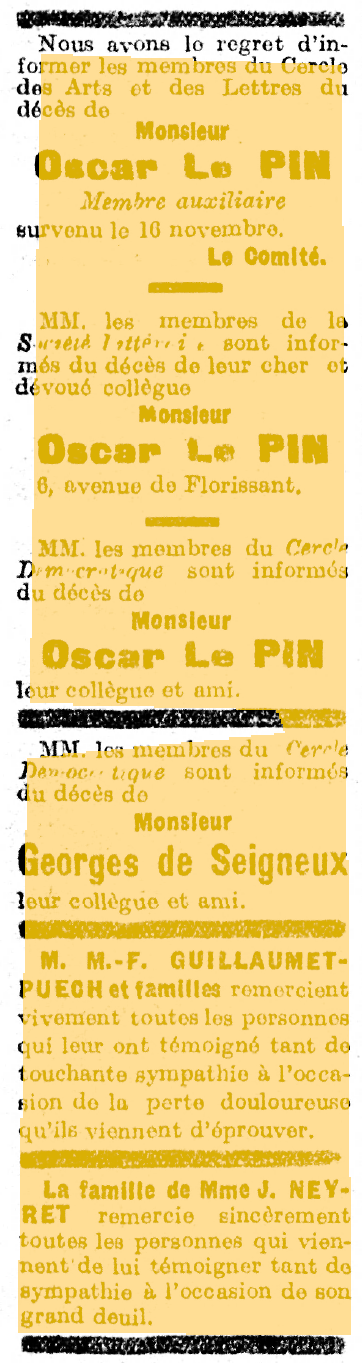}}%
\hspace*{\fill}
\captionsetup{width=0.9\textwidth}
\caption{Examples of the behaviour of the IoU metric.
\label{fig:iou_comp}
}
\end{figure}

\paragraph{Mean Intersection over Union} The Intersection over Union (IoU) is the standard metric for  semantic image segmentation and measures how well two sets of pixels are aligned. It is computed as follows. Given an image $i$ belonging to the set of images
$I$, a class $c$ belonging to the set of classes $C$, a set of predicted pixels $P_{ic}$ of image $i$ belonging to class $c$, and a set of ground-truth pixels $G_{ic}$ of image $i$ belonging to class $c$, the IoU for image $i$ and class $c$ is:

$$IoU_{ic} = \frac{|P_{ic}\cap G_{ic}|}{|P_{ic} \cup G_{ic}|}$$

Figure \ref{fig:iou_comp} shows a quantified example, where the prediction on images \ref{fig:obituary_bad} and \ref{fig:obituary_good} are compared to the ground truth of image \ref{fig:obituary_gt}. Image \ref{fig:obituary_bad} has a too small and misaligned prediction, and therefore a low IoU, while \ref{fig:obituary_good} is better. It is important to note that the metric is computed at the image (or pixel) level, and not the content item instance level. Indeed, although there are four distinct death notice instances in Figure \ref{fig:iou_comp}, the annotation makes no distinction and the model does not need to separate them to obtain a good score.

The mean Intersection over Union (mIoU) for a class $c$ over the set of Image $I$ corresponds to the average of the IoUs of all images where the union of the predicted and the ground-truth set
has at least a pixel of class $c$ (the true negatives are thus not counted). This can be more formally defined as all images $J = \lbrace i \in I \mid \vert P_{ic} \cup G_{ic}| >
0 \rbrace$. Then the mIoU is:

$$mIoU_c = \frac{1}{|J|}\sum_{j \in J}IoU_{jc}$$

\paragraph{Precision and Recall} The IoU does not qualify performances in terms of true positives (TP), true negatives (TN), false positives (FP) and false negatives (FN). However, those values are of interest when considering whether a model can be used in concrete terms, \textit{i.e.} if most of the segments are correctly recognized. The usual way to measure those values in segmentation is to consider an example as positive when above a certain threshold $\tau \in [0,1]$ of IoU. In this case, the prediction is well enough aligned with the ground truth to be considered as correct. On this base, it is possible to consider a prediction with an IoU $\geq \tau$ as a TP, a prediction with no IoU (i.e. with a union of zero) as a TN, a prediction with an IoU of zero and no predicted pixels (i.e. intersection of zero and non-zero number of pixel in the ground truth) as a FN and, finally, a non-FN prediction with an IoU $< \tau$ as a FP. Given a threshold $\tau$, it is therefore possible to compute precision and recall, as follows:
$$\text{Precision at $\tau$} = P@\tau = \frac{TP}{TP+FP}$$

$$\text{Recall at $\tau$} = R@\tau = \frac{TP}{TP+FN}$$

Finally, it is also possible to compute the average precision and recall over a range of thresholds. A range of threshold, is defined by a start $\tau_{start}$ an end $\tau_{end}$ and step size between two threshold $\tau_{step}$ using the following notation: $\tau_{start}\text{:}\tau_{step}\text{:}\tau_{end}$, for example a threshold between 50 and 95 with a step of 5 would be written as $50\text{:}5\text{:}95$. Given a threshold range the average metric $M$ (which can be precision, recall or anything else) is then computed as follows:

$$M@\tau_{start}\text{:}\tau_{step}\text{:}\tau_{end} = \frac{1}{|\tau_{start}\text{:}\tau_{step}\text{:}\tau_{end}|}\sum_{\tau \in \tau_{start}\text{:}\tau_{step}\text{:}\tau_{end}}M@\tau$$

Let us emphasize once again that these metrics (IoU, mIoU, P, R) are computed at the page level and not at the instance level. If a page contains several instances of a class and the prediction matches some instances, but not enough to reach an IoU threshold larger than $\tau$, the whole page is counted as negative.

\subsubsection{Reported results}
\vspace{1ex}

For each experiment, results are reported in terms of mIoU (as a percentage), and precision and recall with, respectively, a threshold of 60\% and 80\% and the average of threshold $50\text{:}5\text{:}95$ of the IoU.

Since models can have a high variance between runs, each model is trained ten times and the average and standard deviation of their performance are reported in the form of tables and boxplots. Even though most of our analyses are based on the mean, we indicate whether the difference of means between two models is significant by using Welch's $t$-test, following the recommendation of \cite{reimers2018comparing}. The significance is indicated using stars (*), where their numbers corresponds to a certain $p$-value: one star (*) indicates that $p \le 0.05$, two (**) that $p \le 0.01$, three (***) that $p \le 0.001$, and four (****) that $p \le 0.0001$.

\subsection{Material release}
\vspace{1ex}
Annotated material is released in the VIA format as open data (under different right statements according to the source on Zenodo) under DOI \href{https://doi.org/10.5281/zenodo.3706863}{10.5281/zenodo.3706863}.

The model architecture is thought as a plugin of \textit{dhSegment}, named \textit{dhSegment-text}. Two implementations are available. The first one, used for the present experiments and based on the TensorFlow implementation of \textit{dhSegment}, is available on GitHub\footnote{\href{https://github.com/dhlab-epfl/dhSegment-text}{https://github.com/dhlab-epfl/dhSegment-text}} and can be used for reproducibility purposes. The second one, based on the new pyTorch implementation of \textit{dhSegment},\footnote{\href{https://github.com/dhlab-epfl/dhSegment-torch}{https://github.com/dhlab-epfl/dhSegment-torch}} is also available on GitHub\footnote{\href{https://github.com/dhlab-epfl/dhSegment-text-torch}{https://github.com/dhlab-epfl/dhSegment-text-torch}} and can be used for training new models.

Finally, a selection of (best) trained models are released on the \textit{dhSegment-text} repository, under a CC BY-SA 4.0 license.

%% file: 05-experiments.tex
\section{Experiments}
\label{sec:experiments}
\vspace{1ex}

In this section, we motivate and present four series of experiments that address important questions with regard to the automatic recognition of fine-grained semantic segments in historic newspapers.  Since our material was published over a long period of time, we are specifically interested in the diachronic robustness of our models.

In Section \ref{sec:gen_exps}, we examine the predictive power of visual and textual features on our four classes, representative of different difficulties. In Section \ref{sec:gen_through_time}, we test the generalization ability of our multimodal approach with respect to (a) the changes over time in newspaper layout and content, and (b) the transfer of models from one newspaper to a related one, which has not been part of the training material.
In Section \ref{sec:assess_effects}, we examine whether textual features allow to reduce the amount of training data given that they add another source of signal to the models.
In Section \ref{sec:exp-increase-indomain}, we focus on multilingual death notices from a single newspaper and examine (a) how the increase of training material improves the results, and (b) how valuable in-domain text embedding are, meaning character and word embeddings that were specifically trained on the multilingual and noisy OCR source material from the very newspaper.

\subsection{Combining Visual and Textual Information}
\label{sec:gen_exps}
\vspace{1ex}

The first series of experiments addresses the following  questions: 
(a) How well does fine-grained semantic segmentation perform on our four selected classes under the condition that training and test data are sampled representatively from the same newspaper? (b) How strong is the signal contained in the textual embedding maps?  (c) What is the expected benefit of combining visual and textual features? 

\subsubsection{Experiment description}
\vspace{1ex}

Here, models are trained on long-term diachronic \textsc{jdg} data only in order to reserve \textsc{gdl} and \textsc{imp} datasets for generalization experiments (Section \ref{sec:gen_through_time}). The \textsc{jdg} dataset was randomly split to compose a training (1,387 images) and test (595 images) sets. Table \ref{tbl:distribution_classes} presents the class distribution, where it can be observed that class ratios are similar between the training and test sets. Given the homogeneity and representativity of the material, the results of this first series of experiments serve as an upper bound for our approach for diachronic, fine-grained image semantic segmentation.

\begin{table}[t]
\centering
\begin{tabular}{l|rrrr}
  \textbf{Class} & \textbf{Train size (ratio)} & \textbf{Test size (ratio)} \\
  \midrule
  Serial & 101\hphantom{0}  (7.28\%) &  36\hphantom{0} (6.05\%) \\ 
  Weather forecast & 103\hphantom{0} (7.43\%) &  53\hphantom{0} (8.91\%) \\ 
  Death notice & 107\hphantom{0} (7.71\%) &  46\hphantom{0} (7.73\%) \\ 
  Stock exchange table & 189 (13.63\%) &  86 (14.45\%) \\ 
  Pages w/o annotations & 982 (70.80\%) & 411 (69.08\%) \\ 
\end{tabular}
\caption{Distribution of the classes for the training and test sets. \label{tbl:distribution_classes}}
\end{table}

In order to measure the effectiveness of using visual features only, textual features only, or a combination of both, we experimented with three modalities:
\begin{enumerate}
    \item \textsf{Image}: the model receives as input a newspaper image (pixels) only and relies solely on visual features. It is equivalent to the model described in \citep{oliveira18_dhseg}.
    \item \textsf{Text}: the model receives as input a blank image and a text embedding map of a newspaper page and relies solely on textual features.
    \item \textsf{Image+Text}: the model receives as input a newspaper image and its corresponding text embedding map and combines visual and textual features.
\end{enumerate}

\bigskip
Each model with textual features uses the architecture described in Section \ref{sec:model}, where text and image information are fused early in the network, as well as the \textit{fastText-flair-fr} stack embeddings.

\begin{table}[htb!]
    \centering
 \begin{adjustbox}{max width=\textwidth}
\begin{tabular}{cc|rrrr|r}
\textbf{Metric} & \textbf{Modality} &       \textbf{Serial} &         \textbf{Weather} &        \textbf{Death Notice} &          \textbf{Stocks} &         \textbf{Average} \\
\midrule
mIoU & \textsf{Image} &   74.12$\pm$7.59 &  81.27$\pm$2.18 &  75.37$\pm$2.98 &  83.11$\pm$0.88 &   79.30$\pm$2.29 \\
     & \textsf{Text} &   49.05$\pm$9.41 &  73.55$\pm$3.55 &  71.44$\pm$3.26 &  78.87$\pm$2.61 &   69.30$\pm$2.25 \\
     & \textsf{Image+Text} &    76.73$\pm$5.90 &  81.38$\pm$3.34 &  ${}^{****}$\textbf{83.58}$\pm$2.02 &  84.43$\pm$1.84 &  ${}^{**}$\textbf{82.16}$\pm$1.72 \\
\midrule
P@60 & \textsf{Image} &   82.02$\pm$7.24 &  91.08$\pm$4.93 &  83.37$\pm$4.46 &  89.21$\pm$1.42 &   86.86$\pm$2.60 \\
     & \textsf{Text} &  53.97$\pm$12.42 &  82.29$\pm$5.13 &  82.19$\pm$5.42 &  86.85$\pm$2.96 &  77.27$\pm$3.03 \\
     & \textsf{Image+Text} &   83.24$\pm$7.17 &  91.81$\pm$4.67 &   ${}^{***}$\textbf{91.27}$\pm$2.80 &  90.11$\pm$1.77 &  ${}^{*}$\textbf{89.43}$\pm$1.79 \\
\midrule
P@80 & \textsf{Image} &  66.45$\pm$14.87 &  66.94$\pm$7.41 &  67.37$\pm$2.04 &  80.51$\pm$2.19 &  72.29$\pm$3.74 \\
     & \textsf{Text} &  29.95$\pm$23.47 &  58.13$\pm$3.97 &  62.01$\pm$5.34 &   74.07$\pm$3.90 &  57.93$\pm$5.73 \\
     & \textsf{Image+Text} &  71.54$\pm$15.05 &  71.37$\pm$7.28 &  ${}^{***}$\textbf{80.89}$\pm$3.88 &  ${}^{**}$\textbf{83.49}$\pm$2.11 &  ${}^{**}$\textbf{78.07}$\pm$3.76 \\
\midrule
P@50:5:95 & \textsf{Image} &  65.37$\pm$10.22 &  69.10$\pm$2.84 &  66.77$\pm$2.43 &  78.36$\pm$1.31 &  71.53$\pm$2.82 \\
   & \textsf{Text} &  36.97$\pm$14.21 &  60.02$\pm$3.46 &   59.78$\pm$3.70 &  70.99$\pm$3.32 &  58.46$\pm$2.96 \\
   & \textsf{Image + Text} &  68.12$\pm$10.47 &  70.98$\pm$3.81 &   ${}^{****}$\textbf{76.18}$\pm$2.10 &  79.38$\pm$2.55 &  ${}^{*}$\textbf{74.80}$\pm$2.49 \\
\midrule
R@60 & \textsf{Image} &   97.91$\pm$1.95 &  78.74$\pm$1.98 &  93.02$\pm$2.88 &  93.78$\pm$0.73 &  90.64$\pm$1.03 \\
     & \textsf{Text} &    ${}^{**}$\textbf{100.00}$\pm$0.00 &  ${}^{****}$\textbf{90.57}$\pm$3.64 &  88.07$\pm$2.91 &  91.71$\pm$1.23 &  91.81$\pm$1.82 \\
     & \textsf{Image+Text} &    ${}^{**}$\textbf{100.00}$\pm$0.00 &  ${}^{****}$\textbf{87.14}$\pm$2.75 &  90.37$\pm$1.44 &  93.73$\pm$1.08 &  ${}^{***}$\textbf{92.39}$\pm$0.95 \\
\midrule
R@80 & \textsf{Image} &   97.52$\pm$2.26 &  73.06$\pm$2.57 &  91.49$\pm$3.67 &  93.15$\pm$0.81 &   88.94$\pm$1.40 \\
     & \textsf{Text} &    ${}^{**}$\textbf{100.00}$\pm$0.00 &  ${}^{****}$\textbf{87.18}$\pm$4.82 &  84.72$\pm$4.15 &   90.42$\pm$1.30 &  89.26$\pm$2.78 \\
     & \textsf{Image+Text} &    ${}^{**}$\textbf{100.00}$\pm$0.00 &   ${}^{****}$\textbf{84.00}$\pm$3.32 &  89.27$\pm$1.48 &  93.27$\pm$1.14 &   ${}^{***}$\textbf{91.37}$\pm$1.00 \\
\midrule
R@50:5:95 & \textsf{Image} &   95.38$\pm$3.72 &  67.87$\pm$2.38 &  88.30$\pm$4.73 &  92.53$\pm$0.77 &   87.13$\pm$1.40 \\
   & \textsf{Text} &   85.00$\pm$10.8 &  ${}^{****}$\textbf{79.09}$\pm$3.72 &   75.16$\pm$4.00 &  87.42$\pm$1.73 &  84.89$\pm$3.15 \\
   & \textsf{Image} + Text &   96.00$\pm$5.16 &  ${}^{****}$\textbf{77.68}$\pm$3.64 &  85.05$\pm$2.63 &  92.42$\pm$1.27 &  ${}^{**}$\textbf{89.35}$\pm$1.24 \\
\end{tabular}
\end{adjustbox}
\caption{Results of the first series of experiments reported as mean values $\pm$ standard deviation computed from 10 runs. Stars indicate statistically significant improvements from \textsf{Text} and \textsf{Image+Text}  relative to \textsf{Image}.}
\label{tbl:general_results}
\end{table}

 \begin{figure}[tb]
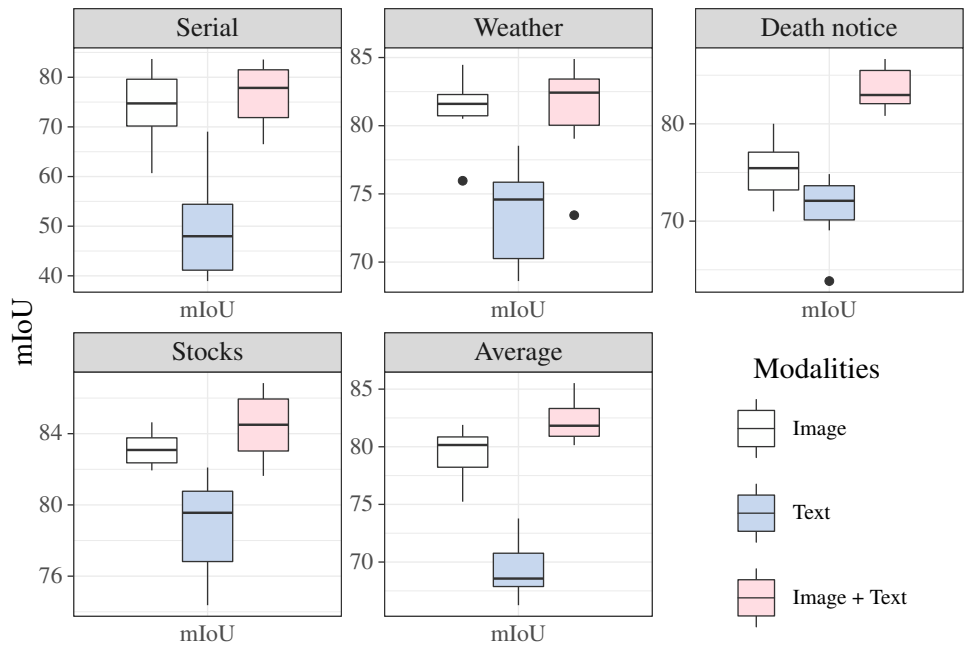

  \makebox[\textwidth][c]{%
  \includestandalone[width=.8\textwidth]{Figures/general_miou}
\  }
    \captionsetup{width=\linewidth}
    \caption{Box plots of the mIoU of the first series of experiments.}
    \label{fig:general_miou}
  \end{figure}
  
Results are shown in Table \ref{tbl:general_results} and  Figure \ref{fig:general_miou}. In general, the \textsf{Image} model outperforms the pure \textsf{Text} model by a large margin in terms of mIoU and precision. Except for recall-oriented setups, there is no advantage of restricting the models to textual features only. As expected, \textsf{Image} model is stronger on classes that are more visually distinct (\textit{Weather} and \textit{Stocks}) than on classes that are mainly text based (\textit{Serial} and \textit{Death notice}).

With respect to the class average\footnote{In all experiments, average corresponds to micro-average.} results, \textsf{Image+Text} models perform significantly better than \textsf{Image} for every metric, attesting  a real gain in using the combination of visual and textual features for the task. The better precision of \textsf{Image} and the good recall of \textsf{Text} play well together, leading also to less variance across models, as the smaller standard deviations indicate.
For all modalities, there is a big drop in precision when augmenting the IoU threshold. This indicates that it is hard to be precise about the location of a segment and that the \textsf{Image+Text} model is more robust than the single modality models.

\begin{figure}[htb!]
\centering
    \begin{minipage}[b]{.45\textwidth }
    \includegraphics[width=1\textwidth]{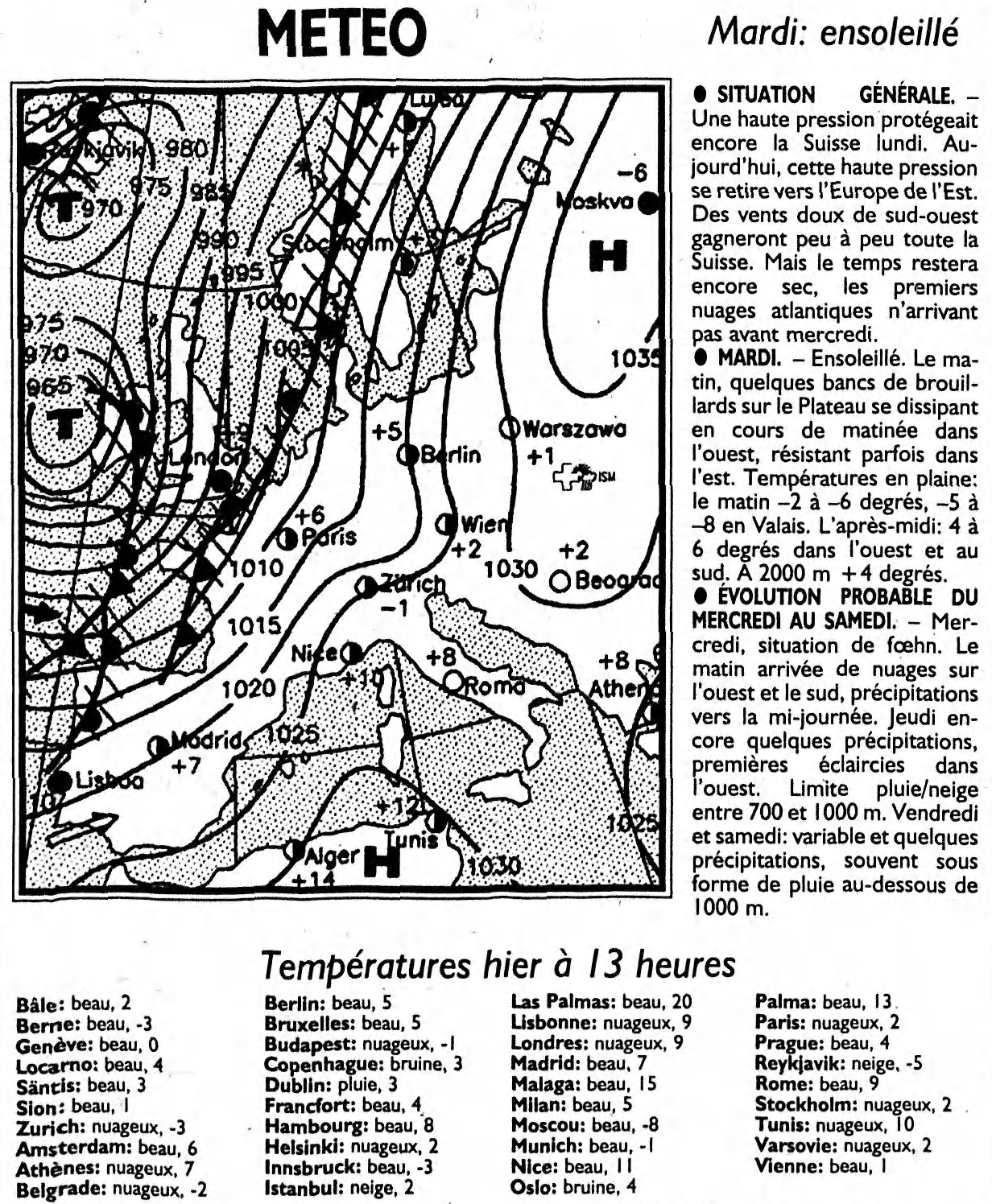}
    \end{minipage}%
\hspace{1cm}
\begin{minipage}[b]{.45\textwidth }
    \includegraphics[width=\textwidth]{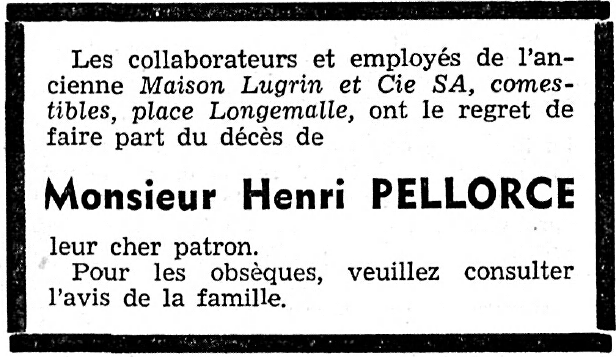}
    \includegraphics[width=\textwidth]{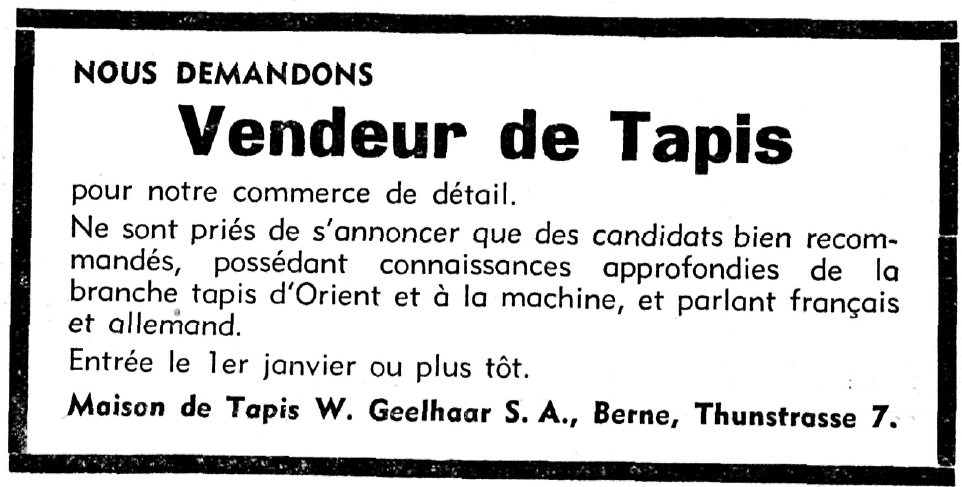}
    \end{minipage}\\
    \hspace*{\fill}
    \begin{minipage}[t]{.45\linewidth}
        \caption{A Weather forecast with both visual and textual features. The \textsf{Image + Text} model finds both the text and the image, whereas the \textsf{Image} model only finds the map and the table.\label{fig:mixed_weather}}
    \end{minipage}%
\hspace{1cm}
        \begin{minipage}[t]{.45\linewidth}
        \captionsetup{width=1\linewidth}
        \caption{A death notice (top) and an advertisement (bottom) with similar layouts, but very different textual features. The \textsf{Text} model correctly  detects only the top example, whereas the \textsf{Image} model is misled by the advertisement.\label{fig:diff_obituary}}
    \end{minipage}%
    \hspace*{\fill}
\end{figure}

The recall of \textit{Weather} is better for both models using textual features. This means that these features are essential for the retrieval of the class \textit{Weather}. As illustrated in Figure \ref{fig:mixed_weather},  weather reports may contain images, maps, and text. While the first two types are visually distinct, vocabulary might be the only semantically distinct feature for purely textual weather reports.

The mIoU and precision of \textit{Death notice} is significantly higher for the \textsf{Image+Text} model than any single modality model. In particular, the gains versus the \textsf{Image} model are important for the for the mIoU (+5.8\%-10.\% at 95\% confidence), for the P@60 (+4.4\%-11.4\% at 95\% confidence), for the P@80 (+10.5\%-16.5\% at 95\% confidence) and for the P@50:5:95 (+7.3\%-11.6\% at 95\% confidence). This strong increase in precision shows that the \textsf{Image+Text} model is much more robust against false positives than the \textsf{Image} one. As illustrated in Figure \ref{fig:diff_obituary}, advertisements can have similar layout, but very different textual content.

The absence of significant differences in terms of mIoU and precision between the \textsf{Image} and \textsf{Image+Text} approaches for \textit{Serial} can be explained by the lack of strong characteristics in either of the modalities. Visually, serials look similar to the rest of the newspaper. Textually, their vocabulary does not differ much from the rest of the newspaper either. However, the fact that it reaches a higher recall score for both models using textual features indicates that they are important for retrieval. The lower precision of the \textsf{Text} model shows that these features are also present in other articles.

Finally, the similar results between all approaches for \textit{Stocks} show that the visual and textual signals are both strong enough to detect this class. The reason for the slightly lower score of the \textsf{Text} model could be crucially missing visual information about purely visual elements such as the lines of a table.

\subsubsection{Summary}
\vspace{1ex}

The  first series of experiments assesses a consistent gain in performance by combining visual and textual features. The gain is particularly strong with content items as \textit{Death Notices} that exhibit easily confusable visual features, but have distinct textual features. We also observe a better recall for \textit{Weather Reports}, that consists of a mix of visual and textual elements. Even though the \textsf{Image+Text} model does not improve much on purely textual classes such as \textit{Serial} or visually distinct classes such as \textit{Stocks}, it still performed at least as well as a model using only the image. 

\subsection{Generalizing Through Time and Across Newspapers}
\label{sec:gen_through_time}
\vspace{1ex}

The second series of experiments addresses the following  questions: 
(a) Do models trained on textual and visual features perform better than purely visual models when applied to an unseen time period of the same newspaper?
(b)  Do models trained on textual and visual features perform better than purely visual models when applied to a related newspaper, where no issue was part of the training material?

\subsubsection{Experiment Description}
\vspace{1ex}

The first experiment on generalization \textit{through time} uses the \textsc{jdg} dataset, with material from the periods 1826-1968 and 1992-1998 as training data (1,394 pages), and 1969-1991 as test data (588 pages) Note that the test period has a different layout than the other periods \citep{buntinx2017layout}.
The second experiment on generalization \textit{across newspapers} trains on the same training set as the first series of experiments (Section \ref{sec:gen_exps}), but tests on the data from \textsc{gdl} (1,008 pages) and \textsc{imp} (1,634 pages). Both experiments compare the generalization ability of the \textsf{Image} and \textsf{Image+Text} models by testing them on layouts  never  seen before.  This setting is therefore more challenging than the previous one where the training set was  sampled uniformly over time and representative of the test set. 

\begin{table*}[hbt]
\centering
\ra{1}
\begin{adjustbox}{max width=\textwidth}
\begin{tabular}{@{}l|r|rr|rr@{}}
\textbf{Class} & \textbf{\textsc{jdg} train} & \textbf{\textsc{jdg} Time train}  & \textbf{\textsc{jdg} Time test} & \textbf{\textsc{gdl} test}  &    \textbf{\textsc{imp} test} \\
\midrule
Serial  &        101\hphantom{0} (7.28\%)&              134\hphantom{0} (9.61\%)&      3\hphantom{0} (0.51\%) &                  108 (10.71\%) &      103\hphantom{0} (6.30\%) \\
Weather       &        103\hphantom{0} (7.43\%) &              132\hphantom{0} (9.47\%) &      24\hphantom{0} (4.08\%) &                  68\hphantom{0} (6.75\%) &      41\hphantom{0} (2.51\%) \\
Death notice        &          107\hphantom{0} (7.71\%) &               124\hphantom{0} (8.90\%) &       29\hphantom{0} (4.93\%) &                   69\hphantom{0} (6.85\%) &      102\hphantom{0} (6.24\%) \\
Stocks        &        189 (13.63\%) &               211 (15.14\%)&       64 (10.88\%)&                  135 (13.39\%)&      79\hphantom{0} (4.83\%)\\
Pages w/o annotations        &        982 (70.80\%) &               923 (66.21\%) &      470 (79.93\%) &                   697 (69.15\%) &     1326 (81.15\%) \\
\end{tabular}
\end{adjustbox}
\caption{Distribution of the classes for the different datasets 
\label{tbl:distrib_contstraints}}
\end{table*}

Distributions of classes for each dataset are shown in Table \ref{tbl:distrib_contstraints}. In general, the distribution between the original \textsc{jdg} dataset and the other one changes. The closest dataset is \textsc{gdl}, which is not surprising since both newspapers come from neighbouring cities.
The largest difference is between the two time periods, with a much lower ratio of content items of the four classes for the test period.
The same embeddings as  in Section \ref{sec:gen_exps} are used, that is to say \textit{Fasttext-Flair-fr} (c.f. Section \ref{sec:embeddings} and Table \ref{tbl:embedd}). 

\subsubsection{Results and Discussion}
\vspace{1ex}

\begin{table}[htb]
  \small
  \centering
  \setlength{\tabcolsep}{2pt}
 \begin{adjustbox}{max width=\textwidth}
\begin{tabular}{cc|rrrr|r}
\textbf{Exp.} & \textbf{Modalities} &  \textbf{Serial} &         \textbf{Weather} &        \textbf{Death Notice} & \textbf{Stocks} & \textbf{Average}\\ 
  \midrule
Time & \textsf{Image}  &    8.00$\pm$2.66 &   29.44$\pm$\hphantom{0}6.28 &  51.29$\pm$12.88 &   ${}^{**}$\textbf{68.30}$\pm$3.31 &  54.65$\pm$5.47 \\
Time & \textsf{Image+Text}  &  ${}^{****}$\textbf{25.08}$\pm$7.37 &  ${}^{****}$\textbf{60.65}$\pm$10.27 &   ${}^{****}$\textbf{77.52}$\pm$\hphantom{0}4.41 &  60.17$\pm$7.42 &  ${}^{**}$\textbf{62.84}$\pm$6.37 \\
  \specialrule{0.10pt}{4pt}{4pt}
  \textsc{gdl} & \textsf{Image}  &  67.79$\pm$6.62 &    58.60$\pm$\hphantom{0}3.00 &  63.06$\pm$\hphantom{0}3.26 &  72.38$\pm$2.35 &  67.59$\pm$3.12 \\
\textsc{gdl} & \textsf{Image+Text}  &  ${}^{*}$\textbf{73.81}$\pm$4.08 &  59.16$\pm$\hphantom{0}2.22 &  ${}^{****}$\textbf{75.32}$\pm$\hphantom{0}1.69 &  72.65$\pm$1.77 &  ${}^{**}$\textbf{71.54}$\pm$1.21 \\
  \midrule
  \textsc{imp} & \textsf{Image}  &  42.45$\pm$7.86 &   7.04$\pm$\hphantom{0}4.92 &  40.14$\pm$\hphantom{0}3.81 &  42.45$\pm$2.08 &  40.71$\pm$2.82 \\
\textsc{imp} & \textsf{Image+Text}  &   ${}^{***}$\textbf{56.70}$\pm$4.23 &  ${}^{****}$\textbf{17.53}$\pm$\hphantom{0}4.11 &  ${}^{****}$\textbf{67.36}$\pm$\hphantom{0}4.43 &  ${}^{***}$\textbf{49.46}$\pm$3.84 &  ${}^{****}$\textbf{54.97}$\pm$3.25 \\
\end{tabular}
\end{adjustbox}
\caption{Results for the mIoU metric. Mean metric $\pm$ standard deviation of the metric (in \%). Stars show statistical difference of mean between modalities.}
\label{tbl:time_newspapers_miou}
\end{table}

 \begin{figure}[htb!]
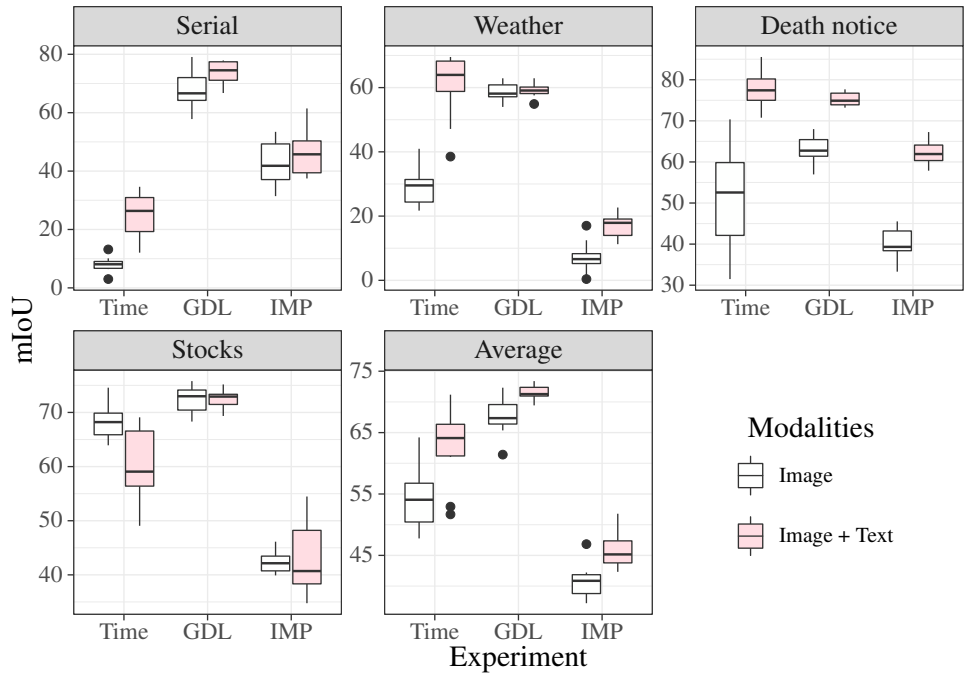

  \makebox[\textwidth][c]{%
  \includestandalone[width=0.8\textwidth]{Figures/generalization}
  }
    \captionsetup{width=\linewidth}
    \caption{Box plots of the mIoU of the generalization experiments.}
    \label{fig:generalization_exps}
  \end{figure}

Results are shown in Table \ref{tbl:time_newspapers_miou} and Figure \ref{fig:generalization_exps}. Compared to the results of the first series of experiments (cf. Section \ref{sec:gen_exps}), the performance is substantially lower. In general, poorer results mean that the examples in the training and test sets are too different. However, it should be noted that all the models using visual and textual features are significantly better than the \textsf{Image} models.

When focusing on the time constraint, it is clear that the \textsf{Image+Text} models perform significantly better for every class, except for \textit{Stocks}. The \textsf{Image} results show that death notices and stocks are visually more stable than serials and weather reports. However, the significant differences between the two models for the classes \textit{Weather} (+23\%- 39\% at 95\% confidence) and \textit{Death Notice} (+16\%-35\% at 95\% confidence) show that textual features are even more stable for these two classes. The poor results with \textit{Serial} reveal that this class is neither visually nor textually stable over time.  Finally, the results of \textit{Stocks} suggest that the visual features are more stable than the textual ones.

When focusing on the model transferability to other newspapers, the overall performance drop is much less pronounced with \textsc{gdl} than with \textsc{imp}, confirming that \textsc{jdg} and \textsc{gdl} have more common features in terms of layout. In particular, the low score of both models for the class \textit{Weather} in  \textsc{imp} dataset shows that this type of segment has great variability in terms of layout. For \textsc{imp}, the gain is, once again, particularly significant for \textit{Death Notice} (+23\%-31\% at 95\% confidence), demonstrating that textual features are particularly good at generalizing for this class.

\subsubsection{Summary}
\vspace{1ex}

The overall performance drop between these experiments and the ones in Section \ref{sec:gen_exps} confirms the variety of newspaper elements, both through time and across newspapers, and stresses the importance of  annotated data representatively sampled across time and newspaper. However, it also demonstrates that model generalization and transferability can be improved by the inclusion of textual features.

When considering the scores, most of them are too low to be considered in any practical use case, except maybe for obituaries. This indicates that even though this method shows some promises in terms of raw performance and generalization, it is still too early to use it at large scale without representative annotated data.

\subsection{Reducing Training Size}
\label{sec:assess_effects}
\vspace{1ex}

The third series of experiments addresses the following question: Do models combining visual and textual features need less training material? 

\subsubsection{Experiment Description}
\vspace{1ex}

These experiments assess the effect of reducing the training size by 60\%. The new training size is of 792 training samples againts 1,387 in the first experiments of Section \ref{sec:gen_exps}. The distribution of the re-sampled datasets can be found in Table \ref{tbl:jdg_less}. Once again it can be seen that the ratios of each class are similar between training and testing sets, and also w.r.t the experiments that used 100\% of the training data.

\begin{table}[ht]
\centering
\begin{tabular}{l|rrrr}
\textbf{Class} & \textbf{Train size (ratio)} & \textbf{Test size (ratio)} \\ 
  \hline
Serial &  56\hphantom{0} (7.07\%) &  81\hphantom{0} (6.81\%)\\ 
  Weather forecast &  61\hphantom{0} (7.70\%) &  95\hphantom{0} (7.98\%)\\ 
  Death Notice &  59\hphantom{0} (7.45\%) &  94\hphantom{0} (7.90\%) \\ 
  Stock Exchange Table &  92 (11.62\%) & 183 (15.38\%) \\ 
  Pages w/o annotations & 578 (72.98\%) & 815 (68.49\%) \\ 
\end{tabular}
\caption{Distribution of the classes for the training and testing sets. \label{tbl:jdg_less}}
\end{table}

\subsubsection{Results and Discussion}
\vspace{1ex}

\begin{figure}[htb]
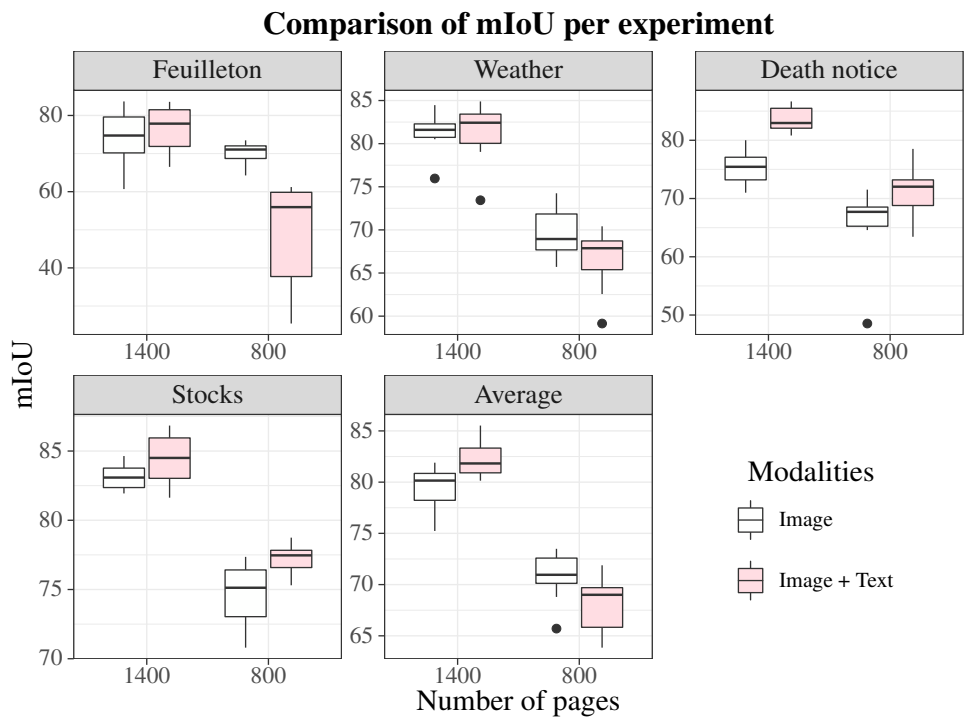

  \makebox[\textwidth][c]{%
  \includestandalone[width=0.8\textwidth]{Figures/JDG_data}
  }
    \captionsetup{width=\linewidth}
    \caption{Box plots of the miou of the \textsc{jdg} with 60\% training data.}
    \label{fig:jdg_less}
\end{figure}

Results are presented in Table \ref{tbl:jdg_less_results} and in Figure \ref{fig:jdg_less}.

\begin{table}[ht]
\centering
\begin{adjustbox}{max width=\textwidth}
\begin{tabular}{cc|rrrr|r}
\textbf{Modalities}             & \textbf{\# pages} &       \textbf{Serial} &         \textbf{Weather} &       \textbf{Death Notice} &          \textbf{Stocks} &         \textbf{Average} \\
\midrule
\textsf{Image}  & 1387 &   74.12$\pm$\hphantom{0}7.59 &  81.27$\pm$2.18 &  75.37$\pm$2.98 &  83.11$\pm$0.88 &   79.30$\pm$2.29 \\
\textsf{Image+Text}  & 1387 &    76.73$\pm$\hphantom{0}5.90 &  81.38$\pm$3.34 &  83.58$\pm$2.02 &  84.43$\pm$1.84 &  82.16$\pm$1.72 \\
\midrule
\textsf{Image}  & 792 &   ${}^{**}$\textbf{70.27}$\pm$\hphantom{0}2.79 &  69.67$\pm$2.76 &  65.88$\pm$6.48 &  74.73$\pm$2.25 &   ${}^{*}$\textbf{70.80}$\pm$2.32 \\
\textsf{Image+Text}  & 792 &  49.36$\pm$14.22 &  66.71$\pm$3.57 &   ${}^{*}$\textbf{71.20}$\pm$4.03 &   ${}^{**}$\textbf{77.30}$\pm$1.07 &  68.22$\pm$2.84 \\
\end{tabular}
\end{adjustbox}
\caption{Results for the mIoU metric. Mean metric $\pm$ standard deviation of the metric (in \%). Stars indicate statistical difference of mean with Image. It is only reported for models using 60\% of the data. 
\label{tbl:jdg_less_results}}
\end{table}

Overall the performance of the models with less data is inferior to the ones using 100\% of the training set. Still considering the average, the drop in mIoU is higher for the \textsf{Image+Text} model than for the \textsf{Image} model. However, this is mainly due to the poor performance of the former on \textit{Serial}. Indeed, the \textsf{Image} model performs significantly better (+10\%-31\% at 95\% confidence) than the \textsf{Image+Text} model. This may be due to the fact that \textsf{Image+Text} has not enough data to learn the textual features of \textit{Serial} and is thus more confused.

Regarding \textit{Death Notice} and \textit{Stocks}, the \textsf{Image+Text} model improves over the \textsf{Image} model. This may indicate that the textual features of these two classes are easier to learn than for \textit{Serial}, and that the model using both text and image features leverages better the small amount of training data for these classes by combining the two signals.

Finally, the fact that \textit{Weather} results have no significant difference between the two models may show that textual features are not as complex to learn as they are for \textit{Serial}, but do not influence much the results.

\subsubsection{Summary}
\vspace{1ex}

This experiment shows that not all classes are equal when the training size is reduced. It indicates that for content items with non focused textual content, such as \textit{Serial}, \textsf{Image+Text} requires more data to efficiently combine both signals. In contrast, it suggests that for more domain specific content items, such as \textit{Stocks} and \textit{Death Notice}, \textsf{Image+Text} model easily leverages the additional signal provided by the textual information.


\subsection{Assessing the Benefits of In-domain Embeddings}
\label{sec:exp-increase-indomain}
\vspace{1ex}

The fourth series of experiments addresses the following three questions: (a)  How big is the advantage if we train in-domain textual embeddings instead of using off-the-shelf embeddings trained on contemporary text data (without noisy OCR)? (b) Can adding more training data compensate for the expected benefit of in-domain embeddings? (c) What is the impact of adding more training material on the performance of the models and is it possible to identify the point at which adding more data becomes ineffective, i.e. a plateau is reached?

\subsubsection{Experiment Description}
\vspace{1ex}

These experiments make uses of several training sets with different numbers of pages, while the test set is kept constant. Training sets of different sizes are iteratively built starting from biggest to smallest by sampling, at each iteration, half of the number of issues per year: the first dataset has 26 issues per year, the next one 13 (therefore a subset of the previous one), the next 6, and so on. Training set statistics are shown in Table \ref{tbl:luxwort_exp_classes}.  The embeddings used are the \textit{BPEmb-flair-multi} and the \textit{fastText-flair-luxwort} stacks (cf. Section \ref{sec:embeddings}). Each experiment is thus characterized by its amount of training data and the embeddings used (or lack of it).

This experiment uses the newspaper \textit{Luxemburger Wort} and focuses on the \textit{Death Notice} class only since, as seen in previous experiments, it is the one that benefits most from the addition of textual features.

\begin{table}[ht]
\centering
\begin{adjustbox}{max width=\textwidth}
\begin{tabular}{r|rrrrr}
  \textbf{Dataset} &  \textbf{\# issues/year} & \textbf{\# issues} & \textbf{\# pages} & \textbf{\# death notices (ratio)}  \\ 
  \midrule
  Training pages \hfill  \textit{550} & 1 & 103 & 562 & 49 (8.72\%)\\
  \textit{1100} & 2 & 206 & 1098 & 100 (9.11\%)\\
  \textit{1650} & 3 & 309 & 1642 & 155 (9.44\%)\\
  \textit{3300} & 6 & 618 & 3290 & 315 (9.57\%)\\
   \textit{7000} & 13 & 1339 & 7029 & 683 (9.72\%)\\
  \textit{14100} & 26 & 2678 & 14128 & 1344 (9.51\%)\\
  \midrule
  \textit{test set} & 8 & 824 & 4825 & 421 (8.73\%)\\
 \end{tabular}
\end{adjustbox}
\caption{Distribution of the classes for the training and test sets. \label{tbl:luxwort_exp_classes}}
\end{table}

\begin{figure}[htb]
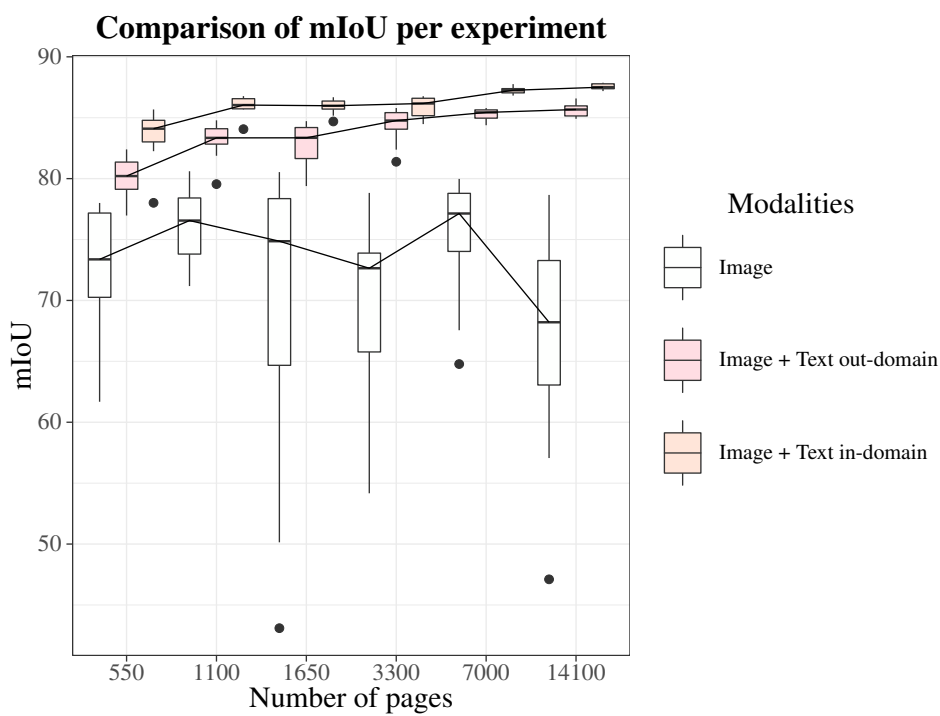

  \makebox[\textwidth][c]{%
  \includestandalone[width=.8\textwidth]{Figures/luxwort}
  }
    \captionsetup{width=\linewidth}
    \caption{Box plots of the mIoU of the Luxwort experiment.}
    \label{fig:luxwort_pages}
  \end{figure}

The results are presented in Table \ref{tbl:luxwort_miou} and  Figure \ref{fig:luxwort_pages}. Overall the results show that there is a significant gain in using textual embeddings maps, both in terms of performance and variance.
\begin{table}[ht]
\centering
\begin{adjustbox}{max width=\textwidth}
\begin{tabular}{r|rrr}
\textbf{\# pages} & \textbf{\textsf{Image}}  &     \textbf{\textsf{Image+Text out-domain}} & \textbf{\textsf{Image+Text in-domain}}  \\
\midrule
550    &   72.77$\pm$\hphantom{0}5.11 &  ${}^{**}$80.06$\pm$1.75 &      ${}^{**}$\textbf{83.49}$\pm$2.18 \\
1100   &    76.20$\pm$\hphantom{0}3.05 &  ${}^{****}$83.12$\pm$1.53 &      ${}^{***}$\textbf{85.99}$\pm$0.79 \\
1650   &  69.31$\pm$13.17 &  ${}^{**}$82.86$\pm$1.77 &      ${}^{***}$\textbf{85.93}$\pm$0.64 \\
3300   &   69.43$\pm$\hphantom{0}7.94 &  ${}^{***}$84.38$\pm$1.45 &      ${}^{*}$\textbf{85.88}$\pm$0.91 \\
7000   &    75.20$\pm$\hphantom{0}5.19 &  ${}^{***}$85.28$\pm$0.46 &      ${}^{****}$\textbf{87.24}$\pm$0.31 \\
14100  &   66.85$\pm$\hphantom{0}9.48 &  ${}^{**}$85.63$\pm$0.54 &      ${}^{****}$\textbf{87.55}$\pm$0.24 \\
\end{tabular}
\end{adjustbox}
\caption{Results for the mIoU metric. Mean metric 
  $\pm$ standard deviation of the metric (in \%). The stars of the \textsf{Image+Text out-domain} column indicate the statistical difference of mean w.r.t to \textsf{Image}, and the ones of \textsf{Image+Text in-domain} the difference w.r.t \textsf{Image+Text out-domain}.}
\label{tbl:luxwort_miou}
\end{table}

The \textsf{Image+Text out-domain} and \textsf{Image+Text in-domain} models always significantly beat the \textsf{Image} model. Moreover, the performance of those models increases and the variance decreases with the size of the training set. However, the performance difference between 550 pages  and 14,100 pages is only around 5\%, even though more than 25 times as many annotated pages are used.

In-domain text embeddings are  beneficial since they provide a consistent gain (+2-3\%) in performance over the out-domain text embeddings. Moreover \textsf{Image+Text in-domain} already surpasses with 1100 examples the performance that \textsf{Image+Text out-domain} achieves with 14100 examples. This is certainly due to the fact that the in-domain embeddings have been trained on enough data to capture the particular semantics of newspapers and the OCR errors present in them.

The results of the model using only the image are quite surprising since the performance decrease with the amount of training data which is counter-intuitive. In order to eliminate the hypothesis that a visual model that uses more data needs more steps for converging, the model with 14,100 pages was trained for the double of training steps, however, it only improved the results by 5\%, while still being 5\% lower than the model with 1,100 pages and having three times its variance.

\subsubsection{Summary}
\vspace{1ex}

In this experiment, the usage of textual features brings a gain in performances, reduces the variance of the model which converges better even with large amount of data. However, when considering the scores, the addition of training data has little impact and this extra annotation does not seem worth the effort. The use of in-domain embeddings shows a decent improvement that cannot be compensated by more training samples over out-domain embeddings while not requiring additional annotations. Indeed, training in-domain embeddings is done completely unsupervised, making it a worthwhile option if the amount of available text data is sufficiently large.

%% file: Figures/general_miou.tex
\begin{tikzpicture}[x=1pt,y=1pt]
\definecolor{fillColor}{RGB}{255,255,255}
\path[use as bounding box,fill=fillColor,fill opacity=0.00] (0,0) rectangle (433.62,325.21);
\begin{scope}
\path[clip] (  0.00,  0.00) rectangle (433.62,325.21);
\definecolor{drawColor}{RGB}{255,255,255}
\definecolor{fillColor}{RGB}{255,255,255}

\path[draw=drawColor,line width= 0.6pt,line join=round,line cap=round,fill=fillColor] (  0.00,  0.00) rectangle (433.62,325.21);
\end{scope}
\begin{scope}
\path[clip] ( 31.71,176.25) rectangle (151.02,285.97);
\definecolor{fillColor}{RGB}{255,255,255}

\path[fill=fillColor] ( 31.71,176.25) rectangle (151.02,285.97);
\definecolor{drawColor}{gray}{0.92}

\path[draw=drawColor,line width= 0.3pt,line join=round] ( 31.71,194.74) --
	(151.02,194.74);

\path[draw=drawColor,line width= 0.3pt,line join=round] ( 31.71,217.04) --
	(151.02,217.04);

\path[draw=drawColor,line width= 0.3pt,line join=round] ( 31.71,239.33) --
	(151.02,239.33);

\path[draw=drawColor,line width= 0.3pt,line join=round] ( 31.71,261.63) --
	(151.02,261.63);

\path[draw=drawColor,line width= 0.3pt,line join=round] ( 31.71,283.92) --
	(151.02,283.92);

\path[draw=drawColor,line width= 0.6pt,line join=round] ( 31.71,183.59) --
	(151.02,183.59);

\path[draw=drawColor,line width= 0.6pt,line join=round] ( 31.71,205.89) --
	(151.02,205.89);

\path[draw=drawColor,line width= 0.6pt,line join=round] ( 31.71,228.18) --
	(151.02,228.18);

\path[draw=drawColor,line width= 0.6pt,line join=round] ( 31.71,250.48) --
	(151.02,250.48);

\path[draw=drawColor,line width= 0.6pt,line join=round] ( 31.71,272.77) --
	(151.02,272.77);

\path[draw=drawColor,line width= 0.6pt,line join=round] ( 91.36,176.25) --
	( 91.36,285.97);
\definecolor{drawColor}{gray}{0.20}

\path[draw=drawColor,line width= 0.6pt,line join=round] ( 66.51,271.89) -- ( 66.51,280.98);

\path[draw=drawColor,line width= 0.6pt,line join=round] ( 66.51,250.87) -- ( 66.51,229.71);
\definecolor{fillColor}{RGB}{254,255,254}

\path[draw=drawColor,line width= 0.6pt,line join=round,line cap=round,fill=fillColor] ( 55.32,271.89) --
	( 55.32,250.87) --
	( 77.69,250.87) --
	( 77.69,271.89) --
	( 55.32,271.89) --
	cycle;

\path[draw=drawColor,line width= 1.1pt,line join=round] ( 55.32,261.03) -- ( 77.69,261.03);

\path[draw=drawColor,line width= 0.6pt,line join=round] ( 91.36,215.73) -- ( 91.36,248.34);

\path[draw=drawColor,line width= 0.6pt,line join=round] ( 91.36,186.13) -- ( 91.36,181.24);
\definecolor{fillColor}{RGB}{199,215,238}

\path[draw=drawColor,line width= 0.6pt,line join=round,line cap=round,fill=fillColor] ( 80.18,215.73) --
	( 80.18,186.13) --
	(102.55,186.13) --
	(102.55,215.73) --
	( 80.18,215.73) --
	cycle;

\path[draw=drawColor,line width= 1.1pt,line join=round] ( 80.18,201.37) -- (102.55,201.37);

\path[draw=drawColor,line width= 0.6pt,line join=round] (116.22,276.12) -- (116.22,280.74);

\path[draw=drawColor,line width= 0.6pt,line join=round] (116.22,254.66) -- (116.22,242.71);
\definecolor{fillColor}{RGB}{255,221,227}

\path[draw=drawColor,line width= 0.6pt,line join=round,line cap=round,fill=fillColor] (105.03,276.12) --
	(105.03,254.66) --
	(127.40,254.66) --
	(127.40,276.12) --
	(105.03,276.12) --
	cycle;

\path[draw=drawColor,line width= 1.1pt,line join=round] (105.03,267.99) -- (127.40,267.99);

\path[draw=drawColor,line width= 0.6pt,line join=round,line cap=round] ( 31.71,176.25) rectangle (151.02,285.97);
\end{scope}
\begin{scope}
\path[clip] ( 31.71, 30.69) rectangle (151.02,140.40);
\definecolor{fillColor}{RGB}{255,255,255}

\path[fill=fillColor] ( 31.71, 30.69) rectangle (151.02,140.40);
\definecolor{drawColor}{gray}{0.92}

\path[draw=drawColor,line width= 0.3pt,line join=round] ( 31.71, 32.71) --
	(151.02, 32.71);

\path[draw=drawColor,line width= 0.3pt,line join=round] ( 31.71, 64.71) --
	(151.02, 64.71);

\path[draw=drawColor,line width= 0.3pt,line join=round] ( 31.71, 96.70) --
	(151.02, 96.70);

\path[draw=drawColor,line width= 0.3pt,line join=round] ( 31.71,128.70) --
	(151.02,128.70);

\path[draw=drawColor,line width= 0.6pt,line join=round] ( 31.71, 48.71) --
	(151.02, 48.71);

\path[draw=drawColor,line width= 0.6pt,line join=round] ( 31.71, 80.70) --
	(151.02, 80.70);

\path[draw=drawColor,line width= 0.6pt,line join=round] ( 31.71,112.70) --
	(151.02,112.70);

\path[draw=drawColor,line width= 0.6pt,line join=round] ( 91.36, 30.69) --
	( 91.36,140.40);
\definecolor{drawColor}{gray}{0.20}

\path[draw=drawColor,line width= 0.6pt,line join=round] ( 66.51,110.83) -- ( 66.51,117.80);

\path[draw=drawColor,line width= 0.6pt,line join=round] ( 66.51, 99.60) -- ( 66.51, 96.21);
\definecolor{fillColor}{RGB}{254,255,254}

\path[draw=drawColor,line width= 0.6pt,line join=round,line cap=round,fill=fillColor] ( 55.32,110.83) --
	( 55.32, 99.60) --
	( 77.69, 99.60) --
	( 77.69,110.83) --
	( 55.32,110.83) --
	cycle;

\path[draw=drawColor,line width= 1.1pt,line join=round] ( 55.32,105.38) -- ( 77.69,105.38);

\path[draw=drawColor,line width= 0.6pt,line join=round] ( 91.36, 86.85) -- ( 91.36, 97.50);

\path[draw=drawColor,line width= 0.6pt,line join=round] ( 91.36, 55.27) -- ( 91.36, 35.67);
\definecolor{fillColor}{RGB}{199,215,238}

\path[draw=drawColor,line width= 0.6pt,line join=round,line cap=round,fill=fillColor] ( 80.18, 86.85) --
	( 80.18, 55.27) --
	(102.55, 55.27) --
	(102.55, 86.85) --
	( 80.18, 86.85) --
	cycle;

\path[draw=drawColor,line width= 1.1pt,line join=round] ( 80.18, 77.17) -- (102.55, 77.17);

\path[draw=drawColor,line width= 0.6pt,line join=round] (116.22,128.27) -- (116.22,135.41);

\path[draw=drawColor,line width= 0.6pt,line join=round] (116.22,104.92) -- (116.22, 93.75);
\definecolor{fillColor}{RGB}{255,221,227}

\path[draw=drawColor,line width= 0.6pt,line join=round,line cap=round,fill=fillColor] (105.03,128.27) --
	(105.03,104.92) --
	(127.40,104.92) --
	(127.40,128.27) --
	(105.03,128.27) --
	cycle;

\path[draw=drawColor,line width= 1.1pt,line join=round] (105.03,116.71) -- (127.40,116.71);

\path[draw=drawColor,line width= 0.6pt,line join=round,line cap=round] ( 31.71, 30.69) rectangle (151.02,140.40);
\end{scope}
\begin{scope}
\path[clip] (170.26,176.25) rectangle (289.57,285.97);
\definecolor{fillColor}{RGB}{255,255,255}

\path[fill=fillColor] (170.26,176.25) rectangle (289.57,285.97);
\definecolor{drawColor}{gray}{0.92}

\path[draw=drawColor,line width= 0.3pt,line join=round] (170.26,205.08) --
	(289.57,205.08);

\path[draw=drawColor,line width= 0.3pt,line join=round] (170.26,235.70) --
	(289.57,235.70);

\path[draw=drawColor,line width= 0.3pt,line join=round] (170.26,266.32) --
	(289.57,266.32);

\path[draw=drawColor,line width= 0.6pt,line join=round] (170.26,189.77) --
	(289.57,189.77);

\path[draw=drawColor,line width= 0.6pt,line join=round] (170.26,220.39) --
	(289.57,220.39);

\path[draw=drawColor,line width= 0.6pt,line join=round] (170.26,251.01) --
	(289.57,251.01);

\path[draw=drawColor,line width= 0.6pt,line join=round] (170.26,281.63) --
	(289.57,281.63);

\path[draw=drawColor,line width= 0.6pt,line join=round] (229.92,176.25) --
	(229.92,285.97);
\definecolor{drawColor}{gray}{0.20}
\definecolor{fillColor}{gray}{0.20}

\path[draw=drawColor,line width= 0.4pt,line join=round,line cap=round,fill=fillColor] (205.06,226.28) circle (  1.96);

\path[draw=drawColor,line width= 0.6pt,line join=round] (205.06,265.02) -- (205.06,278.35);

\path[draw=drawColor,line width= 0.6pt,line join=round] (205.06,255.47) -- (205.06,254.14);
\definecolor{fillColor}{RGB}{254,255,254}

\path[draw=drawColor,line width= 0.6pt,line join=round,line cap=round,fill=fillColor] (193.88,265.02) --
	(193.88,255.47) --
	(216.25,255.47) --
	(216.25,265.02) --
	(193.88,265.02) --
	cycle;

\path[draw=drawColor,line width= 1.1pt,line join=round] (193.88,260.82) -- (216.25,260.82);

\path[draw=drawColor,line width= 0.6pt,line join=round] (229.92,225.64) -- (229.92,242.01);

\path[draw=drawColor,line width= 0.6pt,line join=round] (229.92,191.35) -- (229.92,181.24);
\definecolor{fillColor}{RGB}{199,215,238}

\path[draw=drawColor,line width= 0.6pt,line join=round,line cap=round,fill=fillColor] (218.73,225.64) --
	(218.73,191.35) --
	(241.10,191.35) --
	(241.10,225.64) --
	(218.73,225.64) --
	cycle;

\path[draw=drawColor,line width= 1.1pt,line join=round] (218.73,217.87) -- (241.10,217.87);
\definecolor{fillColor}{gray}{0.20}

\path[draw=drawColor,line width= 0.4pt,line join=round,line cap=round,fill=fillColor] (254.77,210.82) circle (  1.96);

\path[draw=drawColor,line width= 0.6pt,line join=round] (254.77,271.97) -- (254.77,280.98);

\path[draw=drawColor,line width= 0.6pt,line join=round] (254.77,251.22) -- (254.77,245.17);
\definecolor{fillColor}{RGB}{255,221,227}

\path[draw=drawColor,line width= 0.6pt,line join=round,line cap=round,fill=fillColor] (243.59,271.97) --
	(243.59,251.22) --
	(265.96,251.22) --
	(265.96,271.97) --
	(243.59,271.97) --
	cycle;

\path[draw=drawColor,line width= 1.1pt,line join=round] (243.59,265.87) -- (265.96,265.87);

\path[draw=drawColor,line width= 0.6pt,line join=round,line cap=round] (170.26,176.25) rectangle (289.57,285.97);
\end{scope}
\begin{scope}
\path[clip] (170.26, 30.69) rectangle (289.57,140.40);
\definecolor{fillColor}{RGB}{255,255,255}

\path[fill=fillColor] (170.26, 30.69) rectangle (289.57,140.40);
\definecolor{drawColor}{gray}{0.92}

\path[draw=drawColor,line width= 0.3pt,line join=round] (170.26, 42.10) --
	(289.57, 42.10);

\path[draw=drawColor,line width= 0.3pt,line join=round] (170.26, 67.99) --
	(289.57, 67.99);

\path[draw=drawColor,line width= 0.3pt,line join=round] (170.26, 93.88) --
	(289.57, 93.88);

\path[draw=drawColor,line width= 0.3pt,line join=round] (170.26,119.77) --
	(289.57,119.77);

\path[draw=drawColor,line width= 0.6pt,line join=round] (170.26, 55.05) --
	(289.57, 55.05);

\path[draw=drawColor,line width= 0.6pt,line join=round] (170.26, 80.94) --
	(289.57, 80.94);

\path[draw=drawColor,line width= 0.6pt,line join=round] (170.26,106.82) --
	(289.57,106.82);

\path[draw=drawColor,line width= 0.6pt,line join=round] (170.26,132.71) --
	(289.57,132.71);

\path[draw=drawColor,line width= 0.6pt,line join=round] (229.92, 30.69) --
	(229.92,140.40);
\definecolor{drawColor}{gray}{0.20}

\path[draw=drawColor,line width= 0.6pt,line join=round] (205.06,111.24) -- (205.06,116.65);

\path[draw=drawColor,line width= 0.6pt,line join=round] (205.06, 97.63) -- (205.06, 82.15);
\definecolor{fillColor}{RGB}{254,255,254}

\path[draw=drawColor,line width= 0.6pt,line join=round,line cap=round,fill=fillColor] (193.88,111.24) --
	(193.88, 97.63) --
	(216.25, 97.63) --
	(216.25,111.24) --
	(193.88,111.24) --
	cycle;

\path[draw=drawColor,line width= 1.1pt,line join=round] (193.88,107.61) -- (216.25,107.61);

\path[draw=drawColor,line width= 0.6pt,line join=round] (229.92, 59.01) -- (229.92, 74.61);

\path[draw=drawColor,line width= 0.6pt,line join=round] (229.92, 44.05) -- (229.92, 35.67);
\definecolor{fillColor}{RGB}{199,215,238}

\path[draw=drawColor,line width= 0.6pt,line join=round,line cap=round,fill=fillColor] (218.73, 59.01) --
	(218.73, 44.05) --
	(241.10, 44.05) --
	(241.10, 59.01) --
	(218.73, 59.01) --
	cycle;

\path[draw=drawColor,line width= 1.1pt,line join=round] (218.73, 47.63) -- (241.10, 47.63);

\path[draw=drawColor,line width= 0.6pt,line join=round] (254.77,124.02) -- (254.77,135.41);

\path[draw=drawColor,line width= 0.6pt,line join=round] (254.77,111.53) -- (254.77,107.55);
\definecolor{fillColor}{RGB}{255,221,227}

\path[draw=drawColor,line width= 0.6pt,line join=round,line cap=round,fill=fillColor] (243.59,124.02) --
	(243.59,111.53) --
	(265.96,111.53) --
	(265.96,124.02) --
	(243.59,124.02) --
	cycle;

\path[draw=drawColor,line width= 1.1pt,line join=round] (243.59,116.27) -- (265.96,116.27);

\path[draw=drawColor,line width= 0.6pt,line join=round,line cap=round] (170.26, 30.69) rectangle (289.57,140.40);
\end{scope}
\begin{scope}
\path[clip] (308.82,176.25) rectangle (428.12,285.97);
\definecolor{fillColor}{RGB}{255,255,255}

\path[fill=fillColor] (308.82,176.25) rectangle (428.12,285.97);
\definecolor{drawColor}{gray}{0.92}

\path[draw=drawColor,line width= 0.3pt,line join=round] (308.82,186.32) --
	(428.12,186.32);

\path[draw=drawColor,line width= 0.3pt,line join=round] (308.82,230.01) --
	(428.12,230.01);

\path[draw=drawColor,line width= 0.3pt,line join=round] (308.82,273.69) --
	(428.12,273.69);

\path[draw=drawColor,line width= 0.6pt,line join=round] (308.82,208.16) --
	(428.12,208.16);

\path[draw=drawColor,line width= 0.6pt,line join=round] (308.82,251.85) --
	(428.12,251.85);

\path[draw=drawColor,line width= 0.6pt,line join=round] (368.47,176.25) --
	(368.47,285.97);
\definecolor{drawColor}{gray}{0.20}

\path[draw=drawColor,line width= 0.6pt,line join=round] (343.61,239.13) -- (343.61,251.89);

\path[draw=drawColor,line width= 0.6pt,line join=round] (343.61,222.16) -- (343.61,212.54);
\definecolor{fillColor}{RGB}{254,255,254}

\path[draw=drawColor,line width= 0.6pt,line join=round,line cap=round,fill=fillColor] (332.43,239.13) --
	(332.43,222.16) --
	(354.80,222.16) --
	(354.80,239.13) --
	(332.43,239.13) --
	cycle;

\path[draw=drawColor,line width= 1.1pt,line join=round] (332.43,231.95) -- (354.80,231.95);
\definecolor{fillColor}{gray}{0.20}

\path[draw=drawColor,line width= 0.4pt,line join=round,line cap=round,fill=fillColor] (368.47,181.24) circle (  1.96);

\path[draw=drawColor,line width= 0.6pt,line join=round] (368.47,224.06) -- (368.47,229.27);

\path[draw=drawColor,line width= 0.6pt,line join=round] (368.47,208.70) -- (368.47,204.02);
\definecolor{fillColor}{RGB}{199,215,238}

\path[draw=drawColor,line width= 0.6pt,line join=round,line cap=round,fill=fillColor] (357.28,224.06) --
	(357.28,208.70) --
	(379.65,208.70) --
	(379.65,224.06) --
	(357.28,224.06) --
	cycle;

\path[draw=drawColor,line width= 1.1pt,line join=round] (357.28,217.29) -- (379.65,217.29);

\path[draw=drawColor,line width= 0.6pt,line join=round] (393.32,275.83) -- (393.32,280.98);

\path[draw=drawColor,line width= 0.6pt,line join=round] (393.32,260.91) -- (393.32,255.43);
\definecolor{fillColor}{RGB}{255,221,227}

\path[draw=drawColor,line width= 0.6pt,line join=round,line cap=round,fill=fillColor] (382.14,275.83) --
	(382.14,260.91) --
	(404.51,260.91) --
	(404.51,275.83) --
	(382.14,275.83) --
	cycle;

\path[draw=drawColor,line width= 1.1pt,line join=round] (382.14,264.80) -- (404.51,264.80);

\path[draw=drawColor,line width= 0.6pt,line join=round,line cap=round] (308.82,176.25) rectangle (428.12,285.97);
\end{scope}
\begin{scope}
\path[clip] ( 31.71,140.40) rectangle (151.02,158.03);
\definecolor{drawColor}{gray}{0.20}
\definecolor{fillColor}{gray}{0.85}

\path[draw=drawColor,line width= 0.6pt,line join=round,line cap=round,fill=fillColor] ( 31.71,140.40) rectangle (151.02,158.03);
\definecolor{drawColor}{gray}{0.10}

\node[text=drawColor,anchor=base,inner sep=0pt, outer sep=0pt, scale=  1.00] at ( 91.36,145.77) {Stocks};
\end{scope}
\begin{scope}
\path[clip] (170.26,140.40) rectangle (289.57,158.03);
\definecolor{drawColor}{gray}{0.20}
\definecolor{fillColor}{gray}{0.85}

\path[draw=drawColor,line width= 0.6pt,line join=round,line cap=round,fill=fillColor] (170.26,140.40) rectangle (289.57,158.03);
\definecolor{drawColor}{gray}{0.10}

\node[text=drawColor,anchor=base,inner sep=0pt, outer sep=0pt, scale=  1.00] at (229.92,145.77) {Average};
\end{scope}
\begin{scope}
\path[clip] ( 31.71,285.97) rectangle (151.02,303.60);
\definecolor{drawColor}{gray}{0.20}
\definecolor{fillColor}{gray}{0.85}

\path[draw=drawColor,line width= 0.6pt,line join=round,line cap=round,fill=fillColor] ( 31.71,285.97) rectangle (151.02,303.60);
\definecolor{drawColor}{gray}{0.10}

\node[text=drawColor,anchor=base,inner sep=0pt, outer sep=0pt, scale=  1.00] at ( 91.36,291.34) {Serial};
\end{scope}
\begin{scope}
\path[clip] (170.26,285.97) rectangle (289.57,303.60);
\definecolor{drawColor}{gray}{0.20}
\definecolor{fillColor}{gray}{0.85}

\path[draw=drawColor,line width= 0.6pt,line join=round,line cap=round,fill=fillColor] (170.26,285.97) rectangle (289.57,303.60);
\definecolor{drawColor}{gray}{0.10}

\node[text=drawColor,anchor=base,inner sep=0pt, outer sep=0pt, scale=  1.00] at (229.92,291.34) {Weather};
\end{scope}
\begin{scope}
\path[clip] (308.82,285.97) rectangle (428.12,303.60);
\definecolor{drawColor}{gray}{0.20}
\definecolor{fillColor}{gray}{0.85}

\path[draw=drawColor,line width= 0.6pt,line join=round,line cap=round,fill=fillColor] (308.82,285.97) rectangle (428.12,303.60);
\definecolor{drawColor}{gray}{0.10}

\node[text=drawColor,anchor=base,inner sep=0pt, outer sep=0pt, scale=  1.00] at (368.47,291.34) {Death notice};
\end{scope}
\begin{scope}
\path[clip] (  0.00,  0.00) rectangle (433.62,325.21);
\definecolor{drawColor}{gray}{0.30}

\node[text=drawColor,anchor=base,inner sep=0pt, outer sep=0pt, scale=  0.88] at ( 91.36, 19.68) {mIoU};
\end{scope}
\begin{scope}
\path[clip] (  0.00,  0.00) rectangle (433.62,325.21);
\definecolor{drawColor}{gray}{0.30}

\node[text=drawColor,anchor=base,inner sep=0pt, outer sep=0pt, scale=  0.88] at (229.92, 19.68) {mIoU};
\end{scope}
\begin{scope}
\path[clip] (  0.00,  0.00) rectangle (433.62,325.21);
\definecolor{drawColor}{gray}{0.30}

\node[text=drawColor,anchor=base,inner sep=0pt, outer sep=0pt, scale=  0.88] at ( 91.36,165.24) {mIoU};
\end{scope}
\begin{scope}
\path[clip] (  0.00,  0.00) rectangle (433.62,325.21);
\definecolor{drawColor}{gray}{0.30}

\node[text=drawColor,anchor=base,inner sep=0pt, outer sep=0pt, scale=  0.88] at (229.92,165.24) {mIoU};
\end{scope}
\begin{scope}
\path[clip] (  0.00,  0.00) rectangle (433.62,325.21);
\definecolor{drawColor}{gray}{0.30}

\node[text=drawColor,anchor=base,inner sep=0pt, outer sep=0pt, scale=  0.88] at (368.47,165.24) {mIoU};
\end{scope}
\begin{scope}
\path[clip] (  0.00,  0.00) rectangle (433.62,325.21);
\definecolor{drawColor}{gray}{0.30}

\node[text=drawColor,anchor=base east,inner sep=0pt, outer sep=0pt, scale=  0.88] at (303.87,205.13) {70};

\node[text=drawColor,anchor=base east,inner sep=0pt, outer sep=0pt, scale=  0.88] at (303.87,248.82) {80};
\end{scope}
\begin{scope}
\path[clip] (  0.00,  0.00) rectangle (433.62,325.21);
\definecolor{drawColor}{gray}{0.20}

\path[draw=drawColor,line width= 0.6pt,line join=round] (306.07,208.16) --
	(308.82,208.16);

\path[draw=drawColor,line width= 0.6pt,line join=round] (306.07,251.85) --
	(308.82,251.85);
\end{scope}
\begin{scope}
\path[clip] (  0.00,  0.00) rectangle (433.62,325.21);
\definecolor{drawColor}{gray}{0.30}

\node[text=drawColor,anchor=base east,inner sep=0pt, outer sep=0pt, scale=  0.88] at (165.31,186.74) {70};

\node[text=drawColor,anchor=base east,inner sep=0pt, outer sep=0pt, scale=  0.88] at (165.31,217.36) {75};

\node[text=drawColor,anchor=base east,inner sep=0pt, outer sep=0pt, scale=  0.88] at (165.31,247.98) {80};

\node[text=drawColor,anchor=base east,inner sep=0pt, outer sep=0pt, scale=  0.88] at (165.31,278.60) {85};
\end{scope}
\begin{scope}
\path[clip] (  0.00,  0.00) rectangle (433.62,325.21);
\definecolor{drawColor}{gray}{0.20}

\path[draw=drawColor,line width= 0.6pt,line join=round] (167.51,189.77) --
	(170.26,189.77);

\path[draw=drawColor,line width= 0.6pt,line join=round] (167.51,220.39) --
	(170.26,220.39);

\path[draw=drawColor,line width= 0.6pt,line join=round] (167.51,251.01) --
	(170.26,251.01);

\path[draw=drawColor,line width= 0.6pt,line join=round] (167.51,281.63) --
	(170.26,281.63);
\end{scope}
\begin{scope}
\path[clip] (  0.00,  0.00) rectangle (433.62,325.21);
\definecolor{drawColor}{gray}{0.30}

\node[text=drawColor,anchor=base east,inner sep=0pt, outer sep=0pt, scale=  0.88] at (165.31, 52.02) {70};

\node[text=drawColor,anchor=base east,inner sep=0pt, outer sep=0pt, scale=  0.88] at (165.31, 77.91) {75};

\node[text=drawColor,anchor=base east,inner sep=0pt, outer sep=0pt, scale=  0.88] at (165.31,103.79) {80};

\node[text=drawColor,anchor=base east,inner sep=0pt, outer sep=0pt, scale=  0.88] at (165.31,129.68) {85};
\end{scope}
\begin{scope}
\path[clip] (  0.00,  0.00) rectangle (433.62,325.21);
\definecolor{drawColor}{gray}{0.20}

\path[draw=drawColor,line width= 0.6pt,line join=round] (167.51, 55.05) --
	(170.26, 55.05);

\path[draw=drawColor,line width= 0.6pt,line join=round] (167.51, 80.94) --
	(170.26, 80.94);

\path[draw=drawColor,line width= 0.6pt,line join=round] (167.51,106.82) --
	(170.26,106.82);

\path[draw=drawColor,line width= 0.6pt,line join=round] (167.51,132.71) --
	(170.26,132.71);
\end{scope}
\begin{scope}
\path[clip] (  0.00,  0.00) rectangle (433.62,325.21);
\definecolor{drawColor}{gray}{0.30}

\node[text=drawColor,anchor=base east,inner sep=0pt, outer sep=0pt, scale=  0.88] at ( 26.76,180.56) {40};

\node[text=drawColor,anchor=base east,inner sep=0pt, outer sep=0pt, scale=  0.88] at ( 26.76,202.86) {50};

\node[text=drawColor,anchor=base east,inner sep=0pt, outer sep=0pt, scale=  0.88] at ( 26.76,225.15) {60};

\node[text=drawColor,anchor=base east,inner sep=0pt, outer sep=0pt, scale=  0.88] at ( 26.76,247.45) {70};

\node[text=drawColor,anchor=base east,inner sep=0pt, outer sep=0pt, scale=  0.88] at ( 26.76,269.74) {80};
\end{scope}
\begin{scope}
\path[clip] (  0.00,  0.00) rectangle (433.62,325.21);
\definecolor{drawColor}{gray}{0.20}

\path[draw=drawColor,line width= 0.6pt,line join=round] ( 28.96,183.59) --
	( 31.71,183.59);

\path[draw=drawColor,line width= 0.6pt,line join=round] ( 28.96,205.89) --
	( 31.71,205.89);

\path[draw=drawColor,line width= 0.6pt,line join=round] ( 28.96,228.18) --
	( 31.71,228.18);

\path[draw=drawColor,line width= 0.6pt,line join=round] ( 28.96,250.48) --
	( 31.71,250.48);

\path[draw=drawColor,line width= 0.6pt,line join=round] ( 28.96,272.77) --
	( 31.71,272.77);
\end{scope}
\begin{scope}
\path[clip] (  0.00,  0.00) rectangle (433.62,325.21);
\definecolor{drawColor}{gray}{0.30}

\node[text=drawColor,anchor=base east,inner sep=0pt, outer sep=0pt, scale=  0.88] at ( 26.76, 45.68) {76};

\node[text=drawColor,anchor=base east,inner sep=0pt, outer sep=0pt, scale=  0.88] at ( 26.76, 77.67) {80};

\node[text=drawColor,anchor=base east,inner sep=0pt, outer sep=0pt, scale=  0.88] at ( 26.76,109.67) {84};
\end{scope}
\begin{scope}
\path[clip] (  0.00,  0.00) rectangle (433.62,325.21);
\definecolor{drawColor}{gray}{0.20}

\path[draw=drawColor,line width= 0.6pt,line join=round] ( 28.96, 48.71) --
	( 31.71, 48.71);

\path[draw=drawColor,line width= 0.6pt,line join=round] ( 28.96, 80.70) --
	( 31.71, 80.70);

\path[draw=drawColor,line width= 0.6pt,line join=round] ( 28.96,112.70) --
	( 31.71,112.70);
\end{scope}
\begin{scope}
\path[clip] (  0.00,  0.00) rectangle (433.62,325.21);
\definecolor{drawColor}{RGB}{0,0,0}

\node[text=drawColor,rotate= 90.00,anchor=base,inner sep=0pt, outer sep=0pt, scale=  1.10] at ( 13.08,158.33) {mIoU};
\end{scope}
\begin{scope}
\path[clip] (  0.00,  0.00) rectangle (433.62,325.21);
\definecolor{fillColor}{RGB}{255,255,255}

\path[fill=fillColor] (319.58, 17.21) rectangle (406.41,151.99);
\end{scope}
\begin{scope}
\path[clip] (  0.00,  0.00) rectangle (433.62,325.21);
\definecolor{drawColor}{RGB}{0,0,0}

\node[text=drawColor,anchor=base,inner sep=0pt, outer sep=0pt, scale=  1.10] at (362.99,137.84) {Modalities};
\end{scope}
\begin{scope}
\path[clip] (  0.00,  0.00) rectangle (433.62,325.21);
\definecolor{fillColor}{RGB}{255,255,255}

\path[fill=fillColor] (325.08, 98.75) rectangle (346.76,131.27);
\end{scope}
\begin{scope}
\path[clip] (  0.00,  0.00) rectangle (433.62,325.21);
\definecolor{drawColor}{gray}{0.20}

\path[draw=drawColor,line width= 0.6pt,line join=round,line cap=round] (335.92,102.00) --
	(335.92,106.88);

\path[draw=drawColor,line width= 0.6pt,line join=round,line cap=round] (335.92,123.14) --
	(335.92,128.02);
\definecolor{fillColor}{RGB}{254,255,254}

\path[draw=drawColor,line width= 0.6pt,line join=round,line cap=round,fill=fillColor] (327.79,106.88) rectangle (344.05,123.14);

\path[draw=drawColor,line width= 0.6pt,line join=round,line cap=round] (327.79,115.01) --
	(344.05,115.01);
\end{scope}
\begin{scope}
\path[clip] (  0.00,  0.00) rectangle (433.62,325.21);
\definecolor{fillColor}{RGB}{255,255,255}

\path[fill=fillColor] (325.08, 60.73) rectangle (346.76, 93.25);
\end{scope}
\begin{scope}
\path[clip] (  0.00,  0.00) rectangle (433.62,325.21);
\definecolor{drawColor}{gray}{0.20}

\path[draw=drawColor,line width= 0.6pt,line join=round,line cap=round] (335.92, 63.98) --
	(335.92, 68.86);

\path[draw=drawColor,line width= 0.6pt,line join=round,line cap=round] (335.92, 85.12) --
	(335.92, 90.00);
\definecolor{fillColor}{RGB}{199,215,238}

\path[draw=drawColor,line width= 0.6pt,line join=round,line cap=round,fill=fillColor] (327.79, 68.86) rectangle (344.05, 85.12);

\path[draw=drawColor,line width= 0.6pt,line join=round,line cap=round] (327.79, 76.99) --
	(344.05, 76.99);
\end{scope}
\begin{scope}
\path[clip] (  0.00,  0.00) rectangle (433.62,325.21);
\definecolor{fillColor}{RGB}{255,255,255}

\path[fill=fillColor] (325.08, 22.71) rectangle (346.76, 55.23);
\end{scope}
\begin{scope}
\path[clip] (  0.00,  0.00) rectangle (433.62,325.21);
\definecolor{drawColor}{gray}{0.20}

\path[draw=drawColor,line width= 0.6pt,line join=round,line cap=round] (335.92, 25.96) --
	(335.92, 30.84);

\path[draw=drawColor,line width= 0.6pt,line join=round,line cap=round] (335.92, 47.10) --
	(335.92, 51.98);
\definecolor{fillColor}{RGB}{255,221,227}

\path[draw=drawColor,line width= 0.6pt,line join=round,line cap=round,fill=fillColor] (327.79, 30.84) rectangle (344.05, 47.10);

\path[draw=drawColor,line width= 0.6pt,line join=round,line cap=round] (327.79, 38.97) --
	(344.05, 38.97);
\end{scope}
\begin{scope}
\path[clip] (  0.00,  0.00) rectangle (433.62,325.21);
\definecolor{drawColor}{RGB}{0,0,0}

\node[text=drawColor,anchor=base west,inner sep=0pt, outer sep=0pt, scale=  0.80] at (352.26,112.26) {Image};
\end{scope}
\begin{scope}
\path[clip] (  0.00,  0.00) rectangle (433.62,325.21);
\definecolor{drawColor}{RGB}{0,0,0}

\node[text=drawColor,anchor=base west,inner sep=0pt, outer sep=0pt, scale=  0.80] at (352.26, 74.24) {Text};
\end{scope}
\begin{scope}
\path[clip] (  0.00,  0.00) rectangle (433.62,325.21);
\definecolor{drawColor}{RGB}{0,0,0}

\node[text=drawColor,anchor=base west,inner sep=0pt, outer sep=0pt, scale=  0.80] at (352.26, 36.21) {Image + Text};
\end{scope}
\begin{scope}
\path[clip] (  0.00,  0.00) rectangle (433.62,325.21);
\definecolor{drawColor}{RGB}{0,0,0}

\node[text=drawColor,anchor=base,inner sep=0pt, outer sep=0pt, scale=  1.20] at (229.92,311.43) {\bfseries Comparison of mIoU per experiment};
\end{scope}
\end{tikzpicture}

%% file: Figures/generalization.tex
\begin{tikzpicture}[x=1pt,y=1pt]
\definecolor{fillColor}{RGB}{255,255,255}
\path[use as bounding box,fill=fillColor,fill opacity=0.00] (0,0) rectangle (433.62,325.21);
\begin{scope}
\path[clip] (  0.00,  0.00) rectangle (433.62,325.21);
\definecolor{drawColor}{RGB}{255,255,255}
\definecolor{fillColor}{RGB}{255,255,255}

\path[draw=drawColor,line width= 0.6pt,line join=round,line cap=round,fill=fillColor] (  0.00,  0.00) rectangle (433.62,325.21);
\end{scope}
\begin{scope}
\path[clip] ( 31.71,176.25) rectangle (151.02,285.97);
\definecolor{fillColor}{RGB}{255,255,255}

\path[fill=fillColor] ( 31.71,176.25) rectangle (151.02,285.97);
\definecolor{drawColor}{gray}{0.92}

\path[draw=drawColor,line width= 0.3pt,line join=round] ( 31.71,190.39) --
	(151.02,190.39);

\path[draw=drawColor,line width= 0.3pt,line join=round] ( 31.71,216.62) --
	(151.02,216.62);

\path[draw=drawColor,line width= 0.3pt,line join=round] ( 31.71,242.85) --
	(151.02,242.85);

\path[draw=drawColor,line width= 0.3pt,line join=round] ( 31.71,269.07) --
	(151.02,269.07);

\path[draw=drawColor,line width= 0.6pt,line join=round] ( 31.71,177.28) --
	(151.02,177.28);

\path[draw=drawColor,line width= 0.6pt,line join=round] ( 31.71,203.51) --
	(151.02,203.51);

\path[draw=drawColor,line width= 0.6pt,line join=round] ( 31.71,229.73) --
	(151.02,229.73);

\path[draw=drawColor,line width= 0.6pt,line join=round] ( 31.71,255.96) --
	(151.02,255.96);

\path[draw=drawColor,line width= 0.6pt,line join=round] ( 31.71,282.19) --
	(151.02,282.19);

\path[draw=drawColor,line width= 0.6pt,line join=round] ( 54.08,176.25) --
	( 54.08,285.97);

\path[draw=drawColor,line width= 0.6pt,line join=round] ( 91.36,176.25) --
	( 91.36,285.97);

\path[draw=drawColor,line width= 0.6pt,line join=round] (128.65,176.25) --
	(128.65,285.97);
\definecolor{drawColor}{gray}{0.20}
\definecolor{fillColor}{gray}{0.20}

\path[draw=drawColor,line width= 0.4pt,line join=round,line cap=round,fill=fillColor] ( 47.09,194.55) circle (  1.96);

\path[draw=drawColor,line width= 0.4pt,line join=round,line cap=round,fill=fillColor] ( 47.09,181.24) circle (  1.96);

\path[draw=drawColor,line width= 0.6pt,line join=round] ( 47.09,189.14) -- ( 47.09,190.54);

\path[draw=drawColor,line width= 0.6pt,line join=round] ( 47.09,186.06) -- ( 47.09,185.72);
\definecolor{fillColor}{RGB}{254,255,254}

\path[draw=drawColor,line width= 0.6pt,line join=round,line cap=round,fill=fillColor] ( 40.80,189.14) --
	( 40.80,186.06) --
	( 53.38,186.06) --
	( 53.38,189.14) --
	( 40.80,189.14) --
	cycle;

\path[draw=drawColor,line width= 1.1pt,line join=round] ( 40.80,187.88) -- ( 53.38,187.88);

\path[draw=drawColor,line width= 0.6pt,line join=round] ( 61.07,217.85) -- ( 61.07,222.68);

\path[draw=drawColor,line width= 0.6pt,line join=round] ( 61.07,202.58) -- ( 61.07,193.10);
\definecolor{fillColor}{RGB}{255,221,227}

\path[draw=drawColor,line width= 0.6pt,line join=round,line cap=round,fill=fillColor] ( 54.78,217.85) --
	( 54.78,202.58) --
	( 67.36,202.58) --
	( 67.36,217.85) --
	( 54.78,217.85) --
	cycle;

\path[draw=drawColor,line width= 1.1pt,line join=round] ( 54.78,211.85) -- ( 67.36,211.85);

\path[draw=drawColor,line width= 0.6pt,line join=round] ( 84.37,271.73) -- ( 84.37,280.98);

\path[draw=drawColor,line width= 0.6pt,line join=round] ( 84.37,261.49) -- ( 84.37,253.10);
\definecolor{fillColor}{RGB}{254,255,254}

\path[draw=drawColor,line width= 0.6pt,line join=round,line cap=round,fill=fillColor] ( 78.08,271.73) --
	( 78.08,261.49) --
	( 90.67,261.49) --
	( 90.67,271.73) --
	( 78.08,271.73) --
	cycle;

\path[draw=drawColor,line width= 1.1pt,line join=round] ( 78.08,264.66) -- ( 90.67,264.66);

\path[draw=drawColor,line width= 0.6pt,line join=round] ( 98.35,278.82) -- ( 98.35,279.42);

\path[draw=drawColor,line width= 0.6pt,line join=round] ( 98.35,270.53) -- ( 98.35,264.80);
\definecolor{fillColor}{RGB}{255,221,227}

\path[draw=drawColor,line width= 0.6pt,line join=round,line cap=round,fill=fillColor] ( 92.06,278.82) --
	( 92.06,270.53) --
	(104.65,270.53) --
	(104.65,278.82) --
	( 92.06,278.82) --
	cycle;

\path[draw=drawColor,line width= 1.1pt,line join=round] ( 92.06,275.00) -- (104.65,275.00);

\path[draw=drawColor,line width= 0.6pt,line join=round] (121.66,241.93) -- (121.66,247.36);

\path[draw=drawColor,line width= 0.6pt,line join=round] (121.66,225.92) -- (121.66,218.49);
\definecolor{fillColor}{RGB}{254,255,254}

\path[draw=drawColor,line width= 0.6pt,line join=round,line cap=round,fill=fillColor] (115.36,241.93) --
	(115.36,225.92) --
	(127.95,225.92) --
	(127.95,241.93) --
	(115.36,241.93) --
	cycle;

\path[draw=drawColor,line width= 1.1pt,line join=round] (115.36,232.11) -- (127.95,232.11);

\path[draw=drawColor,line width= 0.6pt,line join=round] (135.64,243.29) -- (135.64,257.87);

\path[draw=drawColor,line width= 0.6pt,line join=round] (135.64,228.97) -- (135.64,226.43);
\definecolor{fillColor}{RGB}{255,221,227}

\path[draw=drawColor,line width= 0.6pt,line join=round,line cap=round,fill=fillColor] (129.35,243.29) --
	(129.35,228.97) --
	(141.93,228.97) --
	(141.93,243.29) --
	(129.35,243.29) --
	cycle;

\path[draw=drawColor,line width= 1.1pt,line join=round] (129.35,237.27) -- (141.93,237.27);

\path[draw=drawColor,line width= 0.6pt,line join=round,line cap=round] ( 31.71,176.25) rectangle (151.02,285.97);
\end{scope}
\begin{scope}
\path[clip] ( 31.71, 30.69) rectangle (151.02,140.40);
\definecolor{fillColor}{RGB}{255,255,255}

\path[fill=fillColor] ( 31.71, 30.69) rectangle (151.02,140.40);
\definecolor{drawColor}{gray}{0.92}

\path[draw=drawColor,line width= 0.3pt,line join=round] ( 31.71, 36.23) --
	(151.02, 36.23);

\path[draw=drawColor,line width= 0.3pt,line join=round] ( 31.71, 60.55) --
	(151.02, 60.55);

\path[draw=drawColor,line width= 0.3pt,line join=round] ( 31.71, 84.86) --
	(151.02, 84.86);

\path[draw=drawColor,line width= 0.3pt,line join=round] ( 31.71,109.18) --
	(151.02,109.18);

\path[draw=drawColor,line width= 0.3pt,line join=round] ( 31.71,133.50) --
	(151.02,133.50);

\path[draw=drawColor,line width= 0.6pt,line join=round] ( 31.71, 48.39) --
	(151.02, 48.39);

\path[draw=drawColor,line width= 0.6pt,line join=round] ( 31.71, 72.70) --
	(151.02, 72.70);

\path[draw=drawColor,line width= 0.6pt,line join=round] ( 31.71, 97.02) --
	(151.02, 97.02);

\path[draw=drawColor,line width= 0.6pt,line join=round] ( 31.71,121.34) --
	(151.02,121.34);

\path[draw=drawColor,line width= 0.6pt,line join=round] ( 54.08, 30.69) --
	( 54.08,140.40);

\path[draw=drawColor,line width= 0.6pt,line join=round] ( 91.36, 30.69) --
	( 91.36,140.40);

\path[draw=drawColor,line width= 0.6pt,line join=round] (128.65, 30.69) --
	(128.65,140.40);
\definecolor{drawColor}{gray}{0.20}

\path[draw=drawColor,line width= 0.6pt,line join=round] ( 47.09,121.02) -- ( 47.09,132.46);

\path[draw=drawColor,line width= 0.6pt,line join=round] ( 47.09,111.32) -- ( 47.09,106.58);
\definecolor{fillColor}{RGB}{254,255,254}

\path[draw=drawColor,line width= 0.6pt,line join=round,line cap=round,fill=fillColor] ( 40.80,121.02) --
	( 40.80,111.32) --
	( 53.38,111.32) --
	( 53.38,121.02) --
	( 40.80,121.02) --
	cycle;

\path[draw=drawColor,line width= 1.1pt,line join=round] ( 40.80,117.00) -- ( 53.38,117.00);

\path[draw=drawColor,line width= 0.6pt,line join=round] ( 61.07,112.99) -- ( 61.07,119.14);

\path[draw=drawColor,line width= 0.6pt,line join=round] ( 61.07, 88.25) -- ( 61.07, 70.43);
\definecolor{fillColor}{RGB}{255,221,227}

\path[draw=drawColor,line width= 0.6pt,line join=round,line cap=round,fill=fillColor] ( 54.78,112.99) --
	( 54.78, 88.25) --
	( 67.36, 88.25) --
	( 67.36,112.99) --
	( 54.78,112.99) --
	cycle;

\path[draw=drawColor,line width= 1.1pt,line join=round] ( 54.78, 94.76) -- ( 67.36, 94.76);

\path[draw=drawColor,line width= 0.6pt,line join=round] ( 84.37,131.40) -- ( 84.37,135.41);

\path[draw=drawColor,line width= 0.6pt,line join=round] ( 84.37,122.40) -- ( 84.37,117.24);
\definecolor{fillColor}{RGB}{254,255,254}

\path[draw=drawColor,line width= 0.6pt,line join=round,line cap=round,fill=fillColor] ( 78.08,131.40) --
	( 78.08,122.40) --
	( 90.67,122.40) --
	( 90.67,131.40) --
	( 78.08,131.40) --
	cycle;

\path[draw=drawColor,line width= 1.1pt,line join=round] ( 78.08,128.61) -- ( 90.67,128.61);

\path[draw=drawColor,line width= 0.6pt,line join=round] ( 98.35,129.49) -- ( 98.35,133.91);

\path[draw=drawColor,line width= 0.6pt,line join=round] ( 98.35,124.93) -- ( 98.35,119.76);
\definecolor{fillColor}{RGB}{255,221,227}

\path[draw=drawColor,line width= 0.6pt,line join=round,line cap=round,fill=fillColor] ( 92.06,129.49) --
	( 92.06,124.93) --
	(104.65,124.93) --
	(104.65,129.49) --
	( 92.06,129.49) --
	cycle;

\path[draw=drawColor,line width= 1.1pt,line join=round] ( 92.06,128.46) -- (104.65,128.46);

\path[draw=drawColor,line width= 0.6pt,line join=round] (121.66, 56.81) -- (121.66, 63.29);

\path[draw=drawColor,line width= 0.6pt,line join=round] (121.66, 50.19) -- (121.66, 48.15);
\definecolor{fillColor}{RGB}{254,255,254}

\path[draw=drawColor,line width= 0.6pt,line join=round,line cap=round,fill=fillColor] (115.36, 56.81) --
	(115.36, 50.19) --
	(127.95, 50.19) --
	(127.95, 56.81) --
	(115.36, 56.81) --
	cycle;

\path[draw=drawColor,line width= 1.1pt,line join=round] (115.36, 53.61) -- (127.95, 53.61);

\path[draw=drawColor,line width= 0.6pt,line join=round] (135.64, 68.38) -- (135.64, 83.58);

\path[draw=drawColor,line width= 0.6pt,line join=round] (135.64, 44.35) -- (135.64, 35.67);
\definecolor{fillColor}{RGB}{255,221,227}

\path[draw=drawColor,line width= 0.6pt,line join=round,line cap=round,fill=fillColor] (129.35, 68.38) --
	(129.35, 44.35) --
	(141.93, 44.35) --
	(141.93, 68.38) --
	(129.35, 68.38) --
	cycle;

\path[draw=drawColor,line width= 1.1pt,line join=round] (129.35, 50.10) -- (141.93, 50.10);

\path[draw=drawColor,line width= 0.6pt,line join=round,line cap=round] ( 31.71, 30.69) rectangle (151.02,140.40);
\end{scope}
\begin{scope}
\path[clip] (170.26,176.25) rectangle (289.57,285.97);
\definecolor{fillColor}{RGB}{255,255,255}

\path[fill=fillColor] (170.26,176.25) rectangle (289.57,285.97);
\definecolor{drawColor}{gray}{0.92}

\path[draw=drawColor,line width= 0.3pt,line join=round] (170.26,195.09) --
	(289.57,195.09);

\path[draw=drawColor,line width= 0.3pt,line join=round] (170.26,223.95) --
	(289.57,223.95);

\path[draw=drawColor,line width= 0.3pt,line join=round] (170.26,252.80) --
	(289.57,252.80);

\path[draw=drawColor,line width= 0.3pt,line join=round] (170.26,281.66) --
	(289.57,281.66);

\path[draw=drawColor,line width= 0.6pt,line join=round] (170.26,180.67) --
	(289.57,180.67);

\path[draw=drawColor,line width= 0.6pt,line join=round] (170.26,209.52) --
	(289.57,209.52);

\path[draw=drawColor,line width= 0.6pt,line join=round] (170.26,238.38) --
	(289.57,238.38);

\path[draw=drawColor,line width= 0.6pt,line join=round] (170.26,267.23) --
	(289.57,267.23);

\path[draw=drawColor,line width= 0.6pt,line join=round] (192.63,176.25) --
	(192.63,285.97);

\path[draw=drawColor,line width= 0.6pt,line join=round] (229.92,176.25) --
	(229.92,285.97);

\path[draw=drawColor,line width= 0.6pt,line join=round] (267.20,176.25) --
	(267.20,285.97);
\definecolor{drawColor}{gray}{0.20}

\path[draw=drawColor,line width= 0.6pt,line join=round] (185.64,225.96) -- (185.64,239.75);

\path[draw=drawColor,line width= 0.6pt,line join=round] (185.64,215.83) -- (185.64,211.98);
\definecolor{fillColor}{RGB}{254,255,254}

\path[draw=drawColor,line width= 0.6pt,line join=round,line cap=round,fill=fillColor] (179.35,225.96) --
	(179.35,215.83) --
	(191.93,215.83) --
	(191.93,225.96) --
	(179.35,225.96) --
	cycle;

\path[draw=drawColor,line width= 1.1pt,line join=round] (179.35,223.30) -- (191.93,223.30);
\definecolor{fillColor}{gray}{0.20}

\path[draw=drawColor,line width= 0.4pt,line join=round,line cap=round,fill=fillColor] (199.62,236.24) circle (  1.96);

\path[draw=drawColor,line width= 0.6pt,line join=round] (199.62,279.13) -- (199.62,280.98);

\path[draw=drawColor,line width= 0.6pt,line join=round] (199.62,265.51) -- (199.62,248.67);
\definecolor{fillColor}{RGB}{255,221,227}

\path[draw=drawColor,line width= 0.6pt,line join=round,line cap=round,fill=fillColor] (193.33,279.13) --
	(193.33,265.51) --
	(205.92,265.51) --
	(205.92,279.13) --
	(193.33,279.13) --
	cycle;

\path[draw=drawColor,line width= 1.1pt,line join=round] (193.33,272.93) -- (205.92,272.93);

\path[draw=drawColor,line width= 0.6pt,line join=round] (222.93,268.45) -- (222.93,271.41);

\path[draw=drawColor,line width= 0.6pt,line join=round] (222.93,263.11) -- (222.93,258.55);
\definecolor{fillColor}{RGB}{254,255,254}

\path[draw=drawColor,line width= 0.6pt,line join=round,line cap=round,fill=fillColor] (216.63,268.45) --
	(216.63,263.11) --
	(229.22,263.11) --
	(229.22,268.45) --
	(216.63,268.45) --
	cycle;

\path[draw=drawColor,line width= 1.1pt,line join=round] (216.63,264.53) -- (229.22,264.53);
\definecolor{fillColor}{gray}{0.20}

\path[draw=drawColor,line width= 0.4pt,line join=round,line cap=round,fill=fillColor] (236.91,259.87) circle (  1.96);

\path[draw=drawColor,line width= 0.6pt,line join=round] (236.91,267.47) -- (236.91,271.42);

\path[draw=drawColor,line width= 0.6pt,line join=round] (236.91,264.55) -- (236.91,263.73);
\definecolor{fillColor}{RGB}{255,221,227}

\path[draw=drawColor,line width= 0.6pt,line join=round,line cap=round,fill=fillColor] (230.62,267.47) --
	(230.62,264.55) --
	(243.20,264.55) --
	(243.20,267.47) --
	(230.62,267.47) --
	cycle;

\path[draw=drawColor,line width= 1.1pt,line join=round] (230.62,265.92) -- (243.20,265.92);
\definecolor{fillColor}{gray}{0.20}

\path[draw=drawColor,line width= 0.4pt,line join=round,line cap=round,fill=fillColor] (260.21,181.24) circle (  1.96);

\path[draw=drawColor,line width= 0.4pt,line join=round,line cap=round,fill=fillColor] (260.21,205.20) circle (  1.96);

\path[draw=drawColor,line width= 0.6pt,line join=round] (260.21,192.65) -- (260.21,198.67);

\path[draw=drawColor,line width= 0.6pt,line join=round] (260.21,188.21) -- (260.21,182.34);
\definecolor{fillColor}{RGB}{254,255,254}

\path[draw=drawColor,line width= 0.6pt,line join=round,line cap=round,fill=fillColor] (253.92,192.65) --
	(253.92,188.21) --
	(266.50,188.21) --
	(266.50,192.65) --
	(253.92,192.65) --
	cycle;

\path[draw=drawColor,line width= 1.1pt,line join=round] (253.92,190.20) -- (266.50,190.20);

\path[draw=drawColor,line width= 0.6pt,line join=round] (274.19,208.20) -- (274.19,213.37);

\path[draw=drawColor,line width= 0.6pt,line join=round] (274.19,200.84) -- (274.19,196.91);
\definecolor{fillColor}{RGB}{255,221,227}

\path[draw=drawColor,line width= 0.6pt,line join=round,line cap=round,fill=fillColor] (267.90,208.20) --
	(267.90,200.84) --
	(280.48,200.84) --
	(280.48,208.20) --
	(267.90,208.20) --
	cycle;

\path[draw=drawColor,line width= 1.1pt,line join=round] (267.90,206.49) -- (280.48,206.49);

\path[draw=drawColor,line width= 0.6pt,line join=round,line cap=round] (170.26,176.25) rectangle (289.57,285.97);
\end{scope}
\begin{scope}
\path[clip] (170.26, 30.69) rectangle (289.57,140.40);
\definecolor{fillColor}{RGB}{255,255,255}

\path[fill=fillColor] (170.26, 30.69) rectangle (289.57,140.40);
\definecolor{drawColor}{gray}{0.92}

\path[draw=drawColor,line width= 0.3pt,line join=round] (170.26, 43.35) --
	(289.57, 43.35);

\path[draw=drawColor,line width= 0.3pt,line join=round] (170.26, 70.92) --
	(289.57, 70.92);

\path[draw=drawColor,line width= 0.3pt,line join=round] (170.26, 98.50) --
	(289.57, 98.50);

\path[draw=drawColor,line width= 0.3pt,line join=round] (170.26,126.07) --
	(289.57,126.07);

\path[draw=drawColor,line width= 0.6pt,line join=round] (170.26, 57.14) --
	(289.57, 57.14);

\path[draw=drawColor,line width= 0.6pt,line join=round] (170.26, 84.71) --
	(289.57, 84.71);

\path[draw=drawColor,line width= 0.6pt,line join=round] (170.26,112.28) --
	(289.57,112.28);

\path[draw=drawColor,line width= 0.6pt,line join=round] (170.26,139.86) --
	(289.57,139.86);

\path[draw=drawColor,line width= 0.6pt,line join=round] (192.63, 30.69) --
	(192.63,140.40);

\path[draw=drawColor,line width= 0.6pt,line join=round] (229.92, 30.69) --
	(229.92,140.40);

\path[draw=drawColor,line width= 0.6pt,line join=round] (267.20, 30.69) --
	(267.20,140.40);
\definecolor{drawColor}{gray}{0.20}

\path[draw=drawColor,line width= 0.6pt,line join=round] (185.64, 89.55) -- (185.64,110.12);

\path[draw=drawColor,line width= 0.6pt,line join=round] (185.64, 72.14) -- (185.64, 64.76);
\definecolor{fillColor}{RGB}{254,255,254}

\path[draw=drawColor,line width= 0.6pt,line join=round,line cap=round,fill=fillColor] (179.35, 89.55) --
	(179.35, 72.14) --
	(191.93, 72.14) --
	(191.93, 89.55) --
	(179.35, 89.55) --
	cycle;

\path[draw=drawColor,line width= 1.1pt,line join=round] (179.35, 82.13) -- (191.93, 82.13);
\definecolor{fillColor}{gray}{0.20}

\path[draw=drawColor,line width= 0.4pt,line join=round,line cap=round,fill=fillColor] (199.62, 75.56) circle (  1.96);

\path[draw=drawColor,line width= 0.4pt,line join=round,line cap=round,fill=fillColor] (199.62, 79.02) circle (  1.96);

\path[draw=drawColor,line width= 0.6pt,line join=round] (199.62,116.02) -- (199.62,129.32);

\path[draw=drawColor,line width= 0.6pt,line join=round] (199.62,101.80) -- (199.62,101.32);
\definecolor{fillColor}{RGB}{255,221,227}

\path[draw=drawColor,line width= 0.6pt,line join=round,line cap=round,fill=fillColor] (193.33,116.02) --
	(193.33,101.80) --
	(205.92,101.80) --
	(205.92,116.02) --
	(193.33,116.02) --
	cycle;

\path[draw=drawColor,line width= 1.1pt,line join=round] (193.33,109.83) -- (205.92,109.83);
\definecolor{fillColor}{gray}{0.20}

\path[draw=drawColor,line width= 0.4pt,line join=round,line cap=round,fill=fillColor] (222.93,102.39) circle (  1.96);

\path[draw=drawColor,line width= 0.6pt,line join=round] (222.93,124.88) -- (222.93,132.45);

\path[draw=drawColor,line width= 0.6pt,line join=round] (222.93,116.13) -- (222.93,113.28);
\definecolor{fillColor}{RGB}{254,255,254}

\path[draw=drawColor,line width= 0.6pt,line join=round,line cap=round,fill=fillColor] (216.63,124.88) --
	(216.63,116.13) --
	(229.22,116.13) --
	(229.22,124.88) --
	(216.63,124.88) --
	cycle;

\path[draw=drawColor,line width= 1.1pt,line join=round] (216.63,118.78) -- (229.22,118.78);

\path[draw=drawColor,line width= 0.6pt,line join=round] (236.91,132.61) -- (236.91,135.41);

\path[draw=drawColor,line width= 0.6pt,line join=round] (236.91,128.71) -- (236.91,124.51);
\definecolor{fillColor}{RGB}{255,221,227}

\path[draw=drawColor,line width= 0.6pt,line join=round,line cap=round,fill=fillColor] (230.62,132.61) --
	(230.62,128.71) --
	(243.20,128.71) --
	(243.20,132.61) --
	(230.62,132.61) --
	cycle;

\path[draw=drawColor,line width= 1.1pt,line join=round] (230.62,129.57) -- (243.20,129.57);
\definecolor{fillColor}{gray}{0.20}

\path[draw=drawColor,line width= 0.4pt,line join=round,line cap=round,fill=fillColor] (260.21, 62.19) circle (  1.96);

\path[draw=drawColor,line width= 0.6pt,line join=round] (260.21, 48.46) -- (260.21, 49.45);

\path[draw=drawColor,line width= 0.6pt,line join=round] (260.21, 40.00) -- (260.21, 35.67);
\definecolor{fillColor}{RGB}{254,255,254}

\path[draw=drawColor,line width= 0.6pt,line join=round,line cap=round,fill=fillColor] (253.92, 48.46) --
	(253.92, 40.00) --
	(266.50, 40.00) --
	(266.50, 48.46) --
	(253.92, 48.46) --
	cycle;

\path[draw=drawColor,line width= 1.1pt,line join=round] (253.92, 45.72) -- (266.50, 45.72);

\path[draw=drawColor,line width= 0.6pt,line join=round] (274.19, 63.66) -- (274.19, 75.85);

\path[draw=drawColor,line width= 0.6pt,line join=round] (274.19, 53.77) -- (274.19, 49.75);
\definecolor{fillColor}{RGB}{255,221,227}

\path[draw=drawColor,line width= 0.6pt,line join=round,line cap=round,fill=fillColor] (267.90, 63.66) --
	(267.90, 53.77) --
	(280.48, 53.77) --
	(280.48, 63.66) --
	(267.90, 63.66) --
	cycle;

\path[draw=drawColor,line width= 1.1pt,line join=round] (267.90, 57.58) -- (280.48, 57.58);

\path[draw=drawColor,line width= 0.6pt,line join=round,line cap=round] (170.26, 30.69) rectangle (289.57,140.40);
\end{scope}
\begin{scope}
\path[clip] (308.82,176.25) rectangle (428.12,285.97);
\definecolor{fillColor}{RGB}{255,255,255}

\path[fill=fillColor] (308.82,176.25) rectangle (428.12,285.97);
\definecolor{drawColor}{gray}{0.92}

\path[draw=drawColor,line width= 0.3pt,line join=round] (308.82,187.72) --
	(428.12,187.72);

\path[draw=drawColor,line width= 0.3pt,line join=round] (308.82,206.17) --
	(428.12,206.17);

\path[draw=drawColor,line width= 0.3pt,line join=round] (308.82,224.62) --
	(428.12,224.62);

\path[draw=drawColor,line width= 0.3pt,line join=round] (308.82,243.07) --
	(428.12,243.07);

\path[draw=drawColor,line width= 0.3pt,line join=round] (308.82,261.52) --
	(428.12,261.52);

\path[draw=drawColor,line width= 0.3pt,line join=round] (308.82,279.97) --
	(428.12,279.97);

\path[draw=drawColor,line width= 0.6pt,line join=round] (308.82,178.49) --
	(428.12,178.49);

\path[draw=drawColor,line width= 0.6pt,line join=round] (308.82,196.94) --
	(428.12,196.94);

\path[draw=drawColor,line width= 0.6pt,line join=round] (308.82,215.39) --
	(428.12,215.39);

\path[draw=drawColor,line width= 0.6pt,line join=round] (308.82,233.84) --
	(428.12,233.84);

\path[draw=drawColor,line width= 0.6pt,line join=round] (308.82,252.30) --
	(428.12,252.30);

\path[draw=drawColor,line width= 0.6pt,line join=round] (308.82,270.75) --
	(428.12,270.75);

\path[draw=drawColor,line width= 0.6pt,line join=round] (331.19,176.25) --
	(331.19,285.97);

\path[draw=drawColor,line width= 0.6pt,line join=round] (368.47,176.25) --
	(368.47,285.97);

\path[draw=drawColor,line width= 0.6pt,line join=round] (405.75,176.25) --
	(405.75,285.97);
\definecolor{drawColor}{gray}{0.20}

\path[draw=drawColor,line width= 0.6pt,line join=round] (324.20,233.55) -- (324.20,252.96);

\path[draw=drawColor,line width= 0.6pt,line join=round] (324.20,200.85) -- (324.20,181.24);
\definecolor{fillColor}{RGB}{254,255,254}

\path[draw=drawColor,line width= 0.6pt,line join=round,line cap=round,fill=fillColor] (317.90,233.55) --
	(317.90,200.85) --
	(330.49,200.85) --
	(330.49,233.55) --
	(317.90,233.55) --
	cycle;

\path[draw=drawColor,line width= 1.1pt,line join=round] (317.90,220.12) -- (330.49,220.12);

\path[draw=drawColor,line width= 0.6pt,line join=round] (338.18,271.11) -- (338.18,280.98);

\path[draw=drawColor,line width= 0.6pt,line join=round] (338.18,261.53) -- (338.18,253.71);
\definecolor{fillColor}{RGB}{255,221,227}

\path[draw=drawColor,line width= 0.6pt,line join=round,line cap=round,fill=fillColor] (331.88,271.11) --
	(331.88,261.53) --
	(344.47,261.53) --
	(344.47,271.11) --
	(331.88,271.11) --
	cycle;

\path[draw=drawColor,line width= 1.1pt,line join=round] (331.88,266.00) -- (344.47,266.00);

\path[draw=drawColor,line width= 0.6pt,line join=round] (361.48,243.87) -- (361.48,248.59);

\path[draw=drawColor,line width= 0.6pt,line join=round] (361.48,236.41) -- (361.48,228.24);
\definecolor{fillColor}{RGB}{254,255,254}

\path[draw=drawColor,line width= 0.6pt,line join=round,line cap=round,fill=fillColor] (355.19,243.87) --
	(355.19,236.41) --
	(367.77,236.41) --
	(367.77,243.87) --
	(355.19,243.87) --
	cycle;

\path[draw=drawColor,line width= 1.1pt,line join=round] (355.19,238.92) -- (367.77,238.92);

\path[draw=drawColor,line width= 0.6pt,line join=round] (375.46,264.73) -- (375.46,266.44);

\path[draw=drawColor,line width= 0.6pt,line join=round] (375.46,259.50) -- (375.46,258.22);
\definecolor{fillColor}{RGB}{255,221,227}

\path[draw=drawColor,line width= 0.6pt,line join=round,line cap=round,fill=fillColor] (369.17,264.73) --
	(369.17,259.50) --
	(381.75,259.50) --
	(381.75,264.73) --
	(369.17,264.73) --
	cycle;

\path[draw=drawColor,line width= 1.1pt,line join=round] (369.17,261.35) -- (381.75,261.35);

\path[draw=drawColor,line width= 0.6pt,line join=round] (398.76,202.84) -- (398.76,207.14);

\path[draw=drawColor,line width= 0.6pt,line join=round] (398.76,193.97) -- (398.76,184.57);
\definecolor{fillColor}{RGB}{254,255,254}

\path[draw=drawColor,line width= 0.6pt,line join=round,line cap=round,fill=fillColor] (392.47,202.84) --
	(392.47,193.97) --
	(405.05,193.97) --
	(405.05,202.84) --
	(392.47,202.84) --
	cycle;

\path[draw=drawColor,line width= 1.1pt,line join=round] (392.47,195.69) -- (405.05,195.69);

\path[draw=drawColor,line width= 0.6pt,line join=round] (412.74,241.42) -- (412.74,247.26);

\path[draw=drawColor,line width= 0.6pt,line join=round] (412.74,234.47) -- (412.74,229.92);
\definecolor{fillColor}{RGB}{255,221,227}

\path[draw=drawColor,line width= 0.6pt,line join=round,line cap=round,fill=fillColor] (406.45,241.42) --
	(406.45,234.47) --
	(419.03,234.47) --
	(419.03,241.42) --
	(406.45,241.42) --
	cycle;

\path[draw=drawColor,line width= 1.1pt,line join=round] (406.45,237.39) -- (419.03,237.39);

\path[draw=drawColor,line width= 0.6pt,line join=round,line cap=round] (308.82,176.25) rectangle (428.12,285.97);
\end{scope}
\begin{scope}
\path[clip] ( 31.71,140.40) rectangle (151.02,158.03);
\definecolor{drawColor}{gray}{0.20}
\definecolor{fillColor}{gray}{0.85}

\path[draw=drawColor,line width= 0.6pt,line join=round,line cap=round,fill=fillColor] ( 31.71,140.40) rectangle (151.02,158.03);
\definecolor{drawColor}{gray}{0.10}

\node[text=drawColor,anchor=base,inner sep=0pt, outer sep=0pt, scale=  1.00] at ( 91.36,145.77) {Stocks};
\end{scope}
\begin{scope}
\path[clip] (170.26,140.40) rectangle (289.57,158.03);
\definecolor{drawColor}{gray}{0.20}
\definecolor{fillColor}{gray}{0.85}

\path[draw=drawColor,line width= 0.6pt,line join=round,line cap=round,fill=fillColor] (170.26,140.40) rectangle (289.57,158.03);
\definecolor{drawColor}{gray}{0.10}

\node[text=drawColor,anchor=base,inner sep=0pt, outer sep=0pt, scale=  1.00] at (229.92,145.77) {Average};
\end{scope}
\begin{scope}
\path[clip] ( 31.71,285.97) rectangle (151.02,303.60);
\definecolor{drawColor}{gray}{0.20}
\definecolor{fillColor}{gray}{0.85}

\path[draw=drawColor,line width= 0.6pt,line join=round,line cap=round,fill=fillColor] ( 31.71,285.97) rectangle (151.02,303.60);
\definecolor{drawColor}{gray}{0.10}

\node[text=drawColor,anchor=base,inner sep=0pt, outer sep=0pt, scale=  1.00] at ( 91.36,291.34) {Serial};
\end{scope}
\begin{scope}
\path[clip] (170.26,285.97) rectangle (289.57,303.60);
\definecolor{drawColor}{gray}{0.20}
\definecolor{fillColor}{gray}{0.85}

\path[draw=drawColor,line width= 0.6pt,line join=round,line cap=round,fill=fillColor] (170.26,285.97) rectangle (289.57,303.60);
\definecolor{drawColor}{gray}{0.10}

\node[text=drawColor,anchor=base,inner sep=0pt, outer sep=0pt, scale=  1.00] at (229.92,291.34) {Weather};
\end{scope}
\begin{scope}
\path[clip] (308.82,285.97) rectangle (428.12,303.60);
\definecolor{drawColor}{gray}{0.20}
\definecolor{fillColor}{gray}{0.85}

\path[draw=drawColor,line width= 0.6pt,line join=round,line cap=round,fill=fillColor] (308.82,285.97) rectangle (428.12,303.60);
\definecolor{drawColor}{gray}{0.10}

\node[text=drawColor,anchor=base,inner sep=0pt, outer sep=0pt, scale=  1.00] at (368.47,291.34) {Death notice};
\end{scope}
\begin{scope}
\path[clip] (  0.00,  0.00) rectangle (433.62,325.21);
\definecolor{drawColor}{gray}{0.20}

\path[draw=drawColor,line width= 0.6pt,line join=round] ( 54.08, 27.94) --
	( 54.08, 30.69);

\path[draw=drawColor,line width= 0.6pt,line join=round] ( 91.36, 27.94) --
	( 91.36, 30.69);

\path[draw=drawColor,line width= 0.6pt,line join=round] (128.65, 27.94) --
	(128.65, 30.69);
\end{scope}
\begin{scope}
\path[clip] (  0.00,  0.00) rectangle (433.62,325.21);
\definecolor{drawColor}{gray}{0.30}

\node[text=drawColor,anchor=base,inner sep=0pt, outer sep=0pt, scale=  0.88] at ( 54.08, 19.68) {Time};

\node[text=drawColor,anchor=base,inner sep=0pt, outer sep=0pt, scale=  0.88] at ( 91.36, 19.68) {GDL};

\node[text=drawColor,anchor=base,inner sep=0pt, outer sep=0pt, scale=  0.88] at (128.65, 19.68) {IMP};
\end{scope}
\begin{scope}
\path[clip] (  0.00,  0.00) rectangle (433.62,325.21);
\definecolor{drawColor}{gray}{0.20}

\path[draw=drawColor,line width= 0.6pt,line join=round] (192.63, 27.94) --
	(192.63, 30.69);

\path[draw=drawColor,line width= 0.6pt,line join=round] (229.92, 27.94) --
	(229.92, 30.69);

\path[draw=drawColor,line width= 0.6pt,line join=round] (267.20, 27.94) --
	(267.20, 30.69);
\end{scope}
\begin{scope}
\path[clip] (  0.00,  0.00) rectangle (433.62,325.21);
\definecolor{drawColor}{gray}{0.30}

\node[text=drawColor,anchor=base,inner sep=0pt, outer sep=0pt, scale=  0.88] at (192.63, 19.68) {Time};

\node[text=drawColor,anchor=base,inner sep=0pt, outer sep=0pt, scale=  0.88] at (229.92, 19.68) {GDL};

\node[text=drawColor,anchor=base,inner sep=0pt, outer sep=0pt, scale=  0.88] at (267.20, 19.68) {IMP};
\end{scope}
\begin{scope}
\path[clip] (  0.00,  0.00) rectangle (433.62,325.21);
\definecolor{drawColor}{gray}{0.20}

\path[draw=drawColor,line width= 0.6pt,line join=round] ( 54.08,173.50) --
	( 54.08,176.25);

\path[draw=drawColor,line width= 0.6pt,line join=round] ( 91.36,173.50) --
	( 91.36,176.25);

\path[draw=drawColor,line width= 0.6pt,line join=round] (128.65,173.50) --
	(128.65,176.25);
\end{scope}
\begin{scope}
\path[clip] (  0.00,  0.00) rectangle (433.62,325.21);
\definecolor{drawColor}{gray}{0.30}

\node[text=drawColor,anchor=base,inner sep=0pt, outer sep=0pt, scale=  0.88] at ( 54.08,165.24) {Time};

\node[text=drawColor,anchor=base,inner sep=0pt, outer sep=0pt, scale=  0.88] at ( 91.36,165.24) {GDL};

\node[text=drawColor,anchor=base,inner sep=0pt, outer sep=0pt, scale=  0.88] at (128.65,165.24) {IMP};
\end{scope}
\begin{scope}
\path[clip] (  0.00,  0.00) rectangle (433.62,325.21);
\definecolor{drawColor}{gray}{0.20}

\path[draw=drawColor,line width= 0.6pt,line join=round] (192.63,173.50) --
	(192.63,176.25);

\path[draw=drawColor,line width= 0.6pt,line join=round] (229.92,173.50) --
	(229.92,176.25);

\path[draw=drawColor,line width= 0.6pt,line join=round] (267.20,173.50) --
	(267.20,176.25);
\end{scope}
\begin{scope}
\path[clip] (  0.00,  0.00) rectangle (433.62,325.21);
\definecolor{drawColor}{gray}{0.30}

\node[text=drawColor,anchor=base,inner sep=0pt, outer sep=0pt, scale=  0.88] at (192.63,165.24) {Time};

\node[text=drawColor,anchor=base,inner sep=0pt, outer sep=0pt, scale=  0.88] at (229.92,165.24) {GDL};

\node[text=drawColor,anchor=base,inner sep=0pt, outer sep=0pt, scale=  0.88] at (267.20,165.24) {IMP};
\end{scope}
\begin{scope}
\path[clip] (  0.00,  0.00) rectangle (433.62,325.21);
\definecolor{drawColor}{gray}{0.20}

\path[draw=drawColor,line width= 0.6pt,line join=round] (331.19,173.50) --
	(331.19,176.25);

\path[draw=drawColor,line width= 0.6pt,line join=round] (368.47,173.50) --
	(368.47,176.25);

\path[draw=drawColor,line width= 0.6pt,line join=round] (405.75,173.50) --
	(405.75,176.25);
\end{scope}
\begin{scope}
\path[clip] (  0.00,  0.00) rectangle (433.62,325.21);
\definecolor{drawColor}{gray}{0.30}

\node[text=drawColor,anchor=base,inner sep=0pt, outer sep=0pt, scale=  0.88] at (331.19,165.24) {Time};

\node[text=drawColor,anchor=base,inner sep=0pt, outer sep=0pt, scale=  0.88] at (368.47,165.24) {GDL};

\node[text=drawColor,anchor=base,inner sep=0pt, outer sep=0pt, scale=  0.88] at (405.75,165.24) {IMP};
\end{scope}
\begin{scope}
\path[clip] (  0.00,  0.00) rectangle (433.62,325.21);
\definecolor{drawColor}{gray}{0.30}

\node[text=drawColor,anchor=base east,inner sep=0pt, outer sep=0pt, scale=  0.88] at (303.87,175.46) {30};

\node[text=drawColor,anchor=base east,inner sep=0pt, outer sep=0pt, scale=  0.88] at (303.87,193.91) {40};

\node[text=drawColor,anchor=base east,inner sep=0pt, outer sep=0pt, scale=  0.88] at (303.87,212.36) {50};

\node[text=drawColor,anchor=base east,inner sep=0pt, outer sep=0pt, scale=  0.88] at (303.87,230.81) {60};

\node[text=drawColor,anchor=base east,inner sep=0pt, outer sep=0pt, scale=  0.88] at (303.87,249.27) {70};

\node[text=drawColor,anchor=base east,inner sep=0pt, outer sep=0pt, scale=  0.88] at (303.87,267.72) {80};
\end{scope}
\begin{scope}
\path[clip] (  0.00,  0.00) rectangle (433.62,325.21);
\definecolor{drawColor}{gray}{0.20}

\path[draw=drawColor,line width= 0.6pt,line join=round] (306.07,178.49) --
	(308.82,178.49);

\path[draw=drawColor,line width= 0.6pt,line join=round] (306.07,196.94) --
	(308.82,196.94);

\path[draw=drawColor,line width= 0.6pt,line join=round] (306.07,215.39) --
	(308.82,215.39);

\path[draw=drawColor,line width= 0.6pt,line join=round] (306.07,233.84) --
	(308.82,233.84);

\path[draw=drawColor,line width= 0.6pt,line join=round] (306.07,252.30) --
	(308.82,252.30);

\path[draw=drawColor,line width= 0.6pt,line join=round] (306.07,270.75) --
	(308.82,270.75);
\end{scope}
\begin{scope}
\path[clip] (  0.00,  0.00) rectangle (433.62,325.21);
\definecolor{drawColor}{gray}{0.30}

\node[text=drawColor,anchor=base east,inner sep=0pt, outer sep=0pt, scale=  0.88] at (165.31,177.64) {0};

\node[text=drawColor,anchor=base east,inner sep=0pt, outer sep=0pt, scale=  0.88] at (165.31,206.49) {20};

\node[text=drawColor,anchor=base east,inner sep=0pt, outer sep=0pt, scale=  0.88] at (165.31,235.35) {40};

\node[text=drawColor,anchor=base east,inner sep=0pt, outer sep=0pt, scale=  0.88] at (165.31,264.20) {60};
\end{scope}
\begin{scope}
\path[clip] (  0.00,  0.00) rectangle (433.62,325.21);
\definecolor{drawColor}{gray}{0.20}

\path[draw=drawColor,line width= 0.6pt,line join=round] (167.51,180.67) --
	(170.26,180.67);

\path[draw=drawColor,line width= 0.6pt,line join=round] (167.51,209.52) --
	(170.26,209.52);

\path[draw=drawColor,line width= 0.6pt,line join=round] (167.51,238.38) --
	(170.26,238.38);

\path[draw=drawColor,line width= 0.6pt,line join=round] (167.51,267.23) --
	(170.26,267.23);
\end{scope}
\begin{scope}
\path[clip] (  0.00,  0.00) rectangle (433.62,325.21);
\definecolor{drawColor}{gray}{0.30}

\node[text=drawColor,anchor=base east,inner sep=0pt, outer sep=0pt, scale=  0.88] at (165.31, 54.10) {45};

\node[text=drawColor,anchor=base east,inner sep=0pt, outer sep=0pt, scale=  0.88] at (165.31, 81.68) {55};

\node[text=drawColor,anchor=base east,inner sep=0pt, outer sep=0pt, scale=  0.88] at (165.31,109.25) {65};

\node[text=drawColor,anchor=base east,inner sep=0pt, outer sep=0pt, scale=  0.88] at (165.31,136.83) {75};
\end{scope}
\begin{scope}
\path[clip] (  0.00,  0.00) rectangle (433.62,325.21);
\definecolor{drawColor}{gray}{0.20}

\path[draw=drawColor,line width= 0.6pt,line join=round] (167.51, 57.14) --
	(170.26, 57.14);

\path[draw=drawColor,line width= 0.6pt,line join=round] (167.51, 84.71) --
	(170.26, 84.71);

\path[draw=drawColor,line width= 0.6pt,line join=round] (167.51,112.28) --
	(170.26,112.28);

\path[draw=drawColor,line width= 0.6pt,line join=round] (167.51,139.86) --
	(170.26,139.86);
\end{scope}
\begin{scope}
\path[clip] (  0.00,  0.00) rectangle (433.62,325.21);
\definecolor{drawColor}{gray}{0.30}

\node[text=drawColor,anchor=base east,inner sep=0pt, outer sep=0pt, scale=  0.88] at ( 26.76,174.25) {0};

\node[text=drawColor,anchor=base east,inner sep=0pt, outer sep=0pt, scale=  0.88] at ( 26.76,200.48) {20};

\node[text=drawColor,anchor=base east,inner sep=0pt, outer sep=0pt, scale=  0.88] at ( 26.76,226.70) {40};

\node[text=drawColor,anchor=base east,inner sep=0pt, outer sep=0pt, scale=  0.88] at ( 26.76,252.93) {60};

\node[text=drawColor,anchor=base east,inner sep=0pt, outer sep=0pt, scale=  0.88] at ( 26.76,279.16) {80};
\end{scope}
\begin{scope}
\path[clip] (  0.00,  0.00) rectangle (433.62,325.21);
\definecolor{drawColor}{gray}{0.20}

\path[draw=drawColor,line width= 0.6pt,line join=round] ( 28.96,177.28) --
	( 31.71,177.28);

\path[draw=drawColor,line width= 0.6pt,line join=round] ( 28.96,203.51) --
	( 31.71,203.51);

\path[draw=drawColor,line width= 0.6pt,line join=round] ( 28.96,229.73) --
	( 31.71,229.73);

\path[draw=drawColor,line width= 0.6pt,line join=round] ( 28.96,255.96) --
	( 31.71,255.96);

\path[draw=drawColor,line width= 0.6pt,line join=round] ( 28.96,282.19) --
	( 31.71,282.19);
\end{scope}
\begin{scope}
\path[clip] (  0.00,  0.00) rectangle (433.62,325.21);
\definecolor{drawColor}{gray}{0.30}

\node[text=drawColor,anchor=base east,inner sep=0pt, outer sep=0pt, scale=  0.88] at ( 26.76, 45.36) {40};

\node[text=drawColor,anchor=base east,inner sep=0pt, outer sep=0pt, scale=  0.88] at ( 26.76, 69.67) {50};

\node[text=drawColor,anchor=base east,inner sep=0pt, outer sep=0pt, scale=  0.88] at ( 26.76, 93.99) {60};

\node[text=drawColor,anchor=base east,inner sep=0pt, outer sep=0pt, scale=  0.88] at ( 26.76,118.31) {70};
\end{scope}
\begin{scope}
\path[clip] (  0.00,  0.00) rectangle (433.62,325.21);
\definecolor{drawColor}{gray}{0.20}

\path[draw=drawColor,line width= 0.6pt,line join=round] ( 28.96, 48.39) --
	( 31.71, 48.39);

\path[draw=drawColor,line width= 0.6pt,line join=round] ( 28.96, 72.70) --
	( 31.71, 72.70);

\path[draw=drawColor,line width= 0.6pt,line join=round] ( 28.96, 97.02) --
	( 31.71, 97.02);

\path[draw=drawColor,line width= 0.6pt,line join=round] ( 28.96,121.34) --
	( 31.71,121.34);
\end{scope}
\begin{scope}
\path[clip] (  0.00,  0.00) rectangle (433.62,325.21);
\definecolor{drawColor}{RGB}{0,0,0}

\node[text=drawColor,anchor=base,inner sep=0pt, outer sep=0pt, scale=  1.10] at (229.92,  7.64) {Experiment};
\end{scope}
\begin{scope}
\path[clip] (  0.00,  0.00) rectangle (433.62,325.21);
\definecolor{drawColor}{RGB}{0,0,0}

\node[text=drawColor,rotate= 90.00,anchor=base,inner sep=0pt, outer sep=0pt, scale=  1.10] at ( 13.08,158.33) {mIoU};
\end{scope}
\begin{scope}
\path[clip] (  0.00,  0.00) rectangle (433.62,325.21);
\definecolor{fillColor}{RGB}{255,255,255}

\path[fill=fillColor] (320.65, 49.45) rectangle (400.26,124.53);
\end{scope}
\begin{scope}
\path[clip] (  0.00,  0.00) rectangle (433.62,325.21);
\definecolor{drawColor}{RGB}{0,0,0}

\node[text=drawColor,anchor=base,inner sep=0pt, outer sep=0pt, scale=  1.10] at (360.45,110.39) {Modalities};
\end{scope}
\begin{scope}
\path[clip] (  0.00,  0.00) rectangle (433.62,325.21);
\definecolor{fillColor}{RGB}{255,255,255}

\path[fill=fillColor] (326.15, 82.14) rectangle (340.60,103.82);
\end{scope}
\begin{scope}
\path[clip] (  0.00,  0.00) rectangle (433.62,325.21);
\definecolor{drawColor}{gray}{0.20}

\path[draw=drawColor,line width= 0.6pt,line join=round,line cap=round] (333.37, 84.30) --
	(333.37, 87.56);

\path[draw=drawColor,line width= 0.6pt,line join=round,line cap=round] (333.37, 98.40) --
	(333.37,101.65);
\definecolor{fillColor}{RGB}{254,255,254}

\path[draw=drawColor,line width= 0.6pt,line join=round,line cap=round,fill=fillColor] (327.95, 87.56) rectangle (338.80, 98.40);

\path[draw=drawColor,line width= 0.6pt,line join=round,line cap=round] (327.95, 92.98) --
	(338.80, 92.98);
\end{scope}
\begin{scope}
\path[clip] (  0.00,  0.00) rectangle (433.62,325.21);
\definecolor{fillColor}{RGB}{255,255,255}

\path[fill=fillColor] (326.15, 54.95) rectangle (340.60, 76.64);
\end{scope}
\begin{scope}
\path[clip] (  0.00,  0.00) rectangle (433.62,325.21);
\definecolor{drawColor}{gray}{0.20}

\path[draw=drawColor,line width= 0.6pt,line join=round,line cap=round] (333.37, 57.12) --
	(333.37, 60.38);

\path[draw=drawColor,line width= 0.6pt,line join=round,line cap=round] (333.37, 71.22) --
	(333.37, 74.47);
\definecolor{fillColor}{RGB}{255,221,227}

\path[draw=drawColor,line width= 0.6pt,line join=round,line cap=round,fill=fillColor] (327.95, 60.38) rectangle (338.80, 71.22);

\path[draw=drawColor,line width= 0.6pt,line join=round,line cap=round] (327.95, 65.80) --
	(338.80, 65.80);
\end{scope}
\begin{scope}
\path[clip] (  0.00,  0.00) rectangle (433.62,325.21);
\definecolor{drawColor}{RGB}{0,0,0}

\node[text=drawColor,anchor=base west,inner sep=0pt, outer sep=0pt, scale=  0.80] at (346.10, 90.22) {Image};
\end{scope}
\begin{scope}
\path[clip] (  0.00,  0.00) rectangle (433.62,325.21);
\definecolor{drawColor}{RGB}{0,0,0}

\node[text=drawColor,anchor=base west,inner sep=0pt, outer sep=0pt, scale=  0.80] at (346.10, 63.04) {Image + Text};
\end{scope}
\begin{scope}
\path[clip] (  0.00,  0.00) rectangle (433.62,325.21);
\definecolor{drawColor}{RGB}{0,0,0}

\node[text=drawColor,anchor=base,inner sep=0pt, outer sep=0pt, scale=  1.20] at (229.92,311.43) {\bfseries Comparison of mIoU per experiment};
\end{scope}
\end{tikzpicture}

%% file: Figures/JDG_data.tex
\begin{tikzpicture}[x=1pt,y=1pt]
\definecolor{fillColor}{RGB}{255,255,255}
\path[use as bounding box,fill=fillColor,fill opacity=0.00] (0,0) rectangle (433.62,325.21);
\begin{scope}
\path[clip] (  0.00,  0.00) rectangle (433.62,325.21);
\definecolor{drawColor}{RGB}{255,255,255}
\definecolor{fillColor}{RGB}{255,255,255}

\path[draw=drawColor,line width= 0.6pt,line join=round,line cap=round,fill=fillColor] (  0.00,  0.00) rectangle (433.62,325.21);
\end{scope}
\begin{scope}
\path[clip] ( 31.71,176.25) rectangle (151.02,285.97);
\definecolor{fillColor}{RGB}{255,255,255}

\path[fill=fillColor] ( 31.71,176.25) rectangle (151.02,285.97);
\definecolor{drawColor}{gray}{0.92}

\path[draw=drawColor,line width= 0.3pt,line join=round] ( 31.71,189.10) --
	(151.02,189.10);

\path[draw=drawColor,line width= 0.3pt,line join=round] ( 31.71,223.33) --
	(151.02,223.33);

\path[draw=drawColor,line width= 0.3pt,line join=round] ( 31.71,257.57) --
	(151.02,257.57);

\path[draw=drawColor,line width= 0.6pt,line join=round] ( 31.71,206.22) --
	(151.02,206.22);

\path[draw=drawColor,line width= 0.6pt,line join=round] ( 31.71,240.45) --
	(151.02,240.45);

\path[draw=drawColor,line width= 0.6pt,line join=round] ( 31.71,274.68) --
	(151.02,274.68);

\path[draw=drawColor,line width= 0.6pt,line join=round] ( 64.25,176.25) --
	( 64.25,285.97);

\path[draw=drawColor,line width= 0.6pt,line join=round] (118.48,176.25) --
	(118.48,285.97);
\definecolor{drawColor}{gray}{0.20}

\path[draw=drawColor,line width= 0.6pt,line join=round] ( 54.08,274.00) -- ( 54.08,280.98);

\path[draw=drawColor,line width= 0.6pt,line join=round] ( 54.08,257.87) -- ( 54.08,241.62);
\definecolor{fillColor}{RGB}{254,255,254}

\path[draw=drawColor,line width= 0.6pt,line join=round,line cap=round,fill=fillColor] ( 44.93,274.00) --
	( 44.93,257.87) --
	( 63.23,257.87) --
	( 63.23,274.00) --
	( 44.93,274.00) --
	cycle;

\path[draw=drawColor,line width= 1.1pt,line join=round] ( 44.93,265.67) -- ( 63.23,265.67);

\path[draw=drawColor,line width= 0.6pt,line join=round] ( 74.42,277.25) -- ( 74.42,280.80);

\path[draw=drawColor,line width= 0.6pt,line join=round] ( 74.42,260.77) -- ( 74.42,251.60);
\definecolor{fillColor}{RGB}{255,221,227}

\path[draw=drawColor,line width= 0.6pt,line join=round,line cap=round,fill=fillColor] ( 65.27,277.25) --
	( 65.27,260.77) --
	( 83.57,260.77) --
	( 83.57,277.25) --
	( 65.27,277.25) --
	cycle;

\path[draw=drawColor,line width= 1.1pt,line join=round] ( 65.27,271.01) -- ( 83.57,271.01);

\path[draw=drawColor,line width= 0.6pt,line join=round] (108.31,261.02) -- (108.31,263.51);

\path[draw=drawColor,line width= 0.6pt,line join=round] (108.31,255.36) -- (108.31,247.75);
\definecolor{fillColor}{RGB}{254,255,254}

\path[draw=drawColor,line width= 0.6pt,line join=round,line cap=round,fill=fillColor] ( 99.16,261.02) --
	( 99.16,255.36) --
	(117.46,255.36) --
	(117.46,261.02) --
	( 99.16,261.02) --
	cycle;

\path[draw=drawColor,line width= 1.1pt,line join=round] ( 99.16,259.38) -- (117.46,259.38);

\path[draw=drawColor,line width= 0.6pt,line join=round] (128.65,240.16) -- (128.65,242.53);

\path[draw=drawColor,line width= 0.6pt,line join=round] (128.65,202.34) -- (128.65,181.24);
\definecolor{fillColor}{RGB}{255,221,227}

\path[draw=drawColor,line width= 0.6pt,line join=round,line cap=round,fill=fillColor] (119.50,240.16) --
	(119.50,202.34) --
	(137.80,202.34) --
	(137.80,240.16) --
	(119.50,240.16) --
	cycle;

\path[draw=drawColor,line width= 1.1pt,line join=round] (119.50,233.51) -- (137.80,233.51);

\path[draw=drawColor,line width= 0.6pt,line join=round,line cap=round] ( 31.71,176.25) rectangle (151.02,285.97);
\end{scope}
\begin{scope}
\path[clip] ( 31.71, 30.69) rectangle (151.02,140.40);
\definecolor{fillColor}{RGB}{255,255,255}

\path[fill=fillColor] ( 31.71, 30.69) rectangle (151.02,140.40);
\definecolor{drawColor}{gray}{0.92}

\path[draw=drawColor,line width= 0.3pt,line join=round] ( 31.71, 46.24) --
	(151.02, 46.24);

\path[draw=drawColor,line width= 0.3pt,line join=round] ( 31.71, 77.33) --
	(151.02, 77.33);

\path[draw=drawColor,line width= 0.3pt,line join=round] ( 31.71,108.43) --
	(151.02,108.43);

\path[draw=drawColor,line width= 0.3pt,line join=round] ( 31.71,139.52) --
	(151.02,139.52);

\path[draw=drawColor,line width= 0.6pt,line join=round] ( 31.71, 30.69) --
	(151.02, 30.69);

\path[draw=drawColor,line width= 0.6pt,line join=round] ( 31.71, 61.78) --
	(151.02, 61.78);

\path[draw=drawColor,line width= 0.6pt,line join=round] ( 31.71, 92.88) --
	(151.02, 92.88);

\path[draw=drawColor,line width= 0.6pt,line join=round] ( 31.71,123.97) --
	(151.02,123.97);

\path[draw=drawColor,line width= 0.6pt,line join=round] ( 64.25, 30.69) --
	( 64.25,140.40);

\path[draw=drawColor,line width= 0.6pt,line join=round] (118.48, 30.69) --
	(118.48,140.40);
\definecolor{drawColor}{gray}{0.20}

\path[draw=drawColor,line width= 0.6pt,line join=round] ( 54.08,116.30) -- ( 54.08,121.72);

\path[draw=drawColor,line width= 0.6pt,line join=round] ( 54.08,107.57) -- ( 54.08,104.93);
\definecolor{fillColor}{RGB}{254,255,254}

\path[draw=drawColor,line width= 0.6pt,line join=round,line cap=round,fill=fillColor] ( 44.93,116.30) --
	( 44.93,107.57) --
	( 63.23,107.57) --
	( 63.23,116.30) --
	( 44.93,116.30) --
	cycle;

\path[draw=drawColor,line width= 1.1pt,line join=round] ( 44.93,112.06) -- ( 63.23,112.06);

\path[draw=drawColor,line width= 0.6pt,line join=round] ( 74.42,129.86) -- ( 74.42,135.41);

\path[draw=drawColor,line width= 0.6pt,line join=round] ( 74.42,111.71) -- ( 74.42,103.02);
\definecolor{fillColor}{RGB}{255,221,227}

\path[draw=drawColor,line width= 0.6pt,line join=round,line cap=round,fill=fillColor] ( 65.27,129.86) --
	( 65.27,111.71) --
	( 83.57,111.71) --
	( 83.57,129.86) --
	( 65.27,129.86) --
	cycle;

\path[draw=drawColor,line width= 1.1pt,line join=round] ( 65.27,120.87) -- ( 83.57,120.87);

\path[draw=drawColor,line width= 0.6pt,line join=round] (108.31, 70.57) -- (108.31, 76.44);

\path[draw=drawColor,line width= 0.6pt,line join=round] (108.31, 49.58) -- (108.31, 35.67);
\definecolor{fillColor}{RGB}{254,255,254}

\path[draw=drawColor,line width= 0.6pt,line join=round,line cap=round,fill=fillColor] ( 99.16, 70.57) --
	( 99.16, 49.58) --
	(117.46, 49.58) --
	(117.46, 70.57) --
	( 99.16, 70.57) --
	cycle;

\path[draw=drawColor,line width= 1.1pt,line join=round] ( 99.16, 62.57) -- (117.46, 62.57);

\path[draw=drawColor,line width= 0.6pt,line join=round] (128.65, 79.39) -- (128.65, 85.07);

\path[draw=drawColor,line width= 0.6pt,line join=round] (128.65, 71.64) -- (128.65, 63.68);
\definecolor{fillColor}{RGB}{255,221,227}

\path[draw=drawColor,line width= 0.6pt,line join=round,line cap=round,fill=fillColor] (119.50, 79.39) --
	(119.50, 71.64) --
	(137.80, 71.64) --
	(137.80, 79.39) --
	(119.50, 79.39) --
	cycle;

\path[draw=drawColor,line width= 1.1pt,line join=round] (119.50, 77.12) -- (137.80, 77.12);

\path[draw=drawColor,line width= 0.6pt,line join=round,line cap=round] ( 31.71, 30.69) rectangle (151.02,140.40);
\end{scope}
\begin{scope}
\path[clip] (170.26,176.25) rectangle (289.57,285.97);
\definecolor{fillColor}{RGB}{255,255,255}

\path[fill=fillColor] (170.26,176.25) rectangle (289.57,285.97);
\definecolor{drawColor}{gray}{0.92}

\path[draw=drawColor,line width= 0.3pt,line join=round] (170.26,194.23) --
	(289.57,194.23);

\path[draw=drawColor,line width= 0.3pt,line join=round] (170.26,213.60) --
	(289.57,213.60);

\path[draw=drawColor,line width= 0.3pt,line join=round] (170.26,232.97) --
	(289.57,232.97);

\path[draw=drawColor,line width= 0.3pt,line join=round] (170.26,252.34) --
	(289.57,252.34);

\path[draw=drawColor,line width= 0.3pt,line join=round] (170.26,271.71) --
	(289.57,271.71);

\path[draw=drawColor,line width= 0.6pt,line join=round] (170.26,184.55) --
	(289.57,184.55);

\path[draw=drawColor,line width= 0.6pt,line join=round] (170.26,203.92) --
	(289.57,203.92);

\path[draw=drawColor,line width= 0.6pt,line join=round] (170.26,223.29) --
	(289.57,223.29);

\path[draw=drawColor,line width= 0.6pt,line join=round] (170.26,242.65) --
	(289.57,242.65);

\path[draw=drawColor,line width= 0.6pt,line join=round] (170.26,262.02) --
	(289.57,262.02);

\path[draw=drawColor,line width= 0.6pt,line join=round] (170.26,281.39) --
	(289.57,281.39);

\path[draw=drawColor,line width= 0.6pt,line join=round] (202.80,176.25) --
	(202.80,285.97);

\path[draw=drawColor,line width= 0.6pt,line join=round] (257.03,176.25) --
	(257.03,285.97);
\definecolor{drawColor}{gray}{0.20}
\definecolor{fillColor}{gray}{0.20}

\path[draw=drawColor,line width= 0.4pt,line join=round,line cap=round,fill=fillColor] (192.63,246.38) circle (  1.96);

\path[draw=drawColor,line width= 0.6pt,line join=round] (192.63,270.89) -- (192.63,279.32);

\path[draw=drawColor,line width= 0.6pt,line join=round] (192.63,264.85) -- (192.63,264.00);
\definecolor{fillColor}{RGB}{254,255,254}

\path[draw=drawColor,line width= 0.6pt,line join=round,line cap=round,fill=fillColor] (183.48,270.89) --
	(183.48,264.85) --
	(201.78,264.85) --
	(201.78,270.89) --
	(183.48,270.89) --
	cycle;

\path[draw=drawColor,line width= 1.1pt,line join=round] (183.48,268.23) -- (201.78,268.23);
\definecolor{fillColor}{gray}{0.20}

\path[draw=drawColor,line width= 0.4pt,line join=round,line cap=round,fill=fillColor] (212.97,236.60) circle (  1.96);

\path[draw=drawColor,line width= 0.6pt,line join=round] (212.97,275.28) -- (212.97,280.98);

\path[draw=drawColor,line width= 0.6pt,line join=round] (212.97,262.16) -- (212.97,258.33);
\definecolor{fillColor}{RGB}{255,221,227}

\path[draw=drawColor,line width= 0.6pt,line join=round,line cap=round,fill=fillColor] (203.82,275.28) --
	(203.82,262.16) --
	(222.12,262.16) --
	(222.12,275.28) --
	(203.82,275.28) --
	cycle;

\path[draw=drawColor,line width= 1.1pt,line join=round] (203.82,271.42) -- (222.12,271.42);

\path[draw=drawColor,line width= 0.6pt,line join=round] (246.86,230.43) -- (246.86,239.73);

\path[draw=drawColor,line width= 0.6pt,line join=round] (246.86,214.32) -- (246.86,206.67);
\definecolor{fillColor}{RGB}{254,255,254}

\path[draw=drawColor,line width= 0.6pt,line join=round,line cap=round,fill=fillColor] (237.71,230.43) --
	(237.71,214.32) --
	(256.01,214.32) --
	(256.01,230.43) --
	(237.71,230.43) --
	cycle;

\path[draw=drawColor,line width= 1.1pt,line join=round] (237.71,219.19) -- (256.01,219.19);
\definecolor{fillColor}{gray}{0.20}

\path[draw=drawColor,line width= 0.4pt,line join=round,line cap=round,fill=fillColor] (267.20,181.24) circle (  1.96);

\path[draw=drawColor,line width= 0.6pt,line join=round] (267.20,218.33) -- (267.20,224.90);

\path[draw=drawColor,line width= 0.6pt,line join=round] (267.20,205.42) -- (267.20,194.48);
\definecolor{fillColor}{RGB}{255,221,227}

\path[draw=drawColor,line width= 0.6pt,line join=round,line cap=round,fill=fillColor] (258.05,218.33) --
	(258.05,205.42) --
	(276.35,205.42) --
	(276.35,218.33) --
	(258.05,218.33) --
	cycle;

\path[draw=drawColor,line width= 1.1pt,line join=round] (258.05,215.06) -- (276.35,215.06);

\path[draw=drawColor,line width= 0.6pt,line join=round,line cap=round] (170.26,176.25) rectangle (289.57,285.97);
\end{scope}
\begin{scope}
\path[clip] (170.26, 30.69) rectangle (289.57,140.40);
\definecolor{fillColor}{RGB}{255,255,255}

\path[fill=fillColor] (170.26, 30.69) rectangle (289.57,140.40);
\definecolor{drawColor}{gray}{0.92}

\path[draw=drawColor,line width= 0.3pt,line join=round] (170.26, 52.48) --
	(289.57, 52.48);

\path[draw=drawColor,line width= 0.3pt,line join=round] (170.26, 75.49) --
	(289.57, 75.49);

\path[draw=drawColor,line width= 0.3pt,line join=round] (170.26, 98.50) --
	(289.57, 98.50);

\path[draw=drawColor,line width= 0.3pt,line join=round] (170.26,121.51) --
	(289.57,121.51);

\path[draw=drawColor,line width= 0.6pt,line join=round] (170.26, 40.98) --
	(289.57, 40.98);

\path[draw=drawColor,line width= 0.6pt,line join=round] (170.26, 63.99) --
	(289.57, 63.99);

\path[draw=drawColor,line width= 0.6pt,line join=round] (170.26, 87.00) --
	(289.57, 87.00);

\path[draw=drawColor,line width= 0.6pt,line join=round] (170.26,110.00) --
	(289.57,110.00);

\path[draw=drawColor,line width= 0.6pt,line join=round] (170.26,133.01) --
	(289.57,133.01);

\path[draw=drawColor,line width= 0.6pt,line join=round] (202.80, 30.69) --
	(202.80,140.40);

\path[draw=drawColor,line width= 0.6pt,line join=round] (257.03, 30.69) --
	(257.03,140.40);
\definecolor{drawColor}{gray}{0.20}

\path[draw=drawColor,line width= 0.6pt,line join=round] (192.63,113.92) -- (192.63,118.74);

\path[draw=drawColor,line width= 0.6pt,line join=round] (192.63,101.84) -- (192.63, 88.07);
\definecolor{fillColor}{RGB}{254,255,254}

\path[draw=drawColor,line width= 0.6pt,line join=round,line cap=round,fill=fillColor] (183.48,113.92) --
	(183.48,101.84) --
	(201.78,101.84) --
	(201.78,113.92) --
	(183.48,113.92) --
	cycle;

\path[draw=drawColor,line width= 1.1pt,line join=round] (183.48,110.70) -- (201.78,110.70);

\path[draw=drawColor,line width= 0.6pt,line join=round] (212.97,125.29) -- (212.97,135.41);

\path[draw=drawColor,line width= 0.6pt,line join=round] (212.97,114.18) -- (212.97,110.65);
\definecolor{fillColor}{RGB}{255,221,227}

\path[draw=drawColor,line width= 0.6pt,line join=round,line cap=round,fill=fillColor] (203.82,125.29) --
	(203.82,114.18) --
	(222.12,114.18) --
	(222.12,125.29) --
	(203.82,125.29) --
	cycle;

\path[draw=drawColor,line width= 1.1pt,line join=round] (203.82,118.40) -- (222.12,118.40);
\definecolor{fillColor}{gray}{0.20}

\path[draw=drawColor,line width= 0.4pt,line join=round,line cap=round,fill=fillColor] (246.86, 44.22) circle (  1.96);

\path[draw=drawColor,line width= 0.6pt,line join=round] (246.86, 75.89) -- (246.86, 80.06);

\path[draw=drawColor,line width= 0.6pt,line join=round] (246.86, 64.55) -- (246.86, 58.45);
\definecolor{fillColor}{RGB}{254,255,254}

\path[draw=drawColor,line width= 0.6pt,line join=round,line cap=round,fill=fillColor] (237.71, 75.89) --
	(237.71, 64.55) --
	(256.01, 64.55) --
	(256.01, 75.89) --
	(237.71, 75.89) --
	cycle;

\path[draw=drawColor,line width= 1.1pt,line join=round] (237.71, 68.39) -- (256.01, 68.39);

\path[draw=drawColor,line width= 0.6pt,line join=round] (267.20, 62.61) -- (267.20, 72.66);

\path[draw=drawColor,line width= 0.6pt,line join=round] (267.20, 44.81) -- (267.20, 35.67);
\definecolor{fillColor}{RGB}{255,221,227}

\path[draw=drawColor,line width= 0.6pt,line join=round,line cap=round,fill=fillColor] (258.05, 62.61) --
	(258.05, 44.81) --
	(276.35, 44.81) --
	(276.35, 62.61) --
	(258.05, 62.61) --
	cycle;

\path[draw=drawColor,line width= 1.1pt,line join=round] (258.05, 59.40) -- (276.35, 59.40);

\path[draw=drawColor,line width= 0.6pt,line join=round,line cap=round] (170.26, 30.69) rectangle (289.57,140.40);
\end{scope}
\begin{scope}
\path[clip] (308.82,176.25) rectangle (428.12,285.97);
\definecolor{fillColor}{RGB}{255,255,255}

\path[fill=fillColor] (308.82,176.25) rectangle (428.12,285.97);
\definecolor{drawColor}{gray}{0.92}

\path[draw=drawColor,line width= 0.3pt,line join=round] (308.82,198.13) --
	(428.12,198.13);

\path[draw=drawColor,line width= 0.3pt,line join=round] (308.82,224.29) --
	(428.12,224.29);

\path[draw=drawColor,line width= 0.3pt,line join=round] (308.82,250.46) --
	(428.12,250.46);

\path[draw=drawColor,line width= 0.3pt,line join=round] (308.82,276.62) --
	(428.12,276.62);

\path[draw=drawColor,line width= 0.6pt,line join=round] (308.82,185.05) --
	(428.12,185.05);

\path[draw=drawColor,line width= 0.6pt,line join=round] (308.82,211.21) --
	(428.12,211.21);

\path[draw=drawColor,line width= 0.6pt,line join=round] (308.82,237.38) --
	(428.12,237.38);

\path[draw=drawColor,line width= 0.6pt,line join=round] (308.82,263.54) --
	(428.12,263.54);

\path[draw=drawColor,line width= 0.6pt,line join=round] (341.35,176.25) --
	(341.35,285.97);

\path[draw=drawColor,line width= 0.6pt,line join=round] (395.58,176.25) --
	(395.58,285.97);
\definecolor{drawColor}{gray}{0.20}

\path[draw=drawColor,line width= 0.6pt,line join=round] (331.19,255.92) -- (331.19,263.56);

\path[draw=drawColor,line width= 0.6pt,line join=round] (331.19,245.76) -- (331.19,240.00);
\definecolor{fillColor}{RGB}{254,255,254}

\path[draw=drawColor,line width= 0.6pt,line join=round,line cap=round,fill=fillColor] (322.03,255.92) --
	(322.03,245.76) --
	(340.34,245.76) --
	(340.34,255.92) --
	(322.03,255.92) --
	cycle;

\path[draw=drawColor,line width= 1.1pt,line join=round] (322.03,251.62) -- (340.34,251.62);

\path[draw=drawColor,line width= 0.6pt,line join=round] (351.52,277.90) -- (351.52,280.98);

\path[draw=drawColor,line width= 0.6pt,line join=round] (351.52,268.96) -- (351.52,265.68);
\definecolor{fillColor}{RGB}{255,221,227}

\path[draw=drawColor,line width= 0.6pt,line join=round,line cap=round,fill=fillColor] (342.37,277.90) --
	(342.37,268.96) --
	(360.67,268.96) --
	(360.67,277.90) --
	(342.37,277.90) --
	cycle;

\path[draw=drawColor,line width= 1.1pt,line join=round] (342.37,271.29) -- (360.67,271.29);
\definecolor{fillColor}{gray}{0.20}

\path[draw=drawColor,line width= 0.4pt,line join=round,line cap=round,fill=fillColor] (385.41,181.24) circle (  1.96);

\path[draw=drawColor,line width= 0.6pt,line join=round] (385.41,233.54) -- (385.41,241.37);

\path[draw=drawColor,line width= 0.6pt,line join=round] (385.41,224.94) -- (385.41,223.20);
\definecolor{fillColor}{RGB}{254,255,254}

\path[draw=drawColor,line width= 0.6pt,line join=round,line cap=round,fill=fillColor] (376.26,233.54) --
	(376.26,224.94) --
	(394.57,224.94) --
	(394.57,233.54) --
	(376.26,233.54) --
	cycle;

\path[draw=drawColor,line width= 1.1pt,line join=round] (376.26,231.39) -- (394.57,231.39);

\path[draw=drawColor,line width= 0.6pt,line join=round] (405.75,245.75) -- (405.75,259.67);

\path[draw=drawColor,line width= 0.6pt,line join=round] (405.75,234.24) -- (405.75,220.20);
\definecolor{fillColor}{RGB}{255,221,227}

\path[draw=drawColor,line width= 0.6pt,line join=round,line cap=round,fill=fillColor] (396.60,245.75) --
	(396.60,234.24) --
	(414.90,234.24) --
	(414.90,245.75) --
	(396.60,245.75) --
	cycle;

\path[draw=drawColor,line width= 1.1pt,line join=round] (396.60,242.69) -- (414.90,242.69);

\path[draw=drawColor,line width= 0.6pt,line join=round,line cap=round] (308.82,176.25) rectangle (428.12,285.97);
\end{scope}
\begin{scope}
\path[clip] ( 31.71,140.40) rectangle (151.02,158.03);
\definecolor{drawColor}{gray}{0.20}
\definecolor{fillColor}{gray}{0.85}

\path[draw=drawColor,line width= 0.6pt,line join=round,line cap=round,fill=fillColor] ( 31.71,140.40) rectangle (151.02,158.03);
\definecolor{drawColor}{gray}{0.10}

\node[text=drawColor,anchor=base,inner sep=0pt, outer sep=0pt, scale=  1.00] at ( 91.36,145.77) {Stocks};
\end{scope}
\begin{scope}
\path[clip] (170.26,140.40) rectangle (289.57,158.03);
\definecolor{drawColor}{gray}{0.20}
\definecolor{fillColor}{gray}{0.85}

\path[draw=drawColor,line width= 0.6pt,line join=round,line cap=round,fill=fillColor] (170.26,140.40) rectangle (289.57,158.03);
\definecolor{drawColor}{gray}{0.10}

\node[text=drawColor,anchor=base,inner sep=0pt, outer sep=0pt, scale=  1.00] at (229.92,145.77) {Average};
\end{scope}
\begin{scope}
\path[clip] ( 31.71,285.97) rectangle (151.02,303.60);
\definecolor{drawColor}{gray}{0.20}
\definecolor{fillColor}{gray}{0.85}

\path[draw=drawColor,line width= 0.6pt,line join=round,line cap=round,fill=fillColor] ( 31.71,285.97) rectangle (151.02,303.60);
\definecolor{drawColor}{gray}{0.10}

\node[text=drawColor,anchor=base,inner sep=0pt, outer sep=0pt, scale=  1.00] at ( 91.36,291.34) {Feuilleton};
\end{scope}
\begin{scope}
\path[clip] (170.26,285.97) rectangle (289.57,303.60);
\definecolor{drawColor}{gray}{0.20}
\definecolor{fillColor}{gray}{0.85}

\path[draw=drawColor,line width= 0.6pt,line join=round,line cap=round,fill=fillColor] (170.26,285.97) rectangle (289.57,303.60);
\definecolor{drawColor}{gray}{0.10}

\node[text=drawColor,anchor=base,inner sep=0pt, outer sep=0pt, scale=  1.00] at (229.92,291.34) {Weather};
\end{scope}
\begin{scope}
\path[clip] (308.82,285.97) rectangle (428.12,303.60);
\definecolor{drawColor}{gray}{0.20}
\definecolor{fillColor}{gray}{0.85}

\path[draw=drawColor,line width= 0.6pt,line join=round,line cap=round,fill=fillColor] (308.82,285.97) rectangle (428.12,303.60);
\definecolor{drawColor}{gray}{0.10}

\node[text=drawColor,anchor=base,inner sep=0pt, outer sep=0pt, scale=  1.00] at (368.47,291.34) {Death notice};
\end{scope}
\begin{scope}
\path[clip] (  0.00,  0.00) rectangle (433.62,325.21);
\definecolor{drawColor}{gray}{0.20}

\path[draw=drawColor,line width= 0.6pt,line join=round] ( 64.25, 27.94) --
	( 64.25, 30.69);

\path[draw=drawColor,line width= 0.6pt,line join=round] (118.48, 27.94) --
	(118.48, 30.69);
\end{scope}
\begin{scope}
\path[clip] (  0.00,  0.00) rectangle (433.62,325.21);
\definecolor{drawColor}{gray}{0.30}

\node[text=drawColor,anchor=base,inner sep=0pt, outer sep=0pt, scale=  0.88] at ( 64.25, 19.68) {1400};

\node[text=drawColor,anchor=base,inner sep=0pt, outer sep=0pt, scale=  0.88] at (118.48, 19.68) {800};
\end{scope}
\begin{scope}
\path[clip] (  0.00,  0.00) rectangle (433.62,325.21);
\definecolor{drawColor}{gray}{0.20}

\path[draw=drawColor,line width= 0.6pt,line join=round] (202.80, 27.94) --
	(202.80, 30.69);

\path[draw=drawColor,line width= 0.6pt,line join=round] (257.03, 27.94) --
	(257.03, 30.69);
\end{scope}
\begin{scope}
\path[clip] (  0.00,  0.00) rectangle (433.62,325.21);
\definecolor{drawColor}{gray}{0.30}

\node[text=drawColor,anchor=base,inner sep=0pt, outer sep=0pt, scale=  0.88] at (202.80, 19.68) {1400};

\node[text=drawColor,anchor=base,inner sep=0pt, outer sep=0pt, scale=  0.88] at (257.03, 19.68) {800};
\end{scope}
\begin{scope}
\path[clip] (  0.00,  0.00) rectangle (433.62,325.21);
\definecolor{drawColor}{gray}{0.20}

\path[draw=drawColor,line width= 0.6pt,line join=round] ( 64.25,173.50) --
	( 64.25,176.25);

\path[draw=drawColor,line width= 0.6pt,line join=round] (118.48,173.50) --
	(118.48,176.25);
\end{scope}
\begin{scope}
\path[clip] (  0.00,  0.00) rectangle (433.62,325.21);
\definecolor{drawColor}{gray}{0.30}

\node[text=drawColor,anchor=base,inner sep=0pt, outer sep=0pt, scale=  0.88] at ( 64.25,165.24) {1400};

\node[text=drawColor,anchor=base,inner sep=0pt, outer sep=0pt, scale=  0.88] at (118.48,165.24) {800};
\end{scope}
\begin{scope}
\path[clip] (  0.00,  0.00) rectangle (433.62,325.21);
\definecolor{drawColor}{gray}{0.20}

\path[draw=drawColor,line width= 0.6pt,line join=round] (202.80,173.50) --
	(202.80,176.25);

\path[draw=drawColor,line width= 0.6pt,line join=round] (257.03,173.50) --
	(257.03,176.25);
\end{scope}
\begin{scope}
\path[clip] (  0.00,  0.00) rectangle (433.62,325.21);
\definecolor{drawColor}{gray}{0.30}

\node[text=drawColor,anchor=base,inner sep=0pt, outer sep=0pt, scale=  0.88] at (202.80,165.24) {1400};

\node[text=drawColor,anchor=base,inner sep=0pt, outer sep=0pt, scale=  0.88] at (257.03,165.24) {800};
\end{scope}
\begin{scope}
\path[clip] (  0.00,  0.00) rectangle (433.62,325.21);
\definecolor{drawColor}{gray}{0.20}

\path[draw=drawColor,line width= 0.6pt,line join=round] (341.35,173.50) --
	(341.35,176.25);

\path[draw=drawColor,line width= 0.6pt,line join=round] (395.58,173.50) --
	(395.58,176.25);
\end{scope}
\begin{scope}
\path[clip] (  0.00,  0.00) rectangle (433.62,325.21);
\definecolor{drawColor}{gray}{0.30}

\node[text=drawColor,anchor=base,inner sep=0pt, outer sep=0pt, scale=  0.88] at (341.35,165.24) {1400};

\node[text=drawColor,anchor=base,inner sep=0pt, outer sep=0pt, scale=  0.88] at (395.58,165.24) {800};
\end{scope}
\begin{scope}
\path[clip] (  0.00,  0.00) rectangle (433.62,325.21);
\definecolor{drawColor}{gray}{0.30}

\node[text=drawColor,anchor=base east,inner sep=0pt, outer sep=0pt, scale=  0.88] at (303.87,182.02) {50};

\node[text=drawColor,anchor=base east,inner sep=0pt, outer sep=0pt, scale=  0.88] at (303.87,208.18) {60};

\node[text=drawColor,anchor=base east,inner sep=0pt, outer sep=0pt, scale=  0.88] at (303.87,234.34) {70};

\node[text=drawColor,anchor=base east,inner sep=0pt, outer sep=0pt, scale=  0.88] at (303.87,260.51) {80};
\end{scope}
\begin{scope}
\path[clip] (  0.00,  0.00) rectangle (433.62,325.21);
\definecolor{drawColor}{gray}{0.20}

\path[draw=drawColor,line width= 0.6pt,line join=round] (306.07,185.05) --
	(308.82,185.05);

\path[draw=drawColor,line width= 0.6pt,line join=round] (306.07,211.21) --
	(308.82,211.21);

\path[draw=drawColor,line width= 0.6pt,line join=round] (306.07,237.38) --
	(308.82,237.38);

\path[draw=drawColor,line width= 0.6pt,line join=round] (306.07,263.54) --
	(308.82,263.54);
\end{scope}
\begin{scope}
\path[clip] (  0.00,  0.00) rectangle (433.62,325.21);
\definecolor{drawColor}{gray}{0.30}

\node[text=drawColor,anchor=base east,inner sep=0pt, outer sep=0pt, scale=  0.88] at (165.31,181.52) {60};

\node[text=drawColor,anchor=base east,inner sep=0pt, outer sep=0pt, scale=  0.88] at (165.31,200.89) {65};

\node[text=drawColor,anchor=base east,inner sep=0pt, outer sep=0pt, scale=  0.88] at (165.31,220.26) {70};

\node[text=drawColor,anchor=base east,inner sep=0pt, outer sep=0pt, scale=  0.88] at (165.31,239.62) {75};

\node[text=drawColor,anchor=base east,inner sep=0pt, outer sep=0pt, scale=  0.88] at (165.31,258.99) {80};

\node[text=drawColor,anchor=base east,inner sep=0pt, outer sep=0pt, scale=  0.88] at (165.31,278.36) {85};
\end{scope}
\begin{scope}
\path[clip] (  0.00,  0.00) rectangle (433.62,325.21);
\definecolor{drawColor}{gray}{0.20}

\path[draw=drawColor,line width= 0.6pt,line join=round] (167.51,184.55) --
	(170.26,184.55);

\path[draw=drawColor,line width= 0.6pt,line join=round] (167.51,203.92) --
	(170.26,203.92);

\path[draw=drawColor,line width= 0.6pt,line join=round] (167.51,223.29) --
	(170.26,223.29);

\path[draw=drawColor,line width= 0.6pt,line join=round] (167.51,242.65) --
	(170.26,242.65);

\path[draw=drawColor,line width= 0.6pt,line join=round] (167.51,262.02) --
	(170.26,262.02);

\path[draw=drawColor,line width= 0.6pt,line join=round] (167.51,281.39) --
	(170.26,281.39);
\end{scope}
\begin{scope}
\path[clip] (  0.00,  0.00) rectangle (433.62,325.21);
\definecolor{drawColor}{gray}{0.30}

\node[text=drawColor,anchor=base east,inner sep=0pt, outer sep=0pt, scale=  0.88] at (165.31, 37.95) {65};

\node[text=drawColor,anchor=base east,inner sep=0pt, outer sep=0pt, scale=  0.88] at (165.31, 60.96) {70};

\node[text=drawColor,anchor=base east,inner sep=0pt, outer sep=0pt, scale=  0.88] at (165.31, 83.96) {75};

\node[text=drawColor,anchor=base east,inner sep=0pt, outer sep=0pt, scale=  0.88] at (165.31,106.97) {80};

\node[text=drawColor,anchor=base east,inner sep=0pt, outer sep=0pt, scale=  0.88] at (165.31,129.98) {85};
\end{scope}
\begin{scope}
\path[clip] (  0.00,  0.00) rectangle (433.62,325.21);
\definecolor{drawColor}{gray}{0.20}

\path[draw=drawColor,line width= 0.6pt,line join=round] (167.51, 40.98) --
	(170.26, 40.98);

\path[draw=drawColor,line width= 0.6pt,line join=round] (167.51, 63.99) --
	(170.26, 63.99);

\path[draw=drawColor,line width= 0.6pt,line join=round] (167.51, 87.00) --
	(170.26, 87.00);

\path[draw=drawColor,line width= 0.6pt,line join=round] (167.51,110.00) --
	(170.26,110.00);

\path[draw=drawColor,line width= 0.6pt,line join=round] (167.51,133.01) --
	(170.26,133.01);
\end{scope}
\begin{scope}
\path[clip] (  0.00,  0.00) rectangle (433.62,325.21);
\definecolor{drawColor}{gray}{0.30}

\node[text=drawColor,anchor=base east,inner sep=0pt, outer sep=0pt, scale=  0.88] at ( 26.76,203.19) {40};

\node[text=drawColor,anchor=base east,inner sep=0pt, outer sep=0pt, scale=  0.88] at ( 26.76,237.42) {60};

\node[text=drawColor,anchor=base east,inner sep=0pt, outer sep=0pt, scale=  0.88] at ( 26.76,271.65) {80};
\end{scope}
\begin{scope}
\path[clip] (  0.00,  0.00) rectangle (433.62,325.21);
\definecolor{drawColor}{gray}{0.20}

\path[draw=drawColor,line width= 0.6pt,line join=round] ( 28.96,206.22) --
	( 31.71,206.22);

\path[draw=drawColor,line width= 0.6pt,line join=round] ( 28.96,240.45) --
	( 31.71,240.45);

\path[draw=drawColor,line width= 0.6pt,line join=round] ( 28.96,274.68) --
	( 31.71,274.68);
\end{scope}
\begin{scope}
\path[clip] (  0.00,  0.00) rectangle (433.62,325.21);
\definecolor{drawColor}{gray}{0.30}

\node[text=drawColor,anchor=base east,inner sep=0pt, outer sep=0pt, scale=  0.88] at ( 26.76, 27.66) {70};

\node[text=drawColor,anchor=base east,inner sep=0pt, outer sep=0pt, scale=  0.88] at ( 26.76, 58.75) {75};

\node[text=drawColor,anchor=base east,inner sep=0pt, outer sep=0pt, scale=  0.88] at ( 26.76, 89.85) {80};

\node[text=drawColor,anchor=base east,inner sep=0pt, outer sep=0pt, scale=  0.88] at ( 26.76,120.94) {85};
\end{scope}
\begin{scope}
\path[clip] (  0.00,  0.00) rectangle (433.62,325.21);
\definecolor{drawColor}{gray}{0.20}

\path[draw=drawColor,line width= 0.6pt,line join=round] ( 28.96, 30.69) --
	( 31.71, 30.69);

\path[draw=drawColor,line width= 0.6pt,line join=round] ( 28.96, 61.78) --
	( 31.71, 61.78);

\path[draw=drawColor,line width= 0.6pt,line join=round] ( 28.96, 92.88) --
	( 31.71, 92.88);

\path[draw=drawColor,line width= 0.6pt,line join=round] ( 28.96,123.97) --
	( 31.71,123.97);
\end{scope}
\begin{scope}
\path[clip] (  0.00,  0.00) rectangle (433.62,325.21);
\definecolor{drawColor}{RGB}{0,0,0}

\node[text=drawColor,anchor=base,inner sep=0pt, outer sep=0pt, scale=  1.10] at (229.92,  7.64) {Number of pages};
\end{scope}
\begin{scope}
\path[clip] (  0.00,  0.00) rectangle (433.62,325.21);
\definecolor{drawColor}{RGB}{0,0,0}

\node[text=drawColor,rotate= 90.00,anchor=base,inner sep=0pt, outer sep=0pt, scale=  1.10] at ( 13.08,158.33) {mIoU};
\end{scope}
\begin{scope}
\path[clip] (  0.00,  0.00) rectangle (433.62,325.21);
\definecolor{fillColor}{RGB}{255,255,255}

\path[fill=fillColor] (320.65, 49.45) rectangle (400.26,124.53);
\end{scope}
\begin{scope}
\path[clip] (  0.00,  0.00) rectangle (433.62,325.21);
\definecolor{drawColor}{RGB}{0,0,0}

\node[text=drawColor,anchor=base,inner sep=0pt, outer sep=0pt, scale=  1.10] at (360.45,110.39) {Modalities};
\end{scope}
\begin{scope}
\path[clip] (  0.00,  0.00) rectangle (433.62,325.21);
\definecolor{fillColor}{RGB}{255,255,255}

\path[fill=fillColor] (326.15, 82.14) rectangle (340.60,103.82);
\end{scope}
\begin{scope}
\path[clip] (  0.00,  0.00) rectangle (433.62,325.21);
\definecolor{drawColor}{gray}{0.20}

\path[draw=drawColor,line width= 0.6pt,line join=round,line cap=round] (333.37, 84.30) --
	(333.37, 87.56);

\path[draw=drawColor,line width= 0.6pt,line join=round,line cap=round] (333.37, 98.40) --
	(333.37,101.65);
\definecolor{fillColor}{RGB}{254,255,254}

\path[draw=drawColor,line width= 0.6pt,line join=round,line cap=round,fill=fillColor] (327.95, 87.56) rectangle (338.80, 98.40);

\path[draw=drawColor,line width= 0.6pt,line join=round,line cap=round] (327.95, 92.98) --
	(338.80, 92.98);
\end{scope}
\begin{scope}
\path[clip] (  0.00,  0.00) rectangle (433.62,325.21);
\definecolor{fillColor}{RGB}{255,255,255}

\path[fill=fillColor] (326.15, 54.95) rectangle (340.60, 76.64);
\end{scope}
\begin{scope}
\path[clip] (  0.00,  0.00) rectangle (433.62,325.21);
\definecolor{drawColor}{gray}{0.20}

\path[draw=drawColor,line width= 0.6pt,line join=round,line cap=round] (333.37, 57.12) --
	(333.37, 60.38);

\path[draw=drawColor,line width= 0.6pt,line join=round,line cap=round] (333.37, 71.22) --
	(333.37, 74.47);
\definecolor{fillColor}{RGB}{255,221,227}

\path[draw=drawColor,line width= 0.6pt,line join=round,line cap=round,fill=fillColor] (327.95, 60.38) rectangle (338.80, 71.22);

\path[draw=drawColor,line width= 0.6pt,line join=round,line cap=round] (327.95, 65.80) --
	(338.80, 65.80);
\end{scope}
\begin{scope}
\path[clip] (  0.00,  0.00) rectangle (433.62,325.21);
\definecolor{drawColor}{RGB}{0,0,0}

\node[text=drawColor,anchor=base west,inner sep=0pt, outer sep=0pt, scale=  0.80] at (346.10, 90.22) {Image};
\end{scope}
\begin{scope}
\path[clip] (  0.00,  0.00) rectangle (433.62,325.21);
\definecolor{drawColor}{RGB}{0,0,0}

\node[text=drawColor,anchor=base west,inner sep=0pt, outer sep=0pt, scale=  0.80] at (346.10, 63.04) {Image + Text};
\end{scope}
\begin{scope}
\path[clip] (  0.00,  0.00) rectangle (433.62,325.21);
\definecolor{drawColor}{RGB}{0,0,0}

\node[text=drawColor,anchor=base,inner sep=0pt, outer sep=0pt, scale=  1.20] at (229.92,311.43) {\bfseries Comparison of mIoU per experiment};
\end{scope}
\end{tikzpicture}

%% file: Figures/luxwort.tex
\begin{tikzpicture}[x=1pt,y=1pt]
\definecolor{fillColor}{RGB}{255,255,255}
\path[use as bounding box,fill=fillColor,fill opacity=0.00] (0,0) rectangle (433.62,325.21);
\begin{scope}
\path[clip] (  0.00,  0.00) rectangle (433.62,325.21);
\definecolor{drawColor}{RGB}{255,255,255}
\definecolor{fillColor}{RGB}{255,255,255}

\path[draw=drawColor,line width= 0.6pt,line join=round,line cap=round,fill=fillColor] (  0.00,  0.00) rectangle (433.62,325.21);
\end{scope}
\begin{scope}
\path[clip] ( 33.44, 32.41) rectangle (281.85,301.87);
\definecolor{fillColor}{RGB}{255,255,255}

\path[fill=fillColor] ( 33.44, 32.41) rectangle (281.85,301.87);
\definecolor{drawColor}{gray}{0.92}

\path[draw=drawColor,line width= 0.3pt,line join=round] ( 33.44, 55.09) --
	(281.85, 55.09);

\path[draw=drawColor,line width= 0.3pt,line join=round] ( 33.44,109.79) --
	(281.85,109.79);

\path[draw=drawColor,line width= 0.3pt,line join=round] ( 33.44,164.50) --
	(281.85,164.50);

\path[draw=drawColor,line width= 0.3pt,line join=round] ( 33.44,219.21) --
	(281.85,219.21);

\path[draw=drawColor,line width= 0.3pt,line join=round] ( 33.44,273.91) --
	(281.85,273.91);

\path[draw=drawColor,line width= 0.6pt,line join=round] ( 33.44, 82.44) --
	(281.85, 82.44);

\path[draw=drawColor,line width= 0.6pt,line join=round] ( 33.44,137.15) --
	(281.85,137.15);

\path[draw=drawColor,line width= 0.6pt,line join=round] ( 33.44,191.85) --
	(281.85,191.85);

\path[draw=drawColor,line width= 0.6pt,line join=round] ( 33.44,246.56) --
	(281.85,246.56);

\path[draw=drawColor,line width= 0.6pt,line join=round] ( 33.44,301.26) --
	(281.85,301.26);

\path[draw=drawColor,line width= 0.6pt,line join=round] ( 57.48, 32.41) --
	( 57.48,301.87);

\path[draw=drawColor,line width= 0.6pt,line join=round] ( 97.55, 32.41) --
	( 97.55,301.87);

\path[draw=drawColor,line width= 0.6pt,line join=round] (137.61, 32.41) --
	(137.61,301.87);

\path[draw=drawColor,line width= 0.6pt,line join=round] (177.68, 32.41) --
	(177.68,301.87);

\path[draw=drawColor,line width= 0.6pt,line join=round] (217.75, 32.41) --
	(217.75,301.87);

\path[draw=drawColor,line width= 0.6pt,line join=round] (257.81, 32.41) --
	(257.81,301.87);
\definecolor{drawColor}{gray}{0.20}

\path[draw=drawColor,line width= 0.6pt,line join=round] ( 45.46,231.10) -- ( 45.46,235.63);

\path[draw=drawColor,line width= 0.6pt,line join=round] ( 45.46,193.25) -- ( 45.46,146.35);
\definecolor{fillColor}{RGB}{254,255,254}

\path[draw=drawColor,line width= 0.6pt,line join=round,line cap=round,fill=fillColor] ( 40.45,231.10) --
	( 40.45,193.25) --
	( 50.47,193.25) --
	( 50.47,231.10) --
	( 40.45,231.10) --
	cycle;

\path[draw=drawColor,line width= 1.1pt,line join=round] ( 40.45,210.30) -- ( 50.47,210.30);

\path[draw=drawColor,line width= 0.6pt,line join=round] ( 57.48,253.99) -- ( 57.48,259.68);

\path[draw=drawColor,line width= 0.6pt,line join=round] ( 57.48,241.77) -- ( 57.48,229.99);
\definecolor{fillColor}{RGB}{255,221,227}

\path[draw=drawColor,line width= 0.6pt,line join=round,line cap=round,fill=fillColor] ( 52.47,253.99) --
	( 52.47,241.77) --
	( 62.49,241.77) --
	( 62.49,253.99) --
	( 52.47,253.99) --
	cycle;

\path[draw=drawColor,line width= 1.1pt,line join=round] ( 52.47,247.71) -- ( 62.49,247.71);
\definecolor{fillColor}{gray}{0.20}

\path[draw=drawColor,line width= 0.4pt,line join=round,line cap=round,fill=fillColor] ( 69.50,235.69) circle (  1.96);

\path[draw=drawColor,line width= 0.6pt,line join=round] ( 69.50,272.75) -- ( 69.50,277.62);

\path[draw=drawColor,line width= 0.6pt,line join=round] ( 69.50,263.05) -- ( 69.50,258.88);
\definecolor{fillColor}{RGB}{255,229,217}

\path[draw=drawColor,line width= 0.6pt,line join=round,line cap=round,fill=fillColor] ( 64.49,272.75) --
	( 64.49,263.05) --
	( 74.51,263.05) --
	( 74.51,272.75) --
	( 64.49,272.75) --
	cycle;

\path[draw=drawColor,line width= 1.1pt,line join=round] ( 64.49,268.99) -- ( 74.51,268.99);

\path[draw=drawColor,line width= 0.6pt,line join=round] ( 85.53,237.87) -- ( 85.53,249.87);

\path[draw=drawColor,line width= 0.6pt,line join=round] ( 85.53,212.67) -- ( 85.53,198.31);
\definecolor{fillColor}{RGB}{254,255,254}

\path[draw=drawColor,line width= 0.6pt,line join=round,line cap=round,fill=fillColor] ( 80.52,237.87) --
	( 80.52,212.67) --
	( 90.53,212.67) --
	( 90.53,237.87) --
	( 80.52,237.87) --
	cycle;

\path[draw=drawColor,line width= 1.1pt,line join=round] ( 80.52,227.73) -- ( 90.53,227.73);
\definecolor{fillColor}{gray}{0.20}

\path[draw=drawColor,line width= 0.4pt,line join=round,line cap=round,fill=fillColor] ( 97.55,244.09) circle (  1.96);

\path[draw=drawColor,line width= 0.6pt,line join=round] ( 97.55,268.97) -- ( 97.55,272.74);

\path[draw=drawColor,line width= 0.6pt,line join=round] ( 97.55,262.06) -- ( 97.55,256.83);
\definecolor{fillColor}{RGB}{255,221,227}

\path[draw=drawColor,line width= 0.6pt,line join=round,line cap=round,fill=fillColor] ( 92.54,268.97) --
	( 92.54,262.06) --
	(102.55,262.06) --
	(102.55,268.97) --
	( 92.54,268.97) --
	cycle;

\path[draw=drawColor,line width= 1.1pt,line join=round] ( 92.54,264.86) -- (102.55,264.86);
\definecolor{fillColor}{gray}{0.20}

\path[draw=drawColor,line width= 0.4pt,line join=round,line cap=round,fill=fillColor] (109.57,268.78) circle (  1.96);

\path[draw=drawColor,line width= 0.6pt,line join=round] (109.57,282.46) -- (109.57,283.61);

\path[draw=drawColor,line width= 0.6pt,line join=round] (109.57,277.90) -- (109.57,277.64);
\definecolor{fillColor}{RGB}{255,229,217}

\path[draw=drawColor,line width= 0.6pt,line join=round,line cap=round,fill=fillColor] (104.56,282.46) --
	(104.56,277.90) --
	(114.57,277.90) --
	(114.57,282.46) --
	(104.56,282.46) --
	cycle;

\path[draw=drawColor,line width= 1.1pt,line join=round] (104.56,279.60) -- (114.57,279.60);
\definecolor{fillColor}{gray}{0.20}

\path[draw=drawColor,line width= 0.4pt,line join=round,line cap=round,fill=fillColor] (125.59, 44.66) circle (  1.96);

\path[draw=drawColor,line width= 0.6pt,line join=round] (125.59,237.58) -- (125.59,249.49);

\path[draw=drawColor,line width= 0.6pt,line join=round] (125.59,162.71) -- (125.59, 83.22);
\definecolor{fillColor}{RGB}{254,255,254}

\path[draw=drawColor,line width= 0.6pt,line join=round,line cap=round,fill=fillColor] (120.58,237.58) --
	(120.58,162.71) --
	(130.60,162.71) --
	(130.60,237.58) --
	(120.58,237.58) --
	cycle;

\path[draw=drawColor,line width= 1.1pt,line join=round] (120.58,218.45) -- (130.60,218.45);

\path[draw=drawColor,line width= 0.6pt,line join=round] (137.61,269.50) -- (137.61,272.37);

\path[draw=drawColor,line width= 0.6pt,line join=round] (137.61,255.57) -- (137.61,243.22);
\definecolor{fillColor}{RGB}{255,221,227}

\path[draw=drawColor,line width= 0.6pt,line join=round,line cap=round,fill=fillColor] (132.60,269.50) --
	(132.60,255.57) --
	(142.62,255.57) --
	(142.62,269.50) --
	(132.60,269.50) --
	cycle;

\path[draw=drawColor,line width= 1.1pt,line join=round] (132.60,264.89) -- (142.62,264.89);
\definecolor{fillColor}{gray}{0.20}

\path[draw=drawColor,line width= 0.4pt,line join=round,line cap=round,fill=fillColor] (149.63,272.23) circle (  1.96);

\path[draw=drawColor,line width= 0.6pt,line join=round] (149.63,281.40) -- (149.63,283.11);

\path[draw=drawColor,line width= 0.6pt,line join=round] (149.63,277.79) -- (149.63,274.82);
\definecolor{fillColor}{RGB}{255,229,217}

\path[draw=drawColor,line width= 0.6pt,line join=round,line cap=round,fill=fillColor] (144.62,281.40) --
	(144.62,277.79) --
	(154.64,277.79) --
	(154.64,281.40) --
	(144.62,281.40) --
	cycle;

\path[draw=drawColor,line width= 1.1pt,line join=round] (144.62,279.28) -- (154.64,279.28);

\path[draw=drawColor,line width= 0.6pt,line join=round] (165.66,213.11) -- (165.66,240.15);

\path[draw=drawColor,line width= 0.6pt,line join=round] (165.66,168.72) -- (165.66,105.24);
\definecolor{fillColor}{RGB}{254,255,254}

\path[draw=drawColor,line width= 0.6pt,line join=round,line cap=round,fill=fillColor] (160.65,213.11) --
	(160.65,168.72) --
	(170.67,168.72) --
	(170.67,213.11) --
	(160.65,213.11) --
	cycle;

\path[draw=drawColor,line width= 1.1pt,line join=round] (160.65,206.31) -- (170.67,206.31);
\definecolor{fillColor}{gray}{0.20}

\path[draw=drawColor,line width= 0.4pt,line join=round,line cap=round,fill=fillColor] (177.68,254.13) circle (  1.96);

\path[draw=drawColor,line width= 0.6pt,line join=round] (177.68,276.20) -- (177.68,278.27);

\path[draw=drawColor,line width= 0.6pt,line join=round] (177.68,268.76) -- (177.68,259.58);
\definecolor{fillColor}{RGB}{255,221,227}

\path[draw=drawColor,line width= 0.6pt,line join=round,line cap=round,fill=fillColor] (172.67,276.20) --
	(172.67,268.76) --
	(182.69,268.76) --
	(182.69,276.20) --
	(172.67,276.20) --
	cycle;

\path[draw=drawColor,line width= 1.1pt,line join=round] (172.67,272.63) -- (182.69,272.63);

\path[draw=drawColor,line width= 0.6pt,line join=round] (189.70,282.65) -- (189.70,283.57);

\path[draw=drawColor,line width= 0.6pt,line join=round] (189.70,274.83) -- (189.70,271.05);
\definecolor{fillColor}{RGB}{255,229,217}

\path[draw=drawColor,line width= 0.6pt,line join=round,line cap=round,fill=fillColor] (184.69,282.65) --
	(184.69,274.83) --
	(194.71,274.83) --
	(194.71,282.65) --
	(184.69,282.65) --
	cycle;

\path[draw=drawColor,line width= 1.1pt,line join=round] (184.69,280.31) -- (194.71,280.31);
\definecolor{fillColor}{gray}{0.20}

\path[draw=drawColor,line width= 0.4pt,line join=round,line cap=round,fill=fillColor] (205.73,163.28) circle (  1.96);

\path[draw=drawColor,line width= 0.6pt,line join=round] (205.73,239.95) -- (205.73,246.43);

\path[draw=drawColor,line width= 0.6pt,line join=round] (205.73,213.86) -- (205.73,178.47);
\definecolor{fillColor}{RGB}{254,255,254}

\path[draw=drawColor,line width= 0.6pt,line join=round,line cap=round,fill=fillColor] (200.72,239.95) --
	(200.72,213.86) --
	(210.73,213.86) --
	(210.73,239.95) --
	(200.72,239.95) --
	cycle;

\path[draw=drawColor,line width= 1.1pt,line join=round] (200.72,230.90) -- (210.73,230.90);

\path[draw=drawColor,line width= 0.6pt,line join=round] (217.75,277.43) -- (217.75,278.26);

\path[draw=drawColor,line width= 0.6pt,line join=round] (217.75,273.72) -- (217.75,270.54);
\definecolor{fillColor}{RGB}{255,221,227}

\path[draw=drawColor,line width= 0.6pt,line join=round,line cap=round,fill=fillColor] (212.74,277.43) --
	(212.74,273.72) --
	(222.75,273.72) --
	(222.75,277.43) --
	(212.74,277.43) --
	cycle;

\path[draw=drawColor,line width= 1.1pt,line join=round] (212.74,276.29) -- (222.75,276.29);

\path[draw=drawColor,line width= 0.6pt,line join=round] (229.77,286.98) -- (229.77,289.00);

\path[draw=drawColor,line width= 0.6pt,line join=round] (229.77,285.08) -- (229.77,283.85);
\definecolor{fillColor}{RGB}{255,229,217}

\path[draw=drawColor,line width= 0.6pt,line join=round,line cap=round,fill=fillColor] (224.76,286.98) --
	(224.76,285.08) --
	(234.77,285.08) --
	(234.77,286.98) --
	(224.76,286.98) --
	cycle;

\path[draw=drawColor,line width= 1.1pt,line join=round] (224.76,286.24) -- (234.77,286.24);
\definecolor{fillColor}{gray}{0.20}

\path[draw=drawColor,line width= 0.4pt,line join=round,line cap=round,fill=fillColor] (245.79, 66.60) circle (  1.96);

\path[draw=drawColor,line width= 0.6pt,line join=round] (245.79,209.78) -- (245.79,239.24);

\path[draw=drawColor,line width= 0.6pt,line join=round] (245.79,153.87) -- (245.79,121.07);
\definecolor{fillColor}{RGB}{254,255,254}

\path[draw=drawColor,line width= 0.6pt,line join=round,line cap=round,fill=fillColor] (240.78,209.78) --
	(240.78,153.87) --
	(250.80,153.87) --
	(250.80,209.78) --
	(240.78,209.78) --
	cycle;

\path[draw=drawColor,line width= 1.1pt,line join=round] (240.78,182.05) -- (250.80,182.05);

\path[draw=drawColor,line width= 0.6pt,line join=round] (257.81,279.23) -- (257.81,282.56);

\path[draw=drawColor,line width= 0.6pt,line join=round] (257.81,274.71) -- (257.81,273.43);
\definecolor{fillColor}{RGB}{255,221,227}

\path[draw=drawColor,line width= 0.6pt,line join=round,line cap=round,fill=fillColor] (252.80,279.23) --
	(252.80,274.71) --
	(262.82,274.71) --
	(262.82,279.23) --
	(252.80,279.23) --
	cycle;

\path[draw=drawColor,line width= 1.1pt,line join=round] (252.80,277.60) -- (262.82,277.60);

\path[draw=drawColor,line width= 0.6pt,line join=round] (269.83,289.14) -- (269.83,289.63);

\path[draw=drawColor,line width= 0.6pt,line join=round] (269.83,286.94) -- (269.83,285.87);
\definecolor{fillColor}{RGB}{255,229,217}

\path[draw=drawColor,line width= 0.6pt,line join=round,line cap=round,fill=fillColor] (264.82,289.14) --
	(264.82,286.94) --
	(274.84,286.94) --
	(274.84,289.14) --
	(264.82,289.14) --
	cycle;

\path[draw=drawColor,line width= 1.1pt,line join=round] (264.82,287.64) -- (274.84,287.64);
\definecolor{drawColor}{RGB}{0,0,0}

\path[draw=drawColor,line width= 0.6pt,line join=round] ( 45.46,210.30) --
	( 85.53,227.73) --
	(125.59,218.45) --
	(165.66,206.31) --
	(205.73,230.90) --
	(245.79,182.05);

\path[draw=drawColor,line width= 0.6pt,line join=round] ( 57.48,247.71) --
	( 97.55,264.86) --
	(137.61,264.89) --
	(177.68,272.63) --
	(217.75,276.29) --
	(257.81,277.60);

\path[draw=drawColor,line width= 0.6pt,line join=round] ( 69.50,268.99) --
	(109.57,279.60) --
	(149.63,279.28) --
	(189.70,280.31) --
	(229.77,286.24) --
	(269.83,287.64);
\definecolor{drawColor}{gray}{0.20}

\path[draw=drawColor,line width= 0.6pt,line join=round,line cap=round] ( 33.44, 32.41) rectangle (281.85,301.87);
\end{scope}
\begin{scope}
\path[clip] (  0.00,  0.00) rectangle (433.62,325.21);
\definecolor{drawColor}{gray}{0.30}

\node[text=drawColor,anchor=base east,inner sep=0pt, outer sep=0pt, scale=  0.88] at ( 28.49, 79.41) {50};

\node[text=drawColor,anchor=base east,inner sep=0pt, outer sep=0pt, scale=  0.88] at ( 28.49,134.12) {60};

\node[text=drawColor,anchor=base east,inner sep=0pt, outer sep=0pt, scale=  0.88] at ( 28.49,188.82) {70};

\node[text=drawColor,anchor=base east,inner sep=0pt, outer sep=0pt, scale=  0.88] at ( 28.49,243.53) {80};

\node[text=drawColor,anchor=base east,inner sep=0pt, outer sep=0pt, scale=  0.88] at ( 28.49,298.23) {90};
\end{scope}
\begin{scope}
\path[clip] (  0.00,  0.00) rectangle (433.62,325.21);
\definecolor{drawColor}{gray}{0.20}

\path[draw=drawColor,line width= 0.6pt,line join=round] ( 30.69, 82.44) --
	( 33.44, 82.44);

\path[draw=drawColor,line width= 0.6pt,line join=round] ( 30.69,137.15) --
	( 33.44,137.15);

\path[draw=drawColor,line width= 0.6pt,line join=round] ( 30.69,191.85) --
	( 33.44,191.85);

\path[draw=drawColor,line width= 0.6pt,line join=round] ( 30.69,246.56) --
	( 33.44,246.56);

\path[draw=drawColor,line width= 0.6pt,line join=round] ( 30.69,301.26) --
	( 33.44,301.26);
\end{scope}
\begin{scope}
\path[clip] (  0.00,  0.00) rectangle (433.62,325.21);
\definecolor{drawColor}{gray}{0.20}

\path[draw=drawColor,line width= 0.6pt,line join=round] ( 57.48, 29.66) --
	( 57.48, 32.41);

\path[draw=drawColor,line width= 0.6pt,line join=round] ( 97.55, 29.66) --
	( 97.55, 32.41);

\path[draw=drawColor,line width= 0.6pt,line join=round] (137.61, 29.66) --
	(137.61, 32.41);

\path[draw=drawColor,line width= 0.6pt,line join=round] (177.68, 29.66) --
	(177.68, 32.41);

\path[draw=drawColor,line width= 0.6pt,line join=round] (217.75, 29.66) --
	(217.75, 32.41);

\path[draw=drawColor,line width= 0.6pt,line join=round] (257.81, 29.66) --
	(257.81, 32.41);
\end{scope}
\begin{scope}
\path[clip] (  0.00,  0.00) rectangle (433.62,325.21);
\definecolor{drawColor}{gray}{0.30}

\node[text=drawColor,anchor=base,inner sep=0pt, outer sep=0pt, scale=  0.88] at ( 57.48, 21.40) {550};

\node[text=drawColor,anchor=base,inner sep=0pt, outer sep=0pt, scale=  0.88] at ( 97.55, 21.40) {1100};

\node[text=drawColor,anchor=base,inner sep=0pt, outer sep=0pt, scale=  0.88] at (137.61, 21.40) {1650};

\node[text=drawColor,anchor=base,inner sep=0pt, outer sep=0pt, scale=  0.88] at (177.68, 21.40) {3300};

\node[text=drawColor,anchor=base,inner sep=0pt, outer sep=0pt, scale=  0.88] at (217.75, 21.40) {7000};

\node[text=drawColor,anchor=base,inner sep=0pt, outer sep=0pt, scale=  0.88] at (257.81, 21.40) {14100};
\end{scope}
\begin{scope}
\path[clip] (  0.00,  0.00) rectangle (433.62,325.21);
\definecolor{drawColor}{RGB}{0,0,0}

\node[text=drawColor,anchor=base,inner sep=0pt, outer sep=0pt, scale=  1.10] at (157.65,  9.37) {Number of pages};
\end{scope}
\begin{scope}
\path[clip] (  0.00,  0.00) rectangle (433.62,325.21);
\definecolor{drawColor}{RGB}{0,0,0}

\node[text=drawColor,rotate= 90.00,anchor=base,inner sep=0pt, outer sep=0pt, scale=  1.10] at ( 14.80,167.14) {mIoU};
\end{scope}
\begin{scope}
\path[clip] (  0.00,  0.00) rectangle (433.62,325.21);

\path[] (288.99, 99.78) rectangle (418.48,245.40);
\end{scope}
\begin{scope}
\path[clip] (  0.00,  0.00) rectangle (433.62,325.21);
\definecolor{drawColor}{RGB}{0,0,0}

\node[text=drawColor,anchor=base,inner sep=0pt, outer sep=0pt, scale=  1.10] at (353.73,231.25) {Modalities};
\end{scope}
\begin{scope}
\path[clip] (  0.00,  0.00) rectangle (433.62,325.21);
\definecolor{fillColor}{RGB}{255,255,255}

\path[fill=fillColor] (294.49,188.55) rectangle (316.17,224.68);
\end{scope}
\begin{scope}
\path[clip] (  0.00,  0.00) rectangle (433.62,325.21);
\definecolor{drawColor}{gray}{0.20}

\path[draw=drawColor,line width= 0.6pt,line join=round,line cap=round] (305.33,192.16) --
	(305.33,197.58);

\path[draw=drawColor,line width= 0.6pt,line join=round,line cap=round] (305.33,215.65) --
	(305.33,221.07);
\definecolor{fillColor}{RGB}{254,255,254}

\path[draw=drawColor,line width= 0.6pt,line join=round,line cap=round,fill=fillColor] (297.20,197.58) rectangle (313.46,215.65);

\path[draw=drawColor,line width= 0.6pt,line join=round,line cap=round] (297.20,206.62) --
	(313.46,206.62);
\end{scope}
\begin{scope}
\path[clip] (  0.00,  0.00) rectangle (433.62,325.21);
\definecolor{fillColor}{RGB}{255,255,255}

\path[fill=fillColor] (294.49,146.91) rectangle (316.17,183.05);
\end{scope}
\begin{scope}
\path[clip] (  0.00,  0.00) rectangle (433.62,325.21);
\definecolor{drawColor}{gray}{0.20}

\path[draw=drawColor,line width= 0.6pt,line join=round,line cap=round] (305.33,150.53) --
	(305.33,155.95);

\path[draw=drawColor,line width= 0.6pt,line join=round,line cap=round] (305.33,174.01) --
	(305.33,179.43);
\definecolor{fillColor}{RGB}{255,221,227}

\path[draw=drawColor,line width= 0.6pt,line join=round,line cap=round,fill=fillColor] (297.20,155.95) rectangle (313.46,174.01);

\path[draw=drawColor,line width= 0.6pt,line join=round,line cap=round] (297.20,164.98) --
	(313.46,164.98);
\end{scope}
\begin{scope}
\path[clip] (  0.00,  0.00) rectangle (433.62,325.21);
\definecolor{fillColor}{RGB}{255,255,255}

\path[fill=fillColor] (294.49,105.28) rectangle (316.17,141.41);
\end{scope}
\begin{scope}
\path[clip] (  0.00,  0.00) rectangle (433.62,325.21);
\definecolor{drawColor}{gray}{0.20}

\path[draw=drawColor,line width= 0.6pt,line join=round,line cap=round] (305.33,108.89) --
	(305.33,114.31);

\path[draw=drawColor,line width= 0.6pt,line join=round,line cap=round] (305.33,132.38) --
	(305.33,137.80);
\definecolor{fillColor}{RGB}{255,229,217}

\path[draw=drawColor,line width= 0.6pt,line join=round,line cap=round,fill=fillColor] (297.20,114.31) rectangle (313.46,132.38);

\path[draw=drawColor,line width= 0.6pt,line join=round,line cap=round] (297.20,123.35) --
	(313.46,123.35);
\end{scope}
\begin{scope}
\path[clip] (  0.00,  0.00) rectangle (433.62,325.21);
\definecolor{drawColor}{RGB}{0,0,0}

\node[text=drawColor,anchor=base west,inner sep=0pt, outer sep=0pt, scale=  0.80] at (321.67,203.86) {Image};
\end{scope}
\begin{scope}
\path[clip] (  0.00,  0.00) rectangle (433.62,325.21);
\definecolor{drawColor}{RGB}{0,0,0}

\node[text=drawColor,anchor=base west,inner sep=0pt, outer sep=0pt, scale=  0.80] at (321.67,162.23) {Image + Text out-domain};
\end{scope}
\begin{scope}
\path[clip] (  0.00,  0.00) rectangle (433.62,325.21);
\definecolor{drawColor}{RGB}{0,0,0}

\node[text=drawColor,anchor=base west,inner sep=0pt, outer sep=0pt, scale=  0.80] at (321.67,120.59) {Image + Text in-domain};
\end{scope}
\begin{scope}
\path[clip] (  0.00,  0.00) rectangle (433.62,325.21);
\definecolor{drawColor}{RGB}{0,0,0}

\node[text=drawColor,anchor=base,inner sep=0pt, outer sep=0pt, scale=  1.20] at (157.65,309.71) {\bfseries Comparison of mIoU per experiment};
\end{scope}
\end{tikzpicture}

%% file: 06-discussion.tex
\section*{Conclusion and Outlook}
\label{sec:discussion}
\vspace{1ex}

We believe these series of experiments led to a better understanding of the interplay between visual and textual features for semantic segmentation of newspapers document images.

The first series of experiments using annotation data that was representatively sampled show a consistent improvement for models that combine textual and visual features relative to a strong baseline using visual features only. Textual features also help to mitigate the problem of high variance that purely visual models have with the varying and diverse material in newspapers published over a long period of time.

The second series of experiments on  the generalization ability over time and across newspapers showed that a simple transfer of models leads to a stark drop of performance. However, on average, models with textual features show substantially better results than the ones without. We can conclude that text characteristics are indeed  more stable than layout characteristics and that they are vital for improving the model's robustness. For practical applications, these experiments make clear that annotation efforts need to be carefully distributed over the diachronic variety and diversity of the material present in historic newspapers.

The third series of experiments on the reduction of training data gave mixed results. On the one hand, the model using both visual and textual still improves on most classes. On the other, there seems to be, for some classes, a lower bound on the number of samples for the proposed method in order to be able to extract the relevant signals from textual features. 

The fourth series of experiments on the benefit of using large amounts of  training data---by taking advantage of the fact that a lot of material with useful semantic classification already exists in digitized archives--- showed that purely visual models have more difficulties to exploit a larger amount of data than combined models. Another important outcome of these experiments is the fact that more training material can not fully compensate for the availability of text embeddings specifically built on in-domain text data, especially with noisy OCR texts. As in-domain text embeddings can be computed without human annotation and are therefore  cost-effective, they should always be considered.

Although proof of concept, the present approach can already support two main use cases. First, similarly as \textit{dhSegment}, the released framework can help scholars and/or non-specialists to easily process document images, provided they can be associated with text (thanks to e.g. an open-source OCR software), and that embeddings are available. Second, even though not perfect, the models can already be used as support for manual annotation (users only need to correct false positives or negatives) and, for some classes, be used to segment real newspaper collections to offer further search facets and/or filter unwanted material.

As future works, we intend to compare this approach with pure text classification in order to bridge the comparison spectrum from pure pixel to pure text, as well as to apply it to other documents than newspapers. 
It could also be interesting to integrate a region proposal module (such as in Mask R-CNN \citep{he2017mask}) in order to segment at instance level.